\documentclass[runningheads]{llncs}

 \usepackage{eccv}

\usepackage{eccvabbrv}
\usepackage{adjustbox}
\usepackage{graphicx}
\usepackage{booktabs}
\usepackage{wrapfig}
\usepackage{arydshln}
\usepackage[accsupp]{axessibility}  % Improves PDF readability for those with disabilities.
\usepackage{algorithm}      % 提供 algorithm 浮动体环境
\usepackage{algorithmic}    % 提供 algorithmic 环境和命令（如 \REQUIRE、\ENSURE）version
\usepackage{booktabs}        % 提供 \toprule, \midrule, \bottomrule 等高质量横线
\usepackage{multirow}        % 支持多行合并
\usepackage{graphicx}        % 提供 \resizebox、插图等功能
\usepackage{array}           % 增强表格列格式控制
\usepackage{makecell}        % 允许单元格内换行、设置粗体/斜体等
\usepackage{tcolorbox}
\usepackage{caption}         % 更好地控制 caption 样式
\usepackage[table,xcdraw]{xcolor}
\usepackage{pifont}

\newcommand{\method}{MedSynapse-V\xspace}
\newcommand{\ours}{MedSynapse-V\xspace}
\definecolor{myred}{rgb}{0.996,0.578,0.574}
\definecolor{myyellow}{rgb}{0.988,0.961,0.898}
\definecolor{mylightred}{rgb}{0.992,0.887,0.883}
\newcolumntype{C}[1]{>{\centering\arraybackslash}p{#1}}
\definecolor{OurColor}{RGB}{220,240,255}  % 浅蓝色用于突出我们的方法

\usepackage{orcidlink}

\begin{document}

% ---------------------------------------------------------------
% TODO REVIEW: Replace with your title
%\title{\method: Anatomical Prior Retrieval and Autonomous Reasoning Internalization \\ for Medical Visual Diagnosis}
\title{MedSynapse-V: Bridging Visual Perception and Clinical Intuition via Latent Memory Evolution}

% TODO REVIEW: If the paper title is too long for the running head, you can set
% an abbreviated paper title here. If not, comment out.
\titlerunning{Medical Latent Memory Evolution}

% TODO FINAL: Replace with your author list. 
% Include the authors' OCRID for the camera-ready version, if at all possible.
%\author{First Author\inst{1}\orcidlink{0000-1111-2222-3333} \and
%Second Author\inst{2,3}\orcidlink{1111-2222-3333-4444} \and
%Third Author\inst{3}\orcidlink{2222--3333-4444-5555}}
%
%% TODO FINAL: Replace with an abbreviated list of authors.
%\authorrunning{F.~Author et al.}
%% First names are abbreviated in the running head.
%% If there are more than two authors, 'et al.' is used.
%
%% TODO FINAL: Replace with your institution list.
%\institute{Princeton University, Princeton NJ 08544, USA \and
%Springer Heidelberg, Tiergartenstr.~17, 69121 Heidelberg, Germany
%\email{lncs@springer.com}\\
%\url{http://www.springer.com/gp/computer-science/lncs} \and
%ABC Institute, Rupert-Karls-University Heidelberg, Heidelberg, Germany\\
%\email{\{abc,lncs\}@uni-heidelberg.de}}

\author{Chunzheng Zhu\inst{1} \and
Jiaqi Zeng\inst{1} \and
Junyu Jiang\inst{1} \and
Jianxin Lin\inst{1}\thanks{Corresponding author.} \and
Yijun Wang\inst{1}}

\authorrunning{C. Zhu et al.}

\institute{Hunan University, Changsha, China\\
\email{\{zhuchzh, zjqxxl, jiangjy, linjianxin, wyjun\}@hnu.edu.cn}}

\maketitle

\begin{abstract}
High-precision medical diagnosis relies not only on static imaging features but also on the implicit diagnostic memory experts instantly invoke during image interpretation. We pinpoint a fundamental cognitive misalignment in medical VLMs caused by discrete tokenization, leading to quantization loss, long-range information dissipation, and missing case-adaptive expertise. To bridge this gap, we propose \ours, a framework for latent diagnostic memory evolution that simulates the experiential invocation of clinicians by dynamically synthesizing implicit diagnostic memories within the model’s hidden stream. Specifically, it begins with a \textbf{\textit{Meta Query for Prior Memorization}} mechanism, where learnable probes retrieve structured priors from an anatomical prior encoder to generate condensed implicit memories. To ensure clinical fidelity, we introduce \textbf{\textit{Causal Counterfactual Refinement (CCR)}} which leverages reinforcement learning and counterfactual rewards derived from region-level feature masking to quantify the causal contribution of each memory, thereby pruning redundancies and aligning latent representations with diagnostic logic. This evolutionary process culminates in \textit{\textbf{Intrinsic Memory Transition (IMT)}}, a privileged-autonomous dual-branch paradigm that internalizes teacher-branch diagnostic patterns into the student-branch via full-vocabulary divergence alignment. Comprehensive empirical evaluations across multiple datasets demonstrate that \ours, by transferring external expertise into endogenous parameters, \textit{significantly outperforms existing state-of-the-art methods, particularly Chain-of-Thought (CoT) paradigms}, in diagnostic accuracy and multi-dataset generalization \textit{without compromising the inference efficiency} of standard VLMs.
\end{abstract}

\begin{figure*}[t]
\centering
\includegraphics[width=1\textwidth]{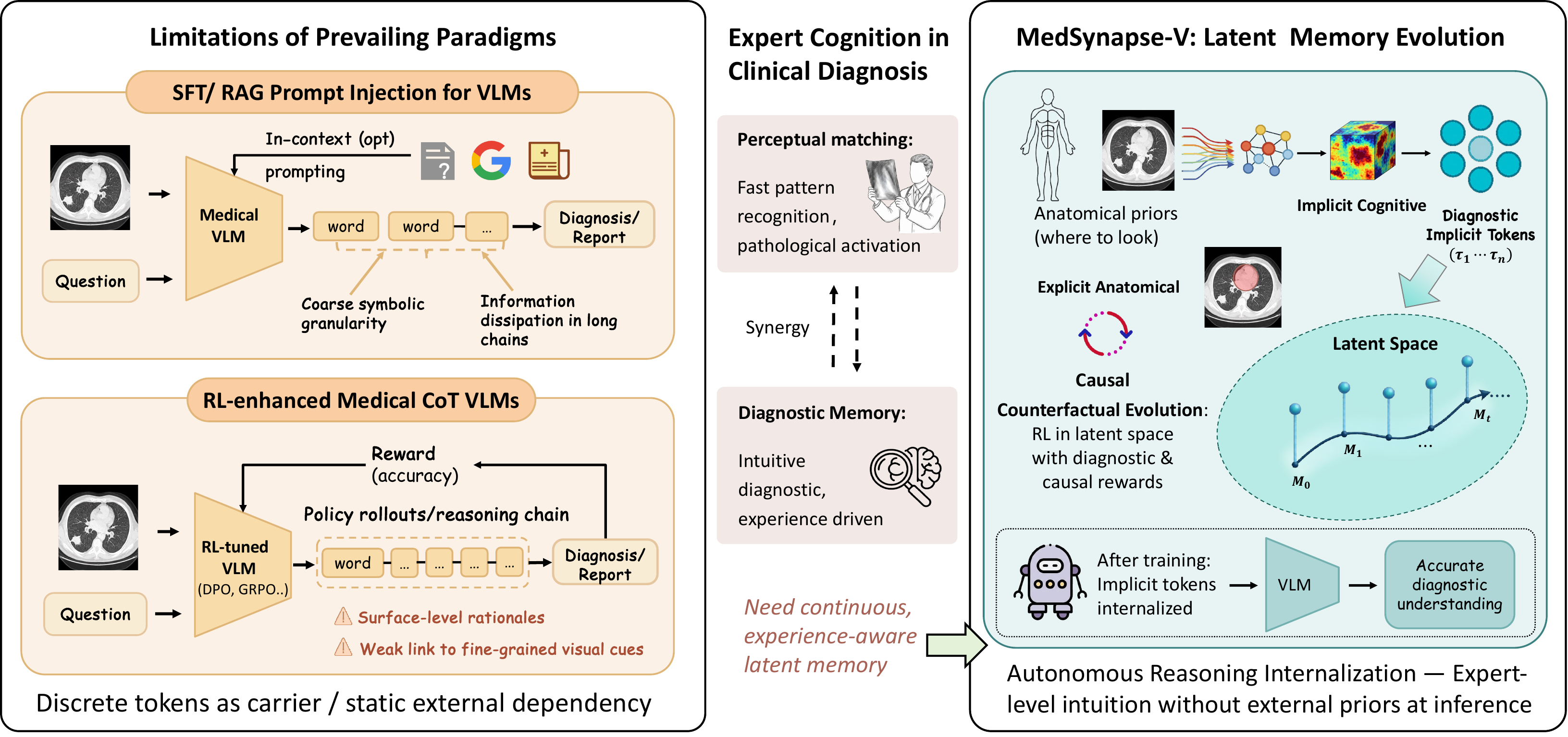}
\caption{Existing medical VLMs suffer from coarse symbolic granularity and long-range information dissipation in discrete reasoning. MedSynapse-V addresses this by evolving diagnostic implicit memory in latent space via anatomical prior condensation, causal counterfactual refinement, and autonomous latent memory internalization.}
\label{fig:motivation}
\end{figure*}

\section{Introduction}
\label{sec:introduction}

Seasoned diagnostic experts do not rely on stepwise logical reasoning when making clinical diagnoses; instead, they activate Implicit Diagnostic Memory, that enables near-instantaneous pattern recognition against accumulated case knowledge~\cite{norman2009dual,waite2017interpretive,brunye2019review}. Although medical vision-language models (VLMs) have made substantial progress in diagnostic assistance~\cite{wu2025towards,sellergren2025medgemma, chen2024towards,li2023llava,moor2023med}, with reinforcement learning from verifiable rewards~\cite{pan2025medvlm,lai2026med,sun2025chiron,su2025gmai} and chain-of-thought (CoT)~\cite{wang2025v2t,chen2025think,le2025s,wei2022chain,gai2024medthink,wu2025medreason} further advancing reasoning capabilities. However, their intrinsic reliance on discrete tokens engenders a profound \textit{Cognitive Misalignment} with the inherently continuous nature of clinical expertise. As illustrated in Fig.~\ref{fig:motivation}, the limited granularity of a fixed vocabulary is inadequate for representing continuous pathological features such as gradual transitions in lesion density or textural heterogeneity, and the autoregressive decoding mechanism is prone to progressive attenuation of visual evidence over extended reasoning chains. Moreover, discrete symbols tend to encode generic linguistic priors rather than dynamic anatomical context, readily giving rise to ``pseudo-logical'' hallucinations that lack grounding in physical evidence.

An intuitive remedy is to supplement models with external diagnostic knowledge. Retrieval-augmented generation (RAG) prepends retrieved text fragments or similar cases to the input context~\cite{zhao2025medrag,wu2024medical,arasteh2407radiorag,zhang2025patho,zakka2024almanac}, while soft-prompt and prefix-tuning methods concatenate learnable vectors to the input sequence to inject domain-specific cues~\cite{van2023open,gu2024lapa,lewis2020retrieval}. However, both strategies inject information that remains \textit{static} and \textit{causally unverified}: it has undergone neither validation of causal relevance to the current diagnostic decision nor evolution into an intrinsic model capability through gradient-based optimization, persisting as a brittle external dependency prone to context saturation and information redundancy as the differential diagnosis space expands.

Recent latent computation paradigms~\cite{hao2024training,shen2025codi,xu2025softcot,li2025seek} offer a principled alternative by performing reasoning in continuous hidden state spaces, circumventing the expressiveness bottleneck of discrete symbols. However, their direct application to medical scenarios encounters two domain-specific obstacles. First, without structured anatomical priors, latent representations degenerate into abstract vectors decoupled from clinical semantics: they capture statistical regularities of the training distribution but fail to encode the structured spatial relationships (organ topology, lesion morphology, tissue boundaries) essential for diagnostic grounding. Second, without causal calibration, the coupling between latent representations and diagnostically critical visual features remains weak, as the model may produce correct answers by exploiting spurious correlations (\eg, dataset-specific formatting cues) rather than attending to pathologically relevant regions, undermining reliability in clinical deployment.

%These observations converge on a fundamental question: \textit{Can a model autonomously evolve diagnostic memory equivalent to that of clinical experts within latent space, enabling instantaneous invocation of case-relevant experiential knowledge, while ensuring that this memory maintains causal consistency with the underlying clinical decision logic?}
These observations converge on a fundamental question: \textit{Can a VLM progressively evolve its latent memory to simulate clinical intuition, enabling the rapid synthesis of case-adaptive diagnostic patterns, while ensuring this autonomous internal reasoning stream and its continuous refinement effectively steer the model toward clinically reliable decisions?}

This paper proposes \ours, a framework for latent diagnostic memory evolution that addresses this question through three synergistic mechanisms operating in a progressive training paradigm. First, \textit{Meta Query for Prior Memorization} (\S\ref{sec:mqpm}) deploys learnable meta-query probes to retrieve multi-scale spatially aware features from a frozen anatomical encoder pre-trained on large-scale segmentation tasks, condensing them into compact diagnostic implicit memory vectors that are injected into the VLM's hidden stream. This mechanism bridges the representation gap between the encoder's anatomical feature space and the VLM's generation space. Second, \textit{Causal Counterfactual Refinement} (CCR; \S\ref{sec:ccr}) performs reinforcement learning-driven memory optimization, introducing a novel causal counterfactual reward that quantifies the causal diagnostic contribution of each memory element through region-level feature masking interventions. By contrasting model behavior under original versus intervened memory conditions, CCR systematically prunes causally irrelevant components while reinforcing those with genuine diagnostic utility. Third, \textit{Intrinsic Memory Transition} (IMT; \S\ref{sec:imt}) employs a privileged-autonomous dual-branch paradigm to distill the refined diagnostic patterns from a teacher branch (with the anatomical encoder) into a lightweight student branch via full-vocabulary Jensen-Shannon divergence alignment. At inference, the anatomical encoder is entirely removed, and the model generates diagnostic memory autonomously with computational overhead nearly identical to a standard VLM.

Comprehensive evaluations across seven medical multimodal benchmarks demonstrate that \method consistently outperforms a broad spectrum of state-of-the-art approaches, spanning medical-specific VLMs, RL-enhanced CoT paradigms, and general-purpose latent reasoning methods, in both diagnostic accuracy and cross-domain generalization, while introducing negligible additional inference cost compared with standard VLMs. Our main contributions are:

%Extensive experiments on multiple medical multimodal benchmarks demonstrate that \method consistently outperforms state-of-the-art approaches spanning medical-specific VLMs, RL-enhanced CoT paradigms, and general-purpose latent reasoning methods. Crucially, these gains are achieved \textit{without} the additional inference latency inherent to CoT reasoning, confirming latent memory evolution as a principled and efficient pathway toward endowing medical VLMs with expert-level clinical intuition. Our main contributions are:

\noindent\textbf{(1)} We propose \method, the first framework that evolves diagnostic implicit memory in latent space for medical diagnosis, shifting from static external knowledge injection to progressive, autonomous memory internalization.

\noindent\textbf{(2)} We design Meta Query–based Prior Memorization coupled with Causal Counterfactual Refinement (CCR), which distills anatomical priors into compact latent memory and calibrates it via counterfactual interventions to retain only causally grounded diagnostic components.

\noindent\textbf{(3)} We introduce Intrinsic Memory Transition (IMT), a privileged–autonomous dual-branch distillation paradigm that internalizes encoder-dependent memory into autonomously generated intrinsic memory via full-vocabulary divergence alignment, eliminating all auxiliary modules at inference.

\noindent\textbf{(4)} In multimodal benchmarks, \method consistently surpasses mainstream CoT paradigms in diagnostic accuracy while maintaining inference efficiency on par with standard VLMs, validating latent memory evolution as a principled alternative to discrete token reasoning.

\section{Methodology}
\label{sec:method}

\subsection{Problem Formulation and Architecture Overview}
\label{sec:overview}

\vspace{0.5mm}\noindent\textbf{Problem Formulation.}
Given an image $X \in \mathbb{R}^{H \times W \times 3}$ and a clinical query $q$, the objective is to generate an output sequence $y$ containing diagnostic analysis and final conclusions. Let $\pi_\theta$ denote the VLM policy, $\mathcal{E}_{\textit{ana}}$ the frozen pretrained anatomical encoder, and $\mathcal{P}_\phi$ the parameterized memory synthesis module. \method dynamically generates a set of diagnostic implicit memory $\mathcal{M} = \{m_1, \ldots, m_N\} \in \mathbb{R}^{N \times d_h}$ for injection into the VLM hidden stream, where $d_h$ is the hidden state dimensionality. The overall policy is formalized as:
\begin{equation}
\pi_\theta(y \mid X, q) = \pi_\theta\!\left(y \;\big|\; X,\, q,\, \mathcal{P}_\phi\!\bigl(\mathcal{E}_{\textit{ana}}(X)\bigr)\right).
\label{eq:overall}
\end{equation}
Unlike explicit CoT, which concatenates reasoning tokens to the input sequence, \method constructs a diagnostic experience invocation mechanism based on implicit memory within the representation space.

\begin{figure*}[t]
\centering
\includegraphics[width=1\textwidth]{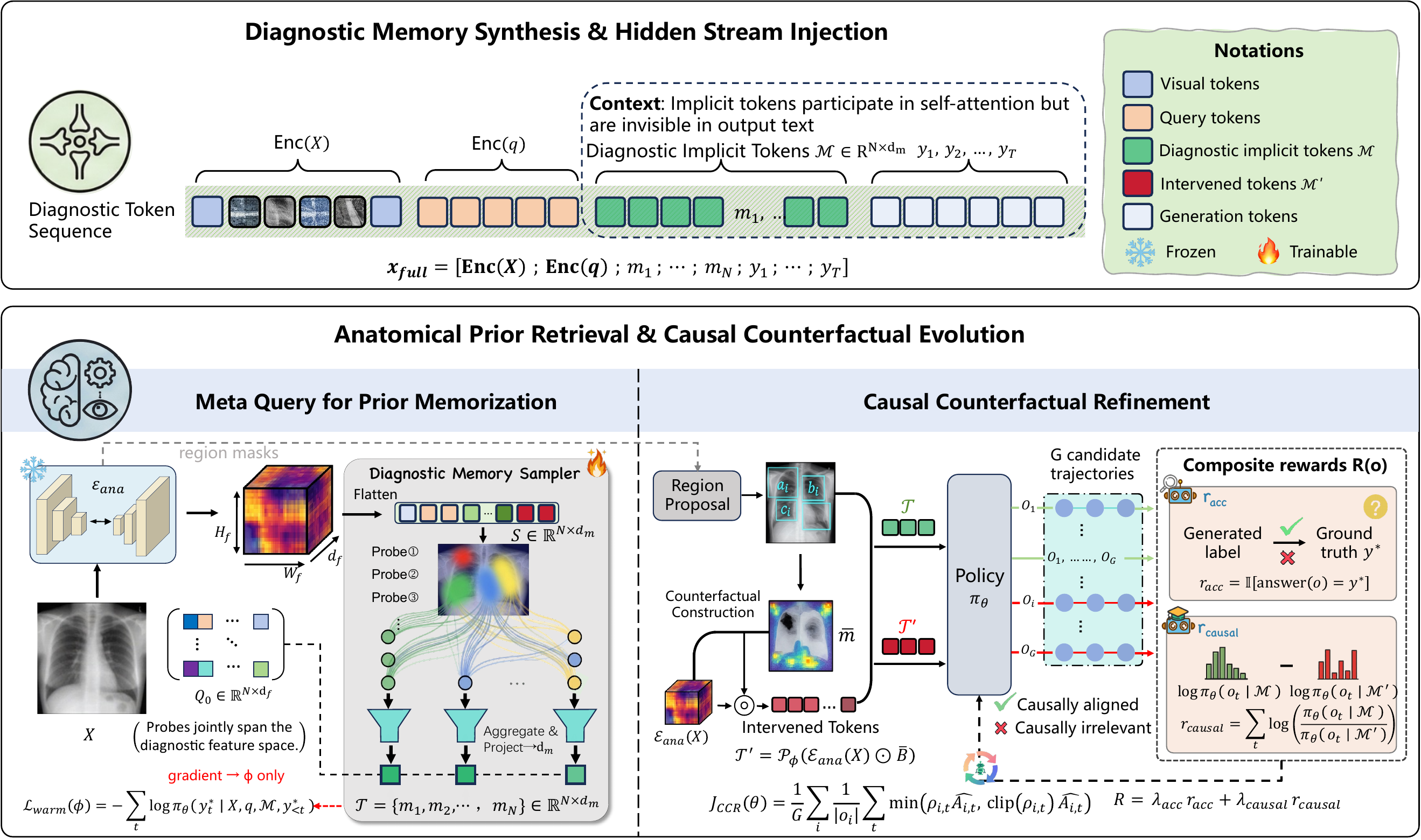}
\caption{Stages I and II of MedSynapse-V. The hook features from an encoder are condensed into diagnostic implicit memory via learnable meta-query probes and injected into the VLM hidden stream. The memory is then refined through RL with composite rewards, ensuring causal alignment between memory and clinical decision logic.}
\label{fig:overview}
\end{figure*}

\vspace{0.5mm}\noindent\textbf{Architecture Overview.}
As illustrated in Figure~\ref{fig:overview}, \method follows a three-stage progressive memory evolution training paradigm. \textit{Stage~\uppercase\expandafter{\romannumeral1}: Meta Query for Prior Memorization} (\S\ref{sec:mqpm}) extracts priors from the frozen anatomical encoder, condenses them into diagnostic implicit memory $\mathcal{M}$ through a learnable memory synthesis module and injects them into the VLM hidden stream, while simultaneously completing semantic alignment warmup. \textit{Stage~\uppercase\expandafter{\romannumeral2}: Causal Counterfactual Refinement} (CCR; \S\ref{sec:ccr}) freezes the synthesis module and performs policy optimization within the $\mathcal{M}$ conditioned latent space based on GRPO, incorporating causal counterfactual rewards for memory refinement. \textit{Stage~\uppercase\expandafter{\romannumeral3}: Intrinsic Memory Transition} (IMT; \S\ref{sec:imt}) writes the refined diagnostic memory into the model's autonomous pathway through privileged autonomous dual branch full vocabulary divergence distillation, completely removing the anatomical encoder at inference. The entire process is represented as a progressive evolution chain of diagnostic memory:
%\begin{equation}
$
\mathbf{F}_{\textit{ana}} \xrightarrow{\;\textbf{Meta Query}\;} \mathcal{M} \xrightarrow{\;\textbf{CCR}\;} \mathcal{M}^{\star} \xrightarrow{\;\textbf{IMT}\;} \mathcal{M}_{\textit{auto}},
$
%\label{eq:evolution}
%\end{equation}
where $\mathbf{F}_{\textit{ana}}$ denotes the anatomical encoder output features, $\mathcal{M}$ is the initial memory after semantic alignment, $\mathcal{M}^{\star}$ is the causally refined memory, and $\mathcal{M}_{\textit{auto}}$ is the intrinsic memory generated by the autonomous module.

% ====================================================================
\subsection{Meta Query for Prior Memorization}
\label{sec:mqpm}

Clinical cognition research has shown that experienced physicians rapidly activate long accumulated anatomical knowledge during image interpretation, forming compressed contextualized expert memory to guide diagnosis~\cite{waite2017interpretive}. Inspired by this finding, this stage models this process as a structured prior retrieval and memorization mechanism: the anatomical encoder extracts spatially aware features, which are then condensed into diagnostic implicit memory through a learnable synthesis module and injected into the VLM hidden states.

\vspace{0.5mm}\noindent\textbf{Structured Prior Elicitation.}
Given an input image $X$, the frozen anatomical encoder $\mathcal{E}_{\textit{ana}}$ outputs spatial features from the final layer:
%\begin{equation}
$
\mathbf{F} = \mathcal{E}_{\textit{ana}}(X) \in \mathbb{R}^{H_f \times W_f \times d_f},
%\label{eq:ana_features}
%\end{equation}
$
where $H_f \times W_f$ is the spatial resolution and $d_f$ is the feature dimensionality. This feature map encodes structured spatial priors learned by the anatomical encoder from large scale medical image segmentation tasks, encompassing multi-granularity information such as lesion boundaries, organ topology, and tissue textures. It is flattened into a sequence $\mathbf{S} = \texttt{flat}(\mathbf{F}) \in \mathbb{R}^{M \times d_f}$, $M = H_f \times W_f$, forming the feature pool for candidate memory.

\vspace{0.5mm}\noindent\textbf{Diagnostic Memory Synthesis.}
%To condense the high dimensional feature pool into compact memory compatible with the VLM hidden state space, we design a lightweight \textit{memory synthesis module} $\mathcal{P}_\phi$. This module maintains $N$ learnable \textit{meta query probes} $\mathbf{Q}_0 \in \mathbb{R}^{N \times d_f}$, where each probe autonomously learns to attend to specific pathological semantic patterns (\textit{e.g.} boundary irregularity, density heterogeneity, or vascular tissue spatial relationships). $\mathcal{P}_\phi$ uses $\mathbf{Q}_0$ as queries and $\mathbf{S}$ as key value pairs, completing selective aggregation and dimensional alignment through cross attention:
To condense the high-dimensional feature pool into compact memory compatible with the VLM hidden state space, we design a lightweight \textit{Diagnostic Memory Sampler} $\mathcal{P}_\phi$ that maintains $N$ learnable \textit{meta-query probes} $\mathbf{Q}_0 \in \mathbb{R}^{N \times d_f}$, each of which learns to attend to specific pathological semantic patterns (\textit{e.g.}, boundary irregularity, density heterogeneity, or vascular-tissue spatial relationships). Using $\mathbf{Q}_0$ as queries and $\mathbf{S}$ as key-value pairs, $\mathcal{P}_\phi$ performs selective aggregation and dimensional alignment:
%\begin{equation}
$
\mathcal{M} = \mathcal{P}_\phi(\mathbf{Q}_0, \mathbf{S}) \in \mathbb{R}^{N \times d_h},
%\label{eq:memory_synthesis}
%\end{equation}
$
where $d_h$ is the VLM hidden state dimensionality. The resulting $N$ diagnostic implicit memory elements $\mathcal{M} = \{m_1, \ldots, m_N\}$ collectively span the feature subspace required for diagnosis. $\mathcal{P}_\phi$ extracts the most diagnostically relevant compact representations from the encoder's high dimensional feature pool according to the task context, bridging them from the anatomical encoder's representation space to the VLM's hidden state space.

\vspace{0.5mm}\noindent\textbf{Diagnostic Memory Injection.}
The diagnostic implicit memory $\mathcal{M}$ is injected into the VLM generation sequence as continuous vectors, positioned after the question encoding and before the answer generation:
$
\mathbf{x}_{\textit{full}} = \bigl[\texttt{Enc}(X);\;\texttt{Enc}(q);\; m_1;\;\\ \ldots;\; m_N;\; y_1;\; \ldots;\; y_T\bigr],
%\label{eq:injection}
%\end{equation}
$
where $\texttt{Enc}(\cdot)$ denotes the encoding operation and $\{y_t\}_{t=1}^T$ are the answer tokens to be generated. Since $\mathcal{M}$ shares the $d_h$ dimensional space with VLM hidden states, the self-attention mechanism enables the generation sequence to dynamically aggregate diagnostic evidence from $\mathcal{M}$, allowing the VLM to modulate its predictive distribution based on latent priors without additional adaptation layers or architectural modifications.

\vspace{0.5mm}\noindent\textbf{Semantic Alignment Warmup.}
Directly injecting outputs from a randomly initialized synthesis module into the VLM would cause semantic mismatch. To address this, a warmup stage is established before formal refinement: the VLM backbone $\theta$ and the anatomical encoder $\mathcal{E}_{\textit{ana}}$ are frozen, and only $\phi$ is optimized with the standard next token prediction loss:
\begin{equation}
\mathcal{L}_{\textit{warm}}(\phi) = -\sum_{t=1}^{T} \log \pi_\theta(y_t^\star \mid X, q, \mathcal{M}, y_{<t}^\star),
\label{eq:warmup}
\end{equation}
where $y^\star$ denotes the reference answer sequence. This stage establishes a well-conditioned initial semantic mapping between the anatomical encoder feature space and the VLM hidden state space, providing a semantically coherent and stable initialization upon which the subsequent RL stage can reliably build.

% ====================================================================
\subsection{Causal Counterfactual Refinement}
\label{sec:ccr}

After warmup, $\mathcal{M}$ possesses basic semantic alignment capability, but directly applying standard SFT has two limitations~\cite{yu2025finemedlm}: binding to fixed reference trajectories restricts out-of-distribution generalization; the model may bypass $\mathcal{M}$ and directly map from $(X, q)$ to answers, causing memory to degenerate into redundant placeholders. In this stage, we perform policy optimization within the $\mathcal{M}$-conditioned latent space, incorporating causal counterfactual intervention to guide memory refinement toward clinical alignment.

\vspace{0.5mm}\noindent\textbf{Conditioned Policy Modeling.}
%First, we freeze $\mathcal{P}_\phi$ and optimizes the VLM through lightweight parameter adapters.
First, we freeze $\mathcal{P}_\phi$ and the VLM backbone, optimizing only lightweight LoRA adapters (collectively denoted $\theta$ below).
 Based on GRPO~\cite{shao2024deepseekmath}, for each sample $(X, q, y^\star)$, $G$ candidate trajectories $\{\mathbf{o}_1, \ldots, \mathbf{o}_G\}$ are sampled. $\mathcal{M}$ is explicitly incorporated into the conditioned context, and the policy gradient indirectly modulates the VLM's utilization pattern of $\mathcal{M}$ through the attention pathway. The policy model optimization objective is:
\begin{equation}
\mathcal{J}_{\textit{CCR}}(\theta) = \frac{1}{G}\sum_{i=1}^{G}\frac{1}{|\mathbf{o}_i|}
\sum_{t=1}^{|\mathbf{o}_i|} 
\min\!\Bigl(\rho_{i,t}\,\hat{A}_i,\;
\texttt{clip}\!\bigl(\rho_{i,t},\,1{-}\varepsilon,\,1{+}\varepsilon\bigr)\hat{A}_i\Bigr),
\label{eq:ccr_obj}
\end{equation}
where $\rho_{i,t}=\frac{\pi_\theta(\mathbf{o}_{i,t}\mid X,q,\mathcal{M},\mathbf{o}_{i,<t})}
{\pi_{\theta_{\textit{old}}}(\mathbf{o}_{i,t}\mid X,q,\mathcal{M},\mathbf{o}_{i,<t})}$ and 
$\hat{A}_i=\frac{R(\mathbf{o}_i)-\mu_G}{\sigma_G+\varepsilon_0}$, with $\mu_G$ and $\sigma_G$ 
denoting the group reward mean and standard deviation, and $\varepsilon_0$ a stability constant. Note that both $\rho_{i,t}$ and $\hat{A}_i$ are conditioned on $\mathcal{M}$, allowing gradients to propagate through the attention pathway linking answer tokens with memory, thereby shaping how the VLM attends to the injected diagnostic cues during generation.

%Note that both $\rho_{i,t}$ and $\hat{A}_i$ are conditioned on $\mathcal{M}$, ensuring that gradient signals propagate through the self-attention pathway that couples answer tokens with memory elements, thereby shaping how the VLM attends to and leverages the injected diagnostic memory during generation.
%where $\rho_{i,t} = {\pi_\theta(\mathbf{o}_{i,t} \mid X, q, \mathcal{M}, \mathbf{o}_{i,<t})}\big/{\pi_{\theta_{\textit{old}}}(\mathbf{o}_{i,t} \mid X, q, \mathcal{M}, \mathbf{o}_{i,<t})}$, $\hat{A}_i = ({R(\mathbf{o}_i) - \mu_G})\big/{(\sigma_G + \varepsilon_0)}$.

\vspace{0.5mm}\noindent\textbf{Interventional Reward Design.}
The composite reward $R(\mathbf{o}) = \lambda_{\textit{acc}} \cdot r_{\textit{acc}} + \lambda_{\textit{causal}} \cdot r_{\textit{causal}}$ consists of two components. The diagnostic accuracy reward is:
\begin{equation}
r_{\textit{acc}}(\mathbf{o}) = \mathbb{I}\bigl[\texttt{answer}(\mathbf{o}) = y^\star\bigr].
\label{eq:r_acc}
\end{equation}
The causal counterfactual reward quantifies the causal contribution of $\mathcal{M}$ through intervention. The frozen anatomical encoder $\mathcal{E}_{\textit{ana}}$ additionally provides highest-confidence region masks for diagnostically relevant areas; zeroing out the corresponding features yields the interventional memory $\mathcal{M}'$:
%\begin{equation}
$
\mathcal{M}' = \mathcal{P}_\phi\!\bigl(\mathcal{E}_{\textit{ana}}(X) \odot \overline{\mathbf{B}}\bigr),
%\label{eq:intervention}
%\end{equation}
$
where $\overline{\mathbf{B}}$ is the inverted binary mask. The causal reward measures the performance discrepancy between the original and interventional conditions:
\begin{equation}
r_{\textit{causal}}(\mathbf{o}) = \sum_{t=1}^{|\mathbf{o}|} \log \frac{\pi_\theta(\mathbf{o}_t \mid X, q, \mathcal{M}, \mathbf{o}_{<t})}{\pi_\theta(\mathbf{o}_t \mid X, q, \mathcal{M}', \mathbf{o}_{<t})}.
\label{eq:r_causal}
\end{equation}
$r_{\textit{causal}} > 0$ indicates that the original memory encodes information with causal contribution to diagnosis; $r_{\textit{causal}} \leq 0$ indicates that the corresponding prior is causally irrelevant to the current decision. This design follows the principle of interventional effect estimation in causal inference, ensuring that the memory retained after refinement is causally consistent with clinical decision logic.

\begin{figure*}[t]
\centering
\includegraphics[width=1\textwidth]{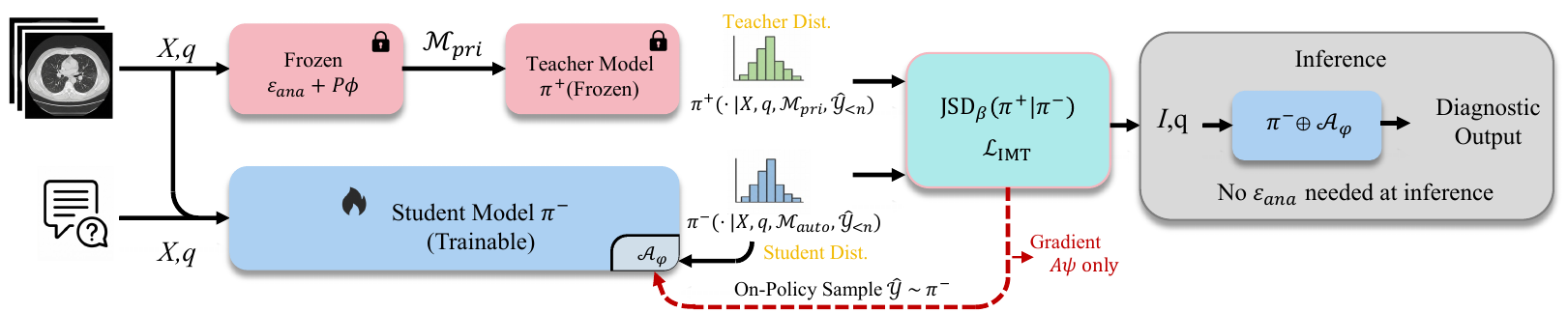}
\caption{Intrinsic Memory Transition (IMT) is achieved via Jensen–Shannon divergence alignment between the teacher ($\pi^{+}$, conditioned on encoder-derived $\mathcal{M}_{pri}$) and student ($\pi^{-}$, conditioned on $\mathcal{M}_{auto}$) branches. Gradients propagate solely to $\mathcal{A}_{\psi}$, enabling complete removal of the anatomical encoder at inference with negligible overhead.}
\label{fig:IMT}
\end{figure*}

% ====================================================================
\subsection{Intrinsic Memory Transition}
\label{sec:imt}

After the causal RL, the model can generate high quality diagnostic outputs under the guidance of externally derived $\mathcal{M}$, but the persistent dependence on the encoder at inference introduces additional computational overhead. As shown in Fig.~\ref{fig:IMT}, IMT reformulates this problem as learning to autonomously generate equivalent latent space embeddings under the condition of removing the auxiliary encoder, employing a privileged autonomous dual-branch paradigm to accomplish memory transition from extrinsic to intrinsic.

\vspace{0.5mm}\noindent\textbf{Privileged Branch and Autonomous Branch.}
The \text{teacher branch} (privileged) retains the complete pipeline with the anatomical encoder, generating reference memory $\mathcal{M}_{\textit{pri}} = \mathcal{P}_\phi(\mathcal{E}_{\textit{ana}}(X)) \in \mathbb{R}^{N \times d_h}$. The \text{student branch} (autonomous) removes the encoder and predicts equivalent memory solely through a lightweight Autonomous Memory Module $\mathcal{A}_\psi$ mounted on the VLM, using the VLM's own visual encoding features:
\begin{equation}
\mathcal{M}_{\textit{auto}} = \mathcal{A}_\psi\!\bigl(\texttt{Enc}_{\textit{VLM}}(X, q)\bigr) \in \mathbb{R}^{N \times d_h}.
\label{eq:m_auto}
\end{equation}
Both branches share the VLM backbone parameters $\theta$, ensuring that alignment is completed within the same function space.

\vspace{0.5mm}\noindent\textbf{Training Objective: Unified Full Vocabulary Divergence.}
The core signal of IMT is the full vocabulary divergence objective covering all generation positions. Given a trajectory $\hat{y} \sim \pi^{-}(\cdot \mid X, q, \mathcal{M}_{\textit{auto}})$ sampled from the student branch, the trajectory averaged per position divergence is defined as:
\begin{equation}
\mathcal{L}_{\textit{IMT}}(\psi)
= \mathbb{E}_{(X,q,y^\star)} \!\left[
\mathbb{E}_{\hat{y}\,\sim\,\pi^{-}} \!\left[
\frac{1}{|\hat{y}|}
\sum_{n=1}^{|\hat{y}|}
D\!\left(
\pi^{+}\!\left(\cdot \mid \hat{y}_{<n}\right)
\;\middle\|\;
\pi^{-}\!\left(\cdot \mid \hat{y}_{<n}\right)
\right)
\right]
\right],
\label{eq:ari_loss}
\end{equation}
where {\small $\pi^{+}(\cdot \mid \hat{y}_{<n}) \triangleq 
\pi_\theta(\cdot \mid X, q, \mathcal{M}_{\textit{pri}}, \hat{y}_{<n})$} 
and {\small $\pi^{-}(\cdot \mid \hat{y}_{<n}) \triangleq 
\pi_\theta(\cdot \mid X, q, \mathcal{M}_{\textit{auto}}, \hat{y}_{<n})$} 
are the full vocabulary next-token distributions conditioned on privileged and autonomous memory, respectively. The divergence $D$ adopts the generalized Jensen--Shannon divergence $\operatorname{JSD}_\beta$~\cite{lin2002divergence}:
\begin{equation}
\operatorname{JSD}_\beta\!\left(\pi^{+} \| \pi^{-}\right)
= \beta\, D_{\textit{KL}}\!\left(\pi^{+} \| \bar{m}\right)
+ (1{-}\beta)\, D_{\textit{KL}}\!\left(\pi^{-} \| \bar{m}\right),
\end{equation}
where $\bar{m} = \beta\,\pi^{+} + (1{-}\beta)\,\pi^{-}$. The teacher branch leverages privileged memory to expose the complete next token probability landscape to the student, driving $\mathcal{A}_\psi$ to learn to generate latent space embeddings behaviorally equivalent to privileged memory. Gradients backpropagate only through the student branch to $\mathcal{A}_\psi$, while the teacher branch serves as a fixed distributional target.

\textbf{At inference}, $\mathcal{A}_\psi$ directly generates $\mathcal{M}_{\textit{auto}}$ from VLM visual encoding features and injects them into the hidden stream. The entire Meta Query pipeline and the anatomical encoder are completely removed, rendering the computational overhead of \method nearly identical to that of a standard VLM.

\section{Experiments}
\label{sec:experiments}

\subsection{Experimental Setup}

\paragraph{Datasets}
\textbf{Training data:}
We conduct comprehensive experiments on seven medical multimodal benchmarks spanning diverse task types and difficulty levels.
\textbf{Stage~\uppercase\expandafter{\romannumeral1}} (MQPM warmup) uses 50K image--text pairs from PubMedVision~\cite{chen2024towards} covering radiology and pathology.
\textbf{Stage~\uppercase\expandafter{\romannumeral2}} (CCR) constructs a mixed-modality RL set: 3K closed-ended VQA samples from OmniMedVQA~\cite{hu2024omnimedvqa} training split (8 modalities: CT, MRI, X-ray, dermoscopy, fundus, OCT, pathology, ultrasound) plus 1K open-ended samples from SLAKE~\cite{liu2021slake} and PathVQA~\cite{he2020pathvqa} training sets, totaling $\sim$4K samples. Region masks for $r_{\textit{causal}}$ are provided by MedSAM3~\cite{liu2025medsam3}.
\textbf{Stage~\uppercase\expandafter{\romannumeral3}} (IMT) reuses the Stage~\uppercase\expandafter{\romannumeral2} data.
\textbf{Evaluation benchmarks:}
\textbf{(i)} Closed-ended VQA: VQA-RAD~\cite{lau2018dataset}, SLAKE~\cite{liu2021slake}, PathVQA~\cite{he2020pathvqa}, PMC-VQA~\cite{zhang2023pmc};
\textbf{(ii)} Clinical reasoning: MMMU Health \& Medicine~\cite{yue2024mmmu} (denoted MMMU*);
\textbf{(iii)} Expert-level reasoning: MedXpertQA-MM~\cite{zuo2025medxpertqa} (Total score);
\textbf{(iv)} Multi-granularity: GMAI-MMBench~\cite{ye2024gmai}.

\paragraph{Baselines}
We compare against four categories of methods:
\textbf{(1)} \textit{General VLMs}: Qwen3-VL-8B~\cite{bai2025qwen3} (our base model), InternVL3-8B~\cite{zhu2025internvl3};
\textbf{(2)} \textit{Medical-specific VLMs}: RadFM~\cite{wu2025towards}, LLaVA-Med~\cite{li2023llava}, GMAI-VL~\cite{li2024gmai}, HuatuoGPT-Vision~\cite{chen2024towards}, BiMediX2-8B~\cite{mullappilly2024bimedix2}, MedMO-8B~\cite{deria2026medmo};
\textbf{(3)} \textit{RL-enhanced medical reasoning}: MedVLM-R1-2B~\cite{pan2025medvlm}, Med-R1-3B~\cite{lai2026med}, MediX-R1-8B~\cite{mullappilly2026medix}, MMedExpert-R1-7B~\cite{ding2026mmedexpert};
\textbf{(4)} \textit{Latent-space reasoning}: Coconut$^\dagger$~\cite{hao2024training}, MCOUT-Multi$^\dagger$~\cite{pham2025multimodal}, IVT-LR$^\dagger$~\cite{chen2025reasoning} ($^\dagger$: adapted with identical Qwen3-VL-8B backbone and training data).
We additionally report \method-4B on the Qwen3-VL-4B backbone to assess scalability.

\paragraph{Implementation Details}
Our framework builds upon Qwen3-VL-8B-Instruct~\cite{bai2025qwen3}. The frozen anatomical encoder $\mathcal{E}_{\textit{ana}}$ employs MedSAM3~\cite{liu2025medsam3} pre-trained on large-scale multi-organ segmentation datasets.
\text{Stage~\uppercase\expandafter{\romannumeral1}} freezes both VLM and $\mathcal{E}_{\textit{ana}}$, training only the Diagnostic Memory Sampler $\mathcal{P}_\phi$ with lr $= 2\times10^{-4}$ for 3 epochs. The diagnostic probe count is $N{=}16$; $\mathcal{P}_\phi$ is a 2-layer cross-attention Transformer with output dimension $d_m{=}4096$ (matching Qwen3-VL-8B). Images are processed at native dynamic resolution following Qwen3-VL's default configuration.
\text{Stage~\uppercase\expandafter{\romannumeral2}} freezes $\mathcal{P}_\phi$ and adapts VLM via LoRA~\cite{hu2022lora} (rank$=$64, applied to all attention layers). GRPO generates $G{=}4$ candidate trajectories per sample, with clipping coefficient $\varepsilon{=}0.2$, reward weights $\lambda_{\textit{acc}}{=}1.0$ and $\lambda_{\textit{causal}}{=}0.5$, training for 200 steps with a rollout batch size of 32. Max generation length is 1024 tokens.
%\text{Stage~\uppercase\expandafter{\romannumeral3}} introduces the Autonomous Memory Module $\mathcal{A}_\psi$ (2-layer MLP + LayerNorm, input from VLM's visual encoder features), with JSD coefficient $\beta{=}0.5$, lr $= 1\times10^{-4}$, 3 epochs. We optimize all stages with AdamW on 4$\times$NVIDIA A100-80GB, bf16 precision with FlashAttention-2.
\text{Stage~\uppercase\expandafter{\romannumeral3}} introduces the Autonomous Memory Module $\mathcal{A}_\psi$ (2-layer MLP + LayerNorm, input from VLM's visual encoder features), with JSD coefficient $\beta{=}0.5$, lr $= 1\times10^{-4}$, 3 epochs. For each sample we draw one on-policy trajectory $\hat{y}\sim\pi^{-}$ per gradient step; the Stage~\uppercase\expandafter{\romannumeral2} data is reused with identical preprocessing.
%\paragraph{Evaluation Metrics}
Closed-ended VQA tasks report overall accuracy (\%). For GMAI-MMBench and MedXpertQA-MM, we follow their respective official evaluation protocols. Inference efficiency is measured quantitatively as ms/sample and peak GPU memory (GB). More details are provided in the supplementary material.

\begin{table*}[t]
\caption{Comprehensive comparison on seven medical benchmarks. Base model: Qwen3-VL-8B (unless otherwise noted). MMMU* = Health \& Medicine track. $^\dagger$: general latent-space methods adapted to medical VQA with identical backbone and training data. $\Delta$: absolute gap (pp) to \method (w/ $\mathcal{E}_{\textit{ana}}$) average. All results are averaged over five independent runs. \textbf{Bold}: best; \underline{underline}: second best.}
\label{tab:main_results}
\centering
\large
\renewcommand{\arraystretch}{1}
\adjustbox{width=\textwidth}{
\begin{tabular}{l|C{1.8cm}|C{2.4cm}C{2cm}C{2cm}C{2.4cm}C{2cm}|C{2.3cm}|C{2cm}|C{1.9cm}|C{1.8cm}}
\toprule[1.1pt]
\textbf{Method} & \textbf{Size} & \textbf{VQA-RAD} & \textbf{SLAKE} & \textbf{PathVQA} & \textbf{PMC-VQA} & \textbf{MMMU*} & \textbf{MedXpert} & \textbf{GMAI} & \textbf{Average} & $\mathbf{\Delta}$ \textbf{(pp)} \\
\midrule[0.9pt]
\rowcolor{myyellow}
\multicolumn{11}{l}{\textit{General Vision-Language Models}} \\
Qwen3-VL-8B~\cite{bai2025qwen3}         & 8B  & 58.6 & 66.2 & 55.4 & 42.5 & 48.3 & 22.1 & 47.2 & 48.6 & -12.8 \\
InternVL3-8B~\cite{zhu2025internvl3}     & 8B  & 57.3 & 64.8 & 50.6 & 41.2 & 50.1 & 21.5 & 49.6 & 47.9 & -13.5 \\
\midrule[0.9pt]
\rowcolor{myyellow}
\multicolumn{11}{l}{\textit{Medical-Specific VLMs}} \\
RadFM~\cite{wu2025towards}               & 14B & 50.6 & 34.6 & 38.7 & 25.9 & 27.0 & 17.3 & 28.5 & 31.8 & -29.6 \\
LLaVA-Med~\cite{li2023llava}             & 7B  & 51.4 & 48.6 & 56.8 & 30.1 & 36.9 & 19.7 & 31.2 & 39.2 & -22.2 \\
GMAI-VL~\cite{li2024gmai}                & 7B  & 64.6 & 71.9 & 47.2 & 52.3 & 51.2 & 23.8 & 45.2 & 50.9 & -10.5 \\
HuatuoGPT-V~\cite{chen2024towards}       & 7B  & 63.8 & 74.5 & 59.9 & 53.4 & 49.1 & 22.7 & 51.3 & 53.5 & -7.9 \\
BiMediX2~\cite{mullappilly2024bimedix2}  & 8B  & 62.4 & 68.3 & 52.7 & 42.8 & 48.6 & 22.2 & 34.6 & 47.4 & -14.0 \\
MedMO-8B~\cite{deria2026medmo}         & 8B  & 59.3 & 66.8 & 48.5 & 36.0 & 46.2 & 20.8 & 38.2 & 45.1 & -16.3 \\
\midrule[0.9pt]
\rowcolor{myyellow}
\multicolumn{11}{l}{\textit{RL-Enhanced Medical CoT / Latent Reasoning}} \\
MedVLM-R1~\cite{pan2025medvlm}           & 2B  & 58.6 & 63.2 & 42.5 & 35.8 & 31.2 & 16.8 & 33.5 & 40.2 & -21.2 \\
Med-R1~\cite{lai2026med}                 & 3B  & 53.2 & 52.8 & 44.1 & 38.5 & 30.4 & 18.5 & 35.2 & 38.9 & -22.5 \\
MediX-R1~\cite{mullappilly2026medix}      & 8B  & 56.4 & 65.8 & 44.2 & 56.2 & 53.5 & 24.9 & 48.2 & 49.9 & -11.5 \\
MMedExpert-R1~\cite{ding2026mmedexpert}   & 7B  & 65.2 & 72.8 & 58.1 & 56.8 & 57.3 & \underline{27.5} & \underline{52.1} & 55.7 & -5.7 \\
Coconut$^\dagger$~\cite{hao2024training}       & 8B & 55.8 & 63.4 & 50.1 & 41.6 & 43.2 & 19.8 & 37.6 & 44.5 & -16.9 \\
MCOUT-Multi$^\dagger$~\cite{pham2025multimodal}& 8B & 59.2 & 67.4 & 53.8 & 44.8 & 47.5 & 22.0 & 40.9 & 47.9 & -13.5 \\
IVT-LR$^\dagger$~\cite{chen2025reasoning}     & 8B & 62.3 & 70.1 & 56.2 & 47.8 & 50.4 & 23.5 & 43.1 & 50.5 & -10.9 \\
\midrule[0.9pt]
\rowcolor[RGB]{231, 248, 254} \method-4B (w/ $\mathcal{E}_{\textit{ana}}$) & 4B & 67.8 & 75.2 & 60.5 & 53.1 & 54.5 & 24.2 & 47.3 & 54.7 & -6.7 \\
\rowcolor[RGB]{231, 248, 254} \method-4B (IMT)  & 4B & 66.0 & 73.1 & 58.6 & 51.2 & 52.4 & 22.6 & 45.0 & 52.7 & -8.7 \\
\rowcolor[RGB]{231, 248, 254} \method (w/ $\mathcal{E}_{\textit{ana}}$) & 8B & \textbf{75.6} & \textbf{81.4} & \textbf{66.2} & \textbf{59.8} & \textbf{62.7} & \textbf{29.4} & \textbf{54.8} & \textbf{61.4} & -- \\
\rowcolor[RGB]{231, 248, 254} \method (IMT) & 8B & \underline{74.2} & \underline{79.8} & \underline{64.8} & \underline{58.5} & \underline{61.4} & 26.8 & 51.6 & \underline{59.6} & -1.8 \\
\bottomrule[1.1pt]
\end{tabular}
}
\end{table*}

%Our framework builds upon Qwen3-VL-8B-Instruct~\cite{bai2025qwen3} with the frozen anatomical encoder $\mathcal{E}_{\textit{ana}}$ from MedSAM3~\cite{liu2025medsam3}.
%\textbf{Stage~\uppercase\expandafter{\romannumeral1}}: freezes VLM and $\mathcal{E}_{\textit{ana}}$, trains only $\mathcal{P}_\phi$ (lr $= 2\times10^{-4}$, 3 epochs, $N{=}16$ probes, 2-layer cross-attention, $d_m{=}4096$).
%\textbf{Stage~\uppercase\expandafter{\romannumeral2}}: freezes $\mathcal{P}_\phi$, adapts VLM via LoRA~\cite{hu2022lora} (rank 64, all attention layers); GRPO with $G{=}4$ trajectories, $\varepsilon{=}0.2$, $\lambda_{\textit{acc}}{=}1.0$, $\lambda_{\textit{causal}}{=}0.5$, 200 steps, rollout batch 32, max 1024 tokens.
%\textbf{Stage~\uppercase\expandafter{\romannumeral3}}: introduces $\mathcal{A}_\psi$ (2-layer MLP + LayerNorm), JSD $\beta{=}0.5$, lr $= 1\times10^{-4}$, 3 epochs.
%For \method-4B we use Qwen3-VL-4B-Instruct ($d_m{=}2560$, LoRA rank 32, $N{=}12$); all other hyperparameters remain identical.
%All experiments are conducted five times with different random seeds; we report the mean. We use AdamW on 4$\times$A100-80GB with bf16 and FlashAttention-2. Further details are in the supplementary material.

\subsection{Main Results}

As shown in Table~\ref{tab:main_results}, \method (w/ $\mathcal{E}_{\textit{ana}}$) achieves the highest average of $\mathbf{61.4\%}$, and the encoder-free \method (IMT) retains $\mathbf{59.6\%}$, surpassing all baselines. Compared to the strongest RL baseline MMedExpert-R1 (55.7\%), \method (IMT) leads by $\mathbf{+3.9}$ pp without any auxiliary module at inference, with \textit{{the largest margins on visual-grounding benchmarks}} (VQA-RAD $\mathbf{+9.0}$, SLAKE $\mathbf{+7.0}$, PathVQA $\mathbf{+6.7}$), where discrete CoT tokens are prone to attenuating early visual evidence across long reasoning chains. On GMAI-MMBench spanning 38 modalities, \method scores 54.8\%, confirming that the anatomical priors generalize beyond the training distribution.

\vspace{1mm}
\noindent\textbf{RL baselines reveal a specialization dilemma.} MediX-R1 benefits from multilingual pretraining and leads on PMC-VQA (56.2\%), yet this breadth dilutes radiology-specific precision (VQA-RAD: 56.4\%); MMedExpert-R1 achieves the most balanced profile by leveraging guideline-based reward. Small-scale models (MedVLM-R1 2B, Med-R1 3B) collapse on out-of-domain tasks (MedXpert below 19\%), confirming that parameter capacity sets a hard ceiling RL alone cannot raise. n contrast, \method sidesteps this dilemma by injecting latent priors that benefit all task types uniformly, achieving the top performance on every benchmark without task-specific tuning.

\vspace{1mm}
\noindent\textbf{Latent methods require domain priors.}
Among adapted latent baselines, the hierarchy Coconut (44.5\%) $<$ MCOUT-Multi (47.9\%) $<$ IVT-LR (50.5\%) tracks optimization sophistication, yet even IVT-LR barely exceeds zero-shot Qwen3-VL-8B (48.6\%). This inversion reveals that \textit{{latent compression without clinical grounding encodes statistical shortcuts}} rather than diagnostic logic; the 10.9 pp gap to \method confirms that prior injection and causal calibration are prerequisites for effective latent reasoning in medicine.

\vspace{1mm}
\noindent\textbf{Scaling efficiency.}
\method-4B (w/ $\mathcal{E}_{\textit{ana}}$) reaches 54.7\% with roughly half the parameters of 7B baselines; after encoder removal the IMT variant still achieves 52.7\% at 85\,ms and 10.8\,GB, surpassing MediX-R1-8B (49.9\%). This efficiency stems from a structural advantage: \textit{diagnostic expertise is distilled into 16 compact memory vectors consumed in a single forward pass}, rather than spread across 150+ verbose reasoning tokens.

\begin{figure}[t]
  \centering
  \includegraphics[width=1\textwidth]{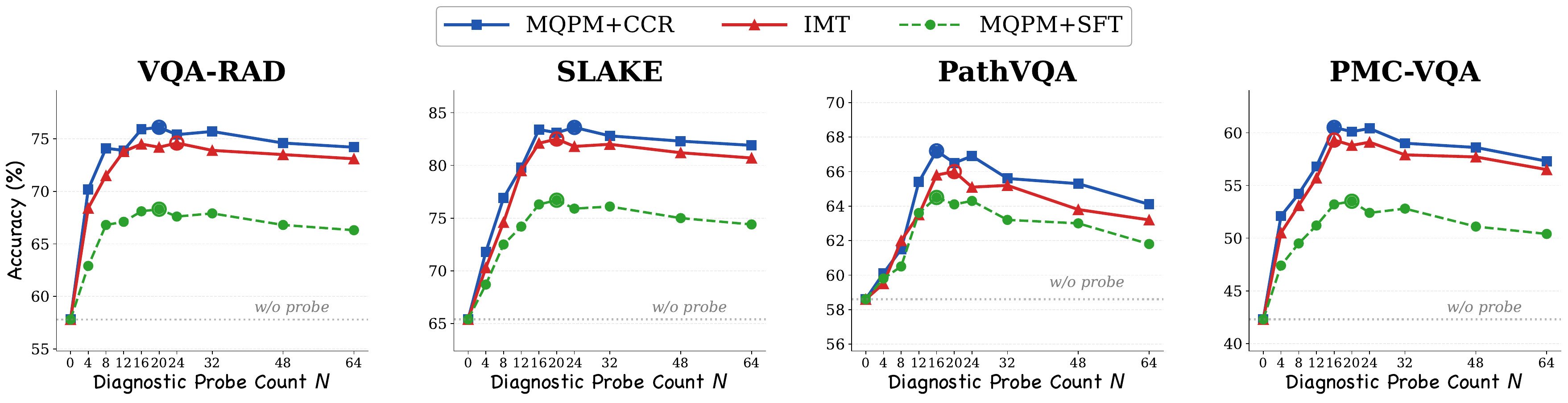}
\caption{\textbf{Effect of diagnostic probe count $N$.}
Performance peaks around $N{=}16$ across benchmarks; further
increasing $N$ dilutes diagnostically relevant signals.}
%\caption{\textbf{Effect of diagnostic probe count $N$. Nearly all configurations peak around $N{=}16$; further increasing $N$ introduces redundant memory that degrades accuracy. MQPM = Meta Query for Prior Memorization.}
\label{fig:latent_size}
\end{figure}

\begin{table*}[t]
\caption{Ablation study on the 8B backbone. All variants undergo the full three-stage pipeline including IMT (\S\ref{sec:imt}); results reflect inference \emph{without} the anatomical encoder unless stated otherwise.  MQPM: Meta Query for Prior Memorization (\S\ref{sec:mqpm}); CCR: Causal Counterfactual Refinement (\S\ref{sec:ccr}); $\mathcal{M}$: diagnostic implicit memory. Best per group in \textbf{bold}.}
\label{tab:ablation}
\centering
\renewcommand{\arraystretch}{1}
\adjustbox{width=\textwidth}{
\begin{tabular}{l|ccccc|cc|C{1.5cm}}
\toprule[1pt]
\textbf{Configuration} & \textbf{VQA-RAD} & \textbf{SLAKE} & \textbf{PathVQA} & \textbf{PMC-VQA} & \textbf{MMMU*} & \textbf{ms/token}$\downarrow$ & \textbf{Mem (GB)}$\downarrow$ & \textbf{Avg.} \\
\midrule[0.85pt]
Qwen3-VL-8B (zero-shot, no $\mathcal{M}$) & 58.6 & 66.2 & 55.4 & 42.5 & 48.3 & 126 & 16.2 & 54.2 \\
\midrule
\rowcolor{myyellow} \multicolumn{9}{l}{\textit{(a) Progressive Training Stages}} \\
MQPM $\to$ IMT (Stage~I only, no RL) & 59.5 & 67.4 & 57.1 & 45.6 & 47.2 & 104 & 16.5 & 55.4 \\
MQPM $\to$ SFT $\to$ IMT (no RL) & 63.2 & 71.2 & 60.4 & 49.8 & 51.3 & 104 & 16.5 & 59.2 \\
Direct RL $\to$ IMT (skip MQPM) & 57.4 & 64.3 & 54.8 & 43.1 & 44.7 & 105 & 16.5 & 52.9 \\
MQPM $\to$ CCR $\to$ IMT (full) & \textbf{74.2} & \textbf{79.8} & \textbf{64.8} & \textbf{58.5} & \textbf{61.4} & 102 & 16.5 & \textbf{67.7} \\
\midrule
\rowcolor{myyellow} \multicolumn{9}{l}{\textit{(b) Reward Design}} \\
$r_{\textit{acc}}$ only & 70.1 & 75.8 & 61.2 & 54.3 & 56.6 & 102 & 16.5 & 63.6 \\
$r_{\textit{acc}} + r_{\textit{causal}}$ (full) & \textbf{74.2} & \textbf{79.8} & \textbf{64.8} & \textbf{58.5} & \textbf{61.4} & 102 & 16.5 & \textbf{67.7} \\
\midrule
\rowcolor{myyellow} \multicolumn{9}{l}{\textit{(c) Encoder Retention vs.\ Removal}} \\
w/ $\mathcal{E}_{\textit{ana}}$ (no privileged branch) & 75.6 & 81.4 & 66.2 & 59.8 & 62.7 & 168 & 22.8 & 69.1 \\
w/o $\mathcal{E}_{\textit{ana}}$ (IMT, default) & 74.2 & 79.8 & 64.8 & 58.5 & 61.4 & \textbf{102} & \textbf{16.5} & 67.7 \\
\midrule
\rowcolor{myyellow} \multicolumn{9}{l}{\textit{(d) Encoder Choice}} \\
SAM-Med2D~\cite{cheng2023sam} & 69.0 & 75.4 & 60.8 & 54.1 & 57.0 & 102 & 16.5 & 63.3 \\
MedSAM3~\cite{liu2025medsam3} (default) & \textbf{74.2} & \textbf{79.8} & \textbf{64.8} & \textbf{58.5} & \textbf{61.4} & 102 & 16.5 & \textbf{67.7} \\
Random init.\ $\to$ IMT & 56.4 & 64.0 & 55.1 & 41.2 & 43.3 & 102 & 16.5 & 52.0 \\
\bottomrule[1pt]
\end{tabular}
}
\end{table*}
% ---- Latent Size Ablation ----

\subsection{Ablation Study}
\label{sec:ablation}

Table~\ref{tab:ablation} reports comprehensive ablation study results.
Specifically, \textbf{\textit{(i)} Progressive training stages.}
\textbf{\textit{MQPM warmup is indispensable}}: skipping it collapses Avg to 52.9, barely above zero-shot (54.2), because randomly initialized memory destabilizes early RL training. Replacing CCR with SFT reaches 59.2 but lags by 8.5 pp due to limited out-of-distribution generalization. The full pipeline (Avg $\mathbf{67.7}$) confirms non-redundant contributions: MQPM grounds semantics, CCR refines via exploration, IMT compresses into an autonomous pathway.
\textbf{\textit{(ii)} Reward design.}
\textit{\textbf{$r_{\textit{causal}}$ is the dominant reward component}} ($\mathbf{+4.1}$ pp, $63.6 \to 67.7$). Without causal pressure the model bypasses $\mathcal{M}$ via direct shortcuts, treating memory as inert padding; the counterfactual intervention penalizes trajectories insensitive to diagnostic regions. The effect concentrates on radiology benchmarks and persists after IMT, indicating stronger memory utilization transfers more faithfully through distillation.
\textbf{\textit{(iii)} Encoder retention vs.\ removal.}
\textit{\textbf{IMT achieves near-lossless removal}}: only 1.4 pp degradation ($69.1 \to 67.7$) while latency drops 39\% and memory decreases 6.3\,GB. The gap is not uniform: core VQA metrics degrade minimally, whereas MedXpert and GMAI suffer more, suggesting \textit{\textbf{complex reasoning depends more on encoder-derived priors}} than closed-ended recognition.
\textbf{\textit{(iv)} Anatomical encoder choice.}
MedSAM3 outperforms SAM-Med2D by 4.4 pp ($67.7$ vs.\ $63.3$), reflecting richer spatial representations from multi-organ segmentation pretraining. Random initialization yields only 52.0, confirming that gains originate from \emph{what} the encoder knows, rather than \emph{how} memory is aggregated.
\textbf{\textit{(v)} Probe count $N$.}
As shown in Fig.~\ref{fig:latent_size}, $N{=}16$ balances expressiveness against redundancy. The CCR to SFT gap widens with $N$ ($3.5$ pp at $N{=}4$ vs.\ $7.2$ pp at $N{=}16$), revealing that \textit{\textbf{larger memory pools amplify bypass shortcuts}} and therefore benefit disproportionately from causal refinement.

\begin{figure*}[t]
\centering
\includegraphics[width=1\textwidth]{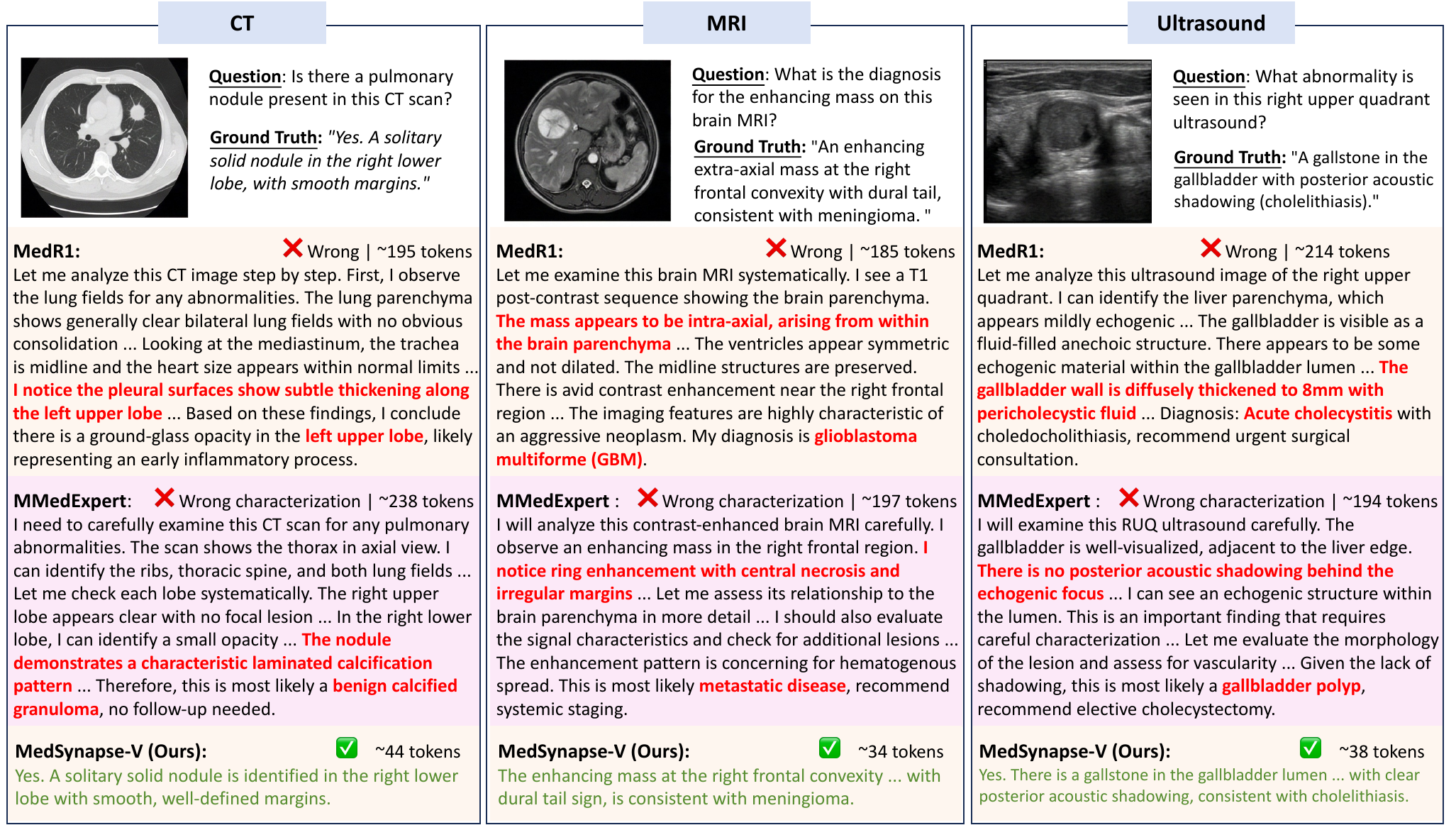}
\caption{Qualitative comparison across CT, MRI, and Ultrasound cases. \method produces concise, correct diagnoses, while Med-R1 and MMedExpert-R1 generate verbose CoT  with hallucinated findings (\textcolor{red}{red}) leading to misdiagnoses.}
\label{fig:attention_case}
\end{figure*}

\subsection{In-Depth Case Analysis}
%As illustrated in Fig.~\ref{fig:attention_case}, we compare \method with two strong RL-enhanced baselines, Med-R1~\cite{lai2026med} and MMedExpert-R1~\cite{ding2026mmedexpert}, across three imaging modalities. Both baselines produce verbose CoT reasoning ($\sim$185--238 tokens) yet arrive at incorrect diagnoses due to hallucinated observations propagating through the reasoning chain. In the CT case, Med-R1 fabricates pleural thickening in the left upper lobe, misdirecting its conclusion to a ground-glass opacity in the wrong location, while MMedExpert-R1 \textit{\textbf{hallucinates a laminated calcification pattern}} and mischaracterizes the nodule as a benign granuloma. In the MRI case, Med-R1 misidentifies the extra-axial mass as intra-axial and concludes glioblastoma, whereas MMedExpert-R1 fabricates ring enhancement with central necrosis and diagnoses metastatic disease, both missing the classic meningioma presentation. In the ultrasound case, Med-R1 hallucinates gallbladder wall thickening with pericholecystic fluid to over-diagnose acute cholecystitis, while MMedExpert-R1 denies the visible posterior acoustic shadowing and misdiagnoses a gallbladder polyp. In contrast, \method generates concise, correct answers ($\sim$34--44 tokens) without explicit CoT, demonstrating that \textbf{\textit{diagnostic implicit memory provides sufficient latent guidance while avoiding the hallucination cascades}} inherent in token-level CoT.
As illustrated in Fig.~\ref{fig:attention_case}, we compare \method with Med-R1~\cite{lai2026med} and MMedExpert-R1~\cite{ding2026mmedexpert} across three distinct imaging modalities. Both baselines produce verbose CoT reasoning ($\sim$185--238 tokens) yet arrive at incorrect diagnoses due to hallucinated observations erroneously propagating through the chain. In the CT case, Med-R1 fabricates pleural thickening in the left upper lobe, while MMedExpert-R1 \textit{\textbf{hallucinates a laminated calcification pattern}} and mischaracterizes the nodule as a benign granuloma. In the MRI case, Med-R1 misidentifies the extra-axial mass as intra-axial and concludes glioblastoma, whereas MMedExpert-R1 fabricates ring enhancement with central necrosis, both missing the classic meningioma presentation. In the ultrasound case, Med-R1 hallucinates gallbladder wall thickening to over-diagnose acute cholecystitis, while MMedExpert-R1 denies posterior acoustic shadowing and misdiagnoses a gallbladder polyp. In contrast, \method generates concise, correct answers ($\sim$34--44 tokens) without explicit CoT, demonstrating that \textbf{\textit{diagnostic implicit memory provides sufficient latent guidance while avoiding the hallucination cascades}} inherent in token-level CoT.

\subsection{Efficiency, RL Dynamics, and Latent Space}
\label{sec:analysis}

%\noindent\textbf{Performance and efficiency.}
%
%\begin{wrapfigure}{r}{0.4\linewidth}
%    \centering
%    \vspace{-10pt}
%    \includegraphics[width=\linewidth]{pic/efficiency_scatter.pdf}
%    \caption{Performance and efficiency trade-off.}
%    \label{fig:efficiency}
%    \vspace{-10pt}
%\end{wrapfigure}
%
%\begin{wrapfigure}{r}{0.4\linewidth}
%    \centering
%    \vspace{-20pt}
%    \includegraphics[width=\linewidth]{pic/training_reward.pdf}
%    \caption{CCR reward curves.}
%    \label{fig:training}
%    \vspace{-10pt}
%\end{wrapfigure}
\begin{wrapfigure}{r}{0.41\linewidth}
    \centering
    \vspace{-25pt}
    % 第一张图
    \begin{minipage}{\linewidth}
        \centering
        \includegraphics[width=\linewidth]{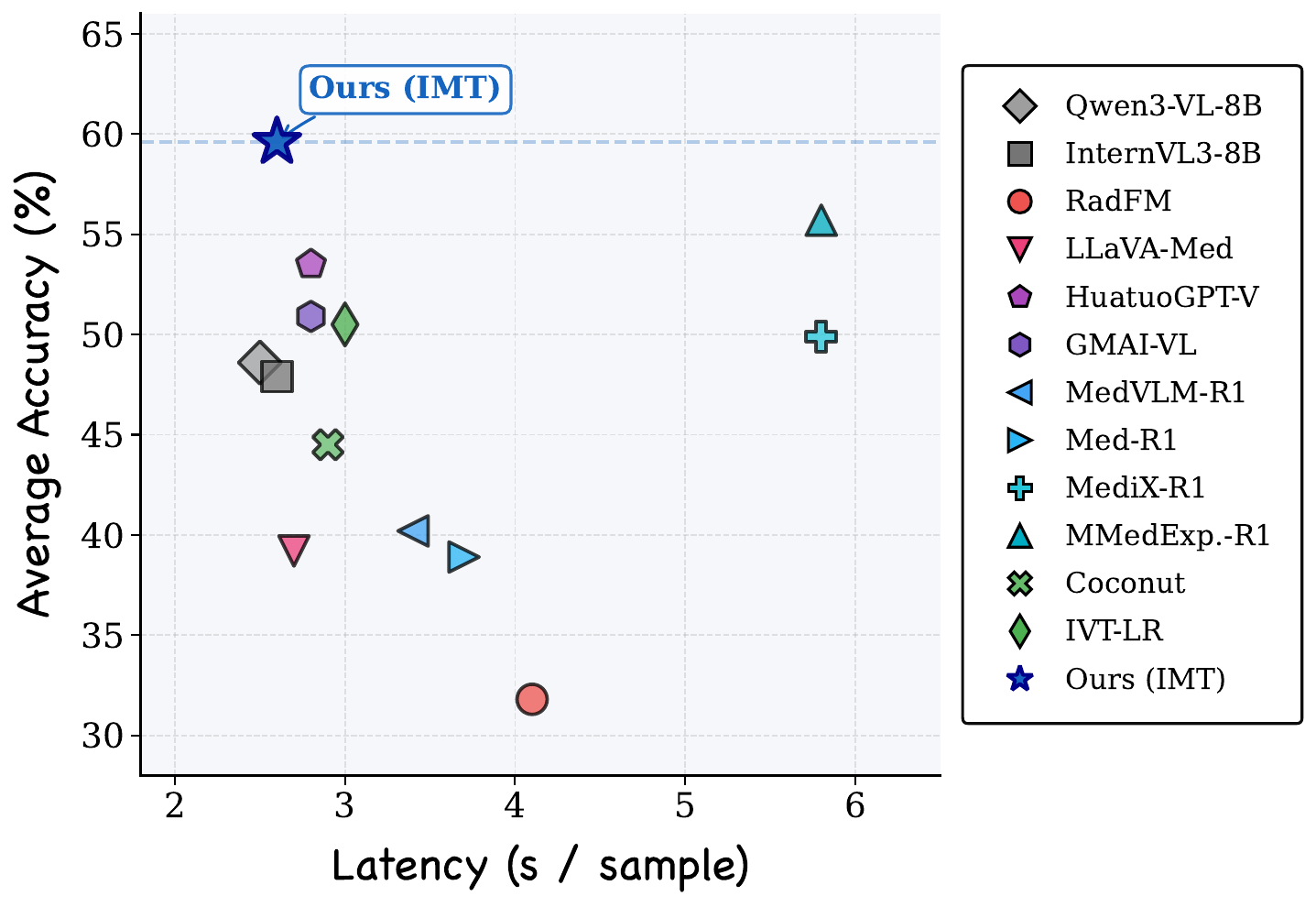}
%        \vspace{-10pt}
        \caption{Accuracy--latency trade-off across compared VLM categories.}
        \label{fig:efficiency}
    \end{minipage}
    
    \vspace{10pt} % 两图之间的间距
    
    % 第二张图
    \begin{minipage}{\linewidth}
        \centering
        \includegraphics[width=\linewidth]{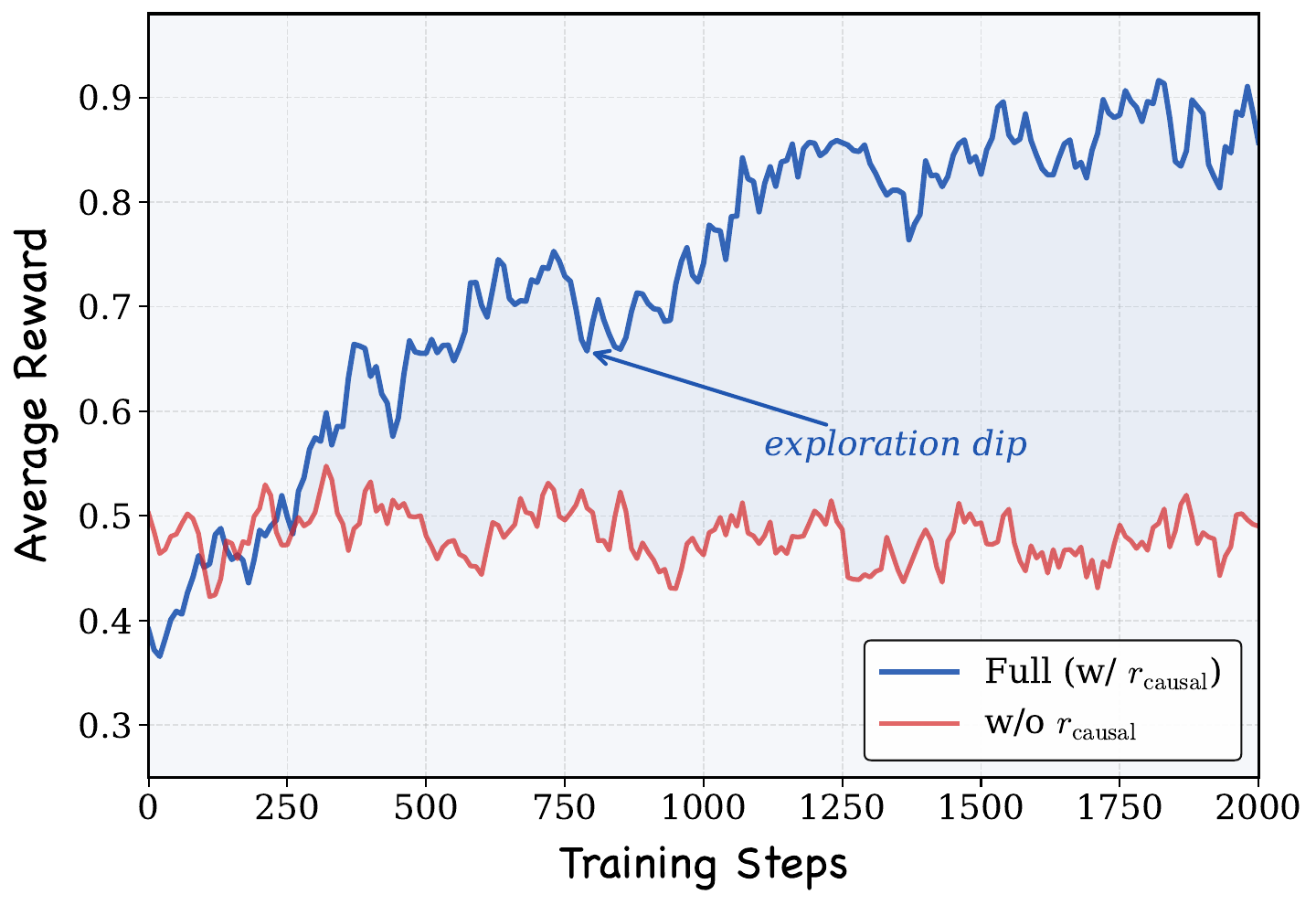}
%        \vspace{-10pt}
        \caption{The RL training reward dynamics with and without $r_{\textit{causal}}$.}
        \label{fig:training}
    \end{minipage}
    \vspace{-15pt}
\end{wrapfigure}
\textbf{Performance--efficiency trade-off.}
As shown in Fig.~\ref{fig:efficiency}, \method (IMT) achieves 59.6\% at 2.6\,s/sample, comparable to zero-shot Qwen3-VL-8B (48.6\%, 2.8\,s) since both share the same backbone and the 16 memory vectors add negligible overhead. Full-scale 7--8B CoT methods (MediX-R1, MMedExpert-R1) require 5.8\,s each due to 300--400 autoregressive reasoning tokens, while smaller CoT models (MedVLM-R1 2B, Med-R1 3B) offset verbosity with faster per-token speed yet remain 18--21\,pp below \method. This confirms that \textit{\textbf{compact latent memory provides diagnostic grounding without the token-generation overhead of full-scale CoT}}.

\vspace{1mm}
\noindent\textbf{Training dynamics.}
Fig.~\ref{fig:training} shows the full model (\textit{w/} $r_{\textit{causal}}$) improving steadily to $\sim$0.88 with a transient exploration dip near step~900, where the policy sacrifices reward to explore memory-reliant generation strategies, while the \textit{w/o} $r_{\textit{causal}}$ ablation plateaus at $\sim$0.48 throughout the training. This confirms that accuracy-only reward cannot distinguish memory-dependent from shortcut trajectories; without causal pressure the model bypasses $\mathcal{M}$ entirely, treating injected memory as inert padding.

\begin{figure}[t]
    \centering
    \includegraphics[width=\linewidth]{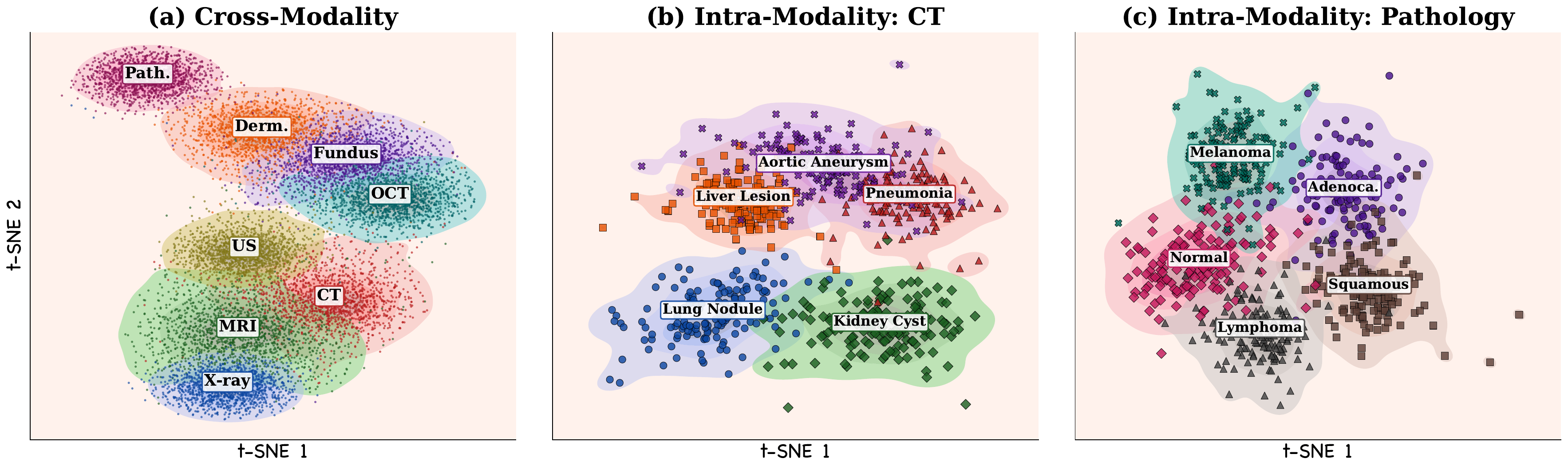}
    \caption{t-SNE visualization of implicit memory $\mathcal{M}_{\textit{auto}}$ after CCR. (a)~Eight imaging modalities form well-separated clusters with clinically coherent proximity. (b,\,c)~Within CT and Pathology, disease subtypes further segregate into distinct regions.}
    \label{fig:latent_tsne}
\end{figure}

\vspace{1mm}
\noindent\textbf{Latent space structure.}
Fig.~\ref{fig:latent_tsne} visualizes the evolved memory $\mathcal{M}_{\textit{auto}}$ via t-SNE across three granularities. At the cross-modality level~(a), eight imaging types form compact clusters with clinically coherent proximity (\eg, CT and MRI lie adjacent; dermoscopy and fundus form a nearby pair). Within individual modalities~(b,\,c), disease subtypes further segregate: CT memory separates lung nodules, liver lesions, kidney cysts, pneumonia, and aortic aneurysms, while pathology memory distinguishes adenocarcinoma, squamous cell carcinoma, normal tissue, lymphoma, and melanoma. This hierarchical organization confirms that $r_{\textit{causal}}$ \textit{\textbf{reshapes the latent space into a diagnostically meaningful manifold}} rather than merely boosting task accuracy.

\vspace{1mm}
\noindent\textbf{Why latent memory evolution works.}
Our ablations pinpoint two necessary conditions that general latent methods lack. \textit{First, structured priors are indispensable}: replacing MedSAM3 with a random encoder collapses Avg from 67.7\% to 52.0\% (Table~\ref{tab:ablation}d). \textit{Second, causal calibration activates the priors}: $r_{\textit{causal}}$ lifts accuracy by 4.1\,pp (Table~\ref{tab:ablation}b) and reorganizes memory into the hierarchical diagnostic manifold shown in Fig.~\ref{fig:latent_tsne}. Neither condition alone suffices, and their synergy is precisely what general latent methods lack.

\section{Conclusion and Future Work}
We propose MedSynapse-V, a medical vision-language model that performs clinical reasoning through compact latent tokens rather than explicit chain-of-thought generation. By combining causal counterfactual rewards with progressive memory evolution, our approach effectively internalizes diagnostic reasoning within a low-latency framework. Experiments across multiple medical benchmarks show that MedSynapse-V outperforms existing medical VLMs, general-purpose VLMs, and RL-based CoT methods in both accuracy and efficiency, confirming that latent cognitive processes guided by well-designed rewards can effectively replace verbose explicit reasoning in the medical domain.

Looking ahead, we aim to extend latent memory evolution to longitudinal analysis and multi-modal report generation by integrating heterogeneous clinical evidence sources. Our research will further investigate scaling implicit memory to accommodate broader differential diagnosis spaces with hundreds of competing hypotheses, validating the generalizability of latent cognitive architectures for complex clinical decision-making in high-stakes diagnostic environments.

%One limitation is that the current evaluation is confined to existing medical VQA benchmarks whose scale and reasoning complexity may underestimate the full potential of latent memory evolution. Extending implicit memory to more demanding scenarios such as longitudinal analysis, multi-modal report generation, and cross-institutional deployment where domain shift challenges memory generalization remains an important direction for future investigation.

%\clearpage  % TODO FINAL: This \clearpage needs to be removed from both review and camera-ready versions.

% \section*{Acknowledgements}
% Please insert your acknowledgments here.

% ---- Bibliography ----
%
% BibTeX users should specify bibliography style 'splncs04'.
% References will then be sorted and formatted in the correct style.
%

% ====================================================================
%  §A  Implementation Details
% ====================================================================
\section{Implementation Details}
\label{sec:implementation}
%This section provides comprehensive implementation specifics and the complete hyperparameter suite to ensure the reproducibility of our results.

\paragraph{Training Configuration}

Table~\ref{tab:hyperparameters} provides the hyperparameter configuration across all three training stages for reproducibility.
We employ standard data augmentation techniques to improve training robustness, including random rotation ($\pm 15°$), horizontal flipping (probability 0.5), brightness/contrast adjustment ($\pm 10\%$), and color jittering, while preserving critical diagnostic features and anatomical orientations.
Images are processed at native dynamic resolution following Qwen3-VL's default configuration (min pixels=$256\times28\times28$, max pixels=$1280\times28\times28$).
All experiments are conducted five times and we report the mean.

\begin{wrapfigure}{r}{0.6\linewidth}
    \centering
    \vspace{-0pt}
    \includegraphics[width=\linewidth]{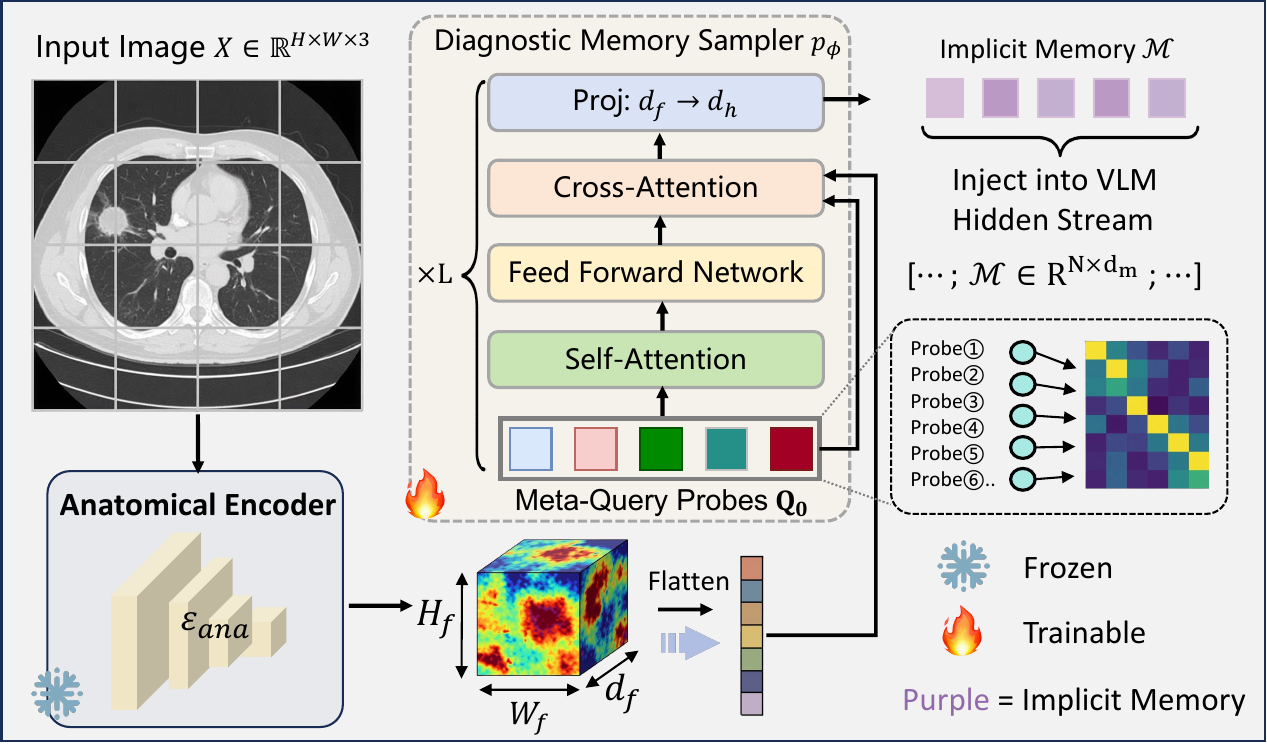}
\caption{Detailed architecture of the Diagnostic Memory Sampler $\mathcal{P}_\phi$. The frozen anatomical encoder $\mathcal{E}_{\textit{ana}}$ extracts spatial features $\mathbf{F} \in \mathbb{R}^{H_f \times W_f \times d_f}$, which are flattened into a token sequence and used as key--value pairs for the learnable meta-query probes $\mathbf{Q}_0$. Through $L$ layers of self-attention, feed-forward processing, cross-attention, and a final linear projection ($d_f \to d_h$), the module produces $N$ compact implicit memory $\mathcal{M} \in \mathbb{R}^{N \times d_h}$ that are injected into the VLM hidden stream between the question encoding and answer positions.}
    \label{fig:module}
    \vspace{-10pt}
\end{wrapfigure}

\begin{table}[t]
\centering
\small
\caption{Complete hyperparameter configuration for all three training stages.}
\renewcommand{\arraystretch}{1.1}
\label{tab:hyperparameters}
\resizebox{\textwidth}{!}{%
\begin{tabular}{l l || l l}
\toprule
\textbf{Hyperparameter} & \textbf{Value} & \textbf{Hyperparameter} & \textbf{Value} \\
\midrule
\rowcolor{myyellow}
%%% --------- Left: Base Architecture / Right: Stage II --------- %%%
\multicolumn{2}{l||}{\textbf{\textit{Base Architecture}}} & \multicolumn{2}{l}{\textbf{\textit{Stage~II: CCR}}} \\
\cmidrule(lr){1-2}\cmidrule(lr){3-4}
VLM Backbone & Qwen3-VL-8B-Instruct & Trainable Module & LoRA adapters on VLM \\
Anatomical Encoder $\mathcal{E}_{\textit{ana}}$ & MedSAM3 (ViT-B, frozen) & LoRA Rank $r$ / Alpha $\alpha$ & 64 / 128 \\
Hidden Dimension $d_h$ & 4096 & LoRA Dropout & 0.05 \\
Diagnostic Probe Count $N$ & 16 & LoRA Parameters & $\sim$83.9M \\
\cmidrule(lr){1-2}
%%% --------- Left: Stage I / Right: Stage II (cont.) --------- %%%
\multicolumn{2}{>{\columncolor{myyellow}}l||}{\textbf{\textit{Stage~I: MQPM Warmup}}} & RL Algorithm & GRPO \\
\cmidrule(lr){1-2}
Trainable Module & $\mathcal{P}_\phi$ only & Group Size $G$ & 4 \\
$\mathcal{P}_\phi$ Attention Heads & 8 & Clipping Coefficient $\varepsilon$ & 0.2 \\
$\mathcal{P}_\phi$ FFN Dimension & 4096 & $\lambda_{\textit{acc}}$ / $\lambda_{\textit{causal}}$ & 1.0 / 0.5 \\
$\mathcal{P}_\phi$ Parameters & $\sim$12.6M & Max Generation Length & 1024 tokens \\
Learning Rate & $2\times10^{-4}$ & Rollout Batch Size & 32 \\
Optimizer & AdamW ($\beta_1{=}0.9$, $\beta_2{=}0.999$) & Training Steps & 200 \\
Weight Decay & $1\times10^{-2}$ & Learning Rate & $1\times10^{-5}$ \\
Training Epochs & 3 & Temperature (sampling) & 0.7 \\
Batch Size & 32 & \multicolumn{2}{l}{} \\
%\cmidrule(lr){3-4}
%%% --------- Left: Stage I (cont.) / Right: Stage III --------- %%%
Warmup Ratio & 0.03 & 
\multicolumn{2}{>{\columncolor{myyellow}}l}{\textbf{\textit{Stage~III: IMT}}} \\
%\cmidrule(lr){3-4}
LR Schedule & Cosine Annealing & Trainable Module & $\mathcal{A}_\psi$ only \\
\cmidrule(lr){1-2}
%%% --------- Left: Infrastructure / Right: Stage III (cont.) -- %%%
\multicolumn{2}{>{\columncolor{myyellow}}l||}{\textbf{\textit{Infrastructure}}} & $\mathcal{A}_\psi$ Architecture & 2-layer MLP + LayerNorm \\
\cmidrule(lr){1-2}
GPUs & $4\times$ A100 (80GB) & $\mathcal{A}_\psi$ Hidden Dim & 4096 \\
Gradient Accumulation & 2 steps & $\mathcal{A}_\psi$ Parameters & $\sim$33.6M \\
Mixed Precision & bf16 + FlashAttention-2 & JSD Coefficient $\beta$ & 0.5 \\
Total Training Time & $\sim$38 hours & Learning Rate & $1\times10^{-4}$ \\
Cross-validation & 5-fold random seed & Training Epochs & 3 \\
\bottomrule
\end{tabular}
}
\end{table}

\paragraph{Architectural Details}

The architecture comprises several integrated components: the Diagnostic Memory Sampler $\mathcal{P}_\phi$ is implemented as a 2-layer (L=2) Transformer featuring 8 heads (head dimension 128) and 16 meta-query probes $\mathbf{Q}_0 \in \mathbb{R}^{16 \times 1024}$ initialized via a truncated normal distribution ($\sigma=0.02$), followed by a final linear projection to the 4096-dimensional hidden space; concurrently, the Autonomous Memory Module $\mathcal{A}_\psi$ processes pooled visual features through two 4096-dimensional linear layers with GELU activation and LayerNorm to produce an $N \times d_h$ representation. For anatomical encoding, we utilize the MedSAM3 ViT-B backbone pretrained on 11 imaging modalities, which extracts $64 \times 64 \times 1024$ spatial features (flattened to $M=4096$ tokens) and provides highest-confidence region masks $\mathbf{B}$ via its segmentation head (threshold 0.7) to guide the causal counterfactual reward. Finally, the model is optimized in Stage II using LoRA adapters ($r=64, \alpha=128$) applied to all attention projection matrices across the 32 layers of Qwen3-VL-8B, resulting in approximately 83.9M trainable parameters ($\sim$1.0\% of the backbone) to ensure efficient RL-driven adaptation while preserving the integrity of the pretrained knowledge.

\paragraph{Evaluation Details}
For quantitative evaluation, VQA-RAD, PMC-VQA, MMMU*, MedXpertQA-MM, and GMAI-MMBench are evaluated exclusively with the closed-ended template (Fig.~\ref{fig:prompt_closed}). SLAKE and PathVQA contain both closed-ended and open-ended subsets: the corresponding template is applied to each subset respectively, and overall accuracy is reported by aggregating both. 
For closed-ended VQA tasks, we extract the predicted answer by matching the first occurrence of option letters (A/B/C/D/E) in the generated response.
If no explicit option is found, we perform fuzzy string matching against candidate answers.
For GMAI-MMBench and MedXpertQA-MM, we follow their official evaluation scripts to ensure cross-study comparability.
All evaluations use greedy decoding (temperature${=}0$, top-$p{=}1.0$) with a maximum generation length of 512 tokens.
The 16 diagnostic memory vectors are injected at positions immediately following the question encoding, as described in \S3.2 of the main paper.

\section{Additional Qualitative Results}
\label{sec:qualitative}

\begin{figure*}[t]
\centering
\includegraphics[width=1\textwidth]{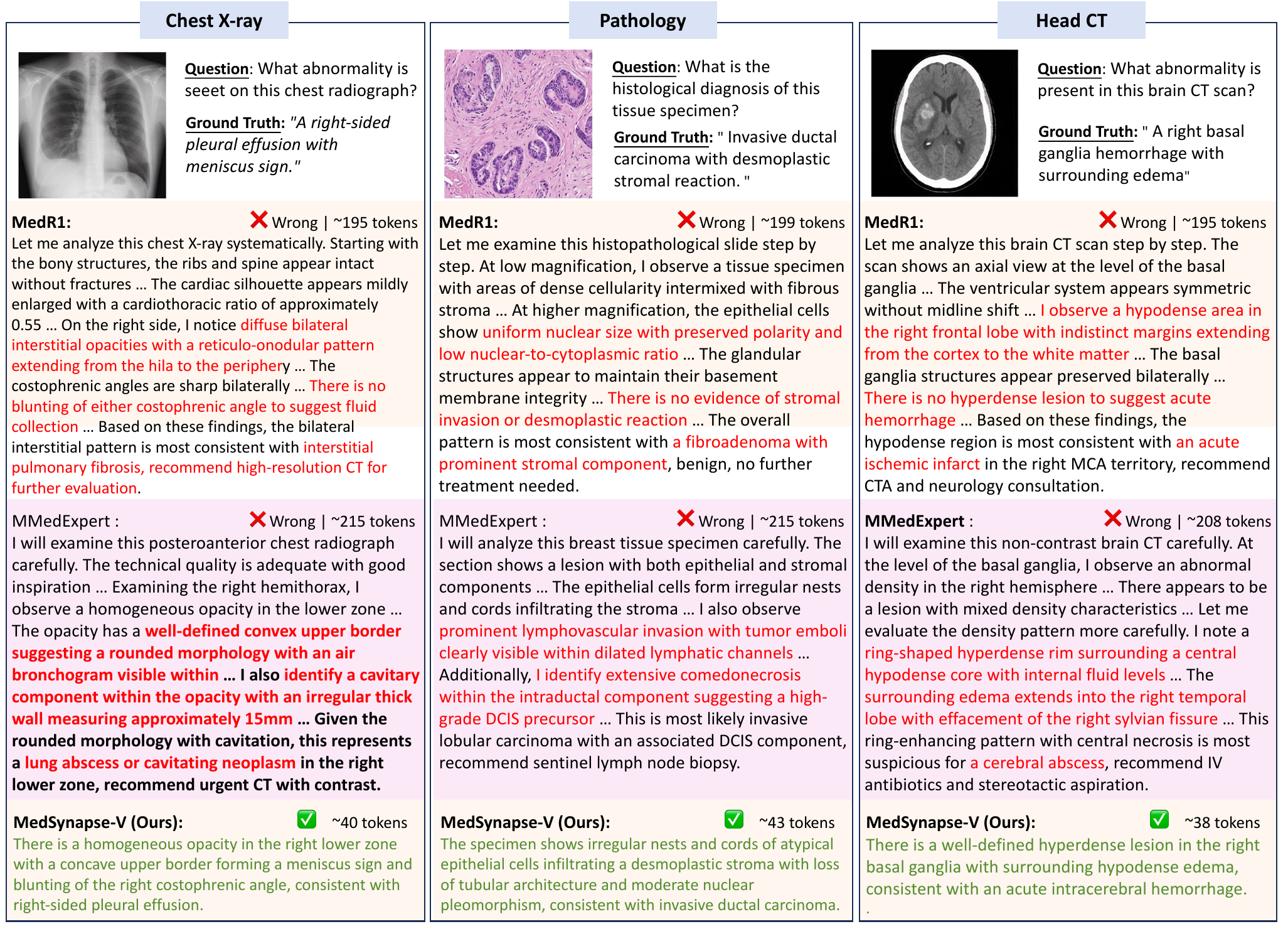}
\caption{Qualitative comparison across Chest X-ray, Pathology, and Head CT cases. \method produces concise, correct diagnoses ($\sim$38\textendash43 tokens), while other methods generate verbose CoT ($\sim$195\textendash215 tokens) with hallucinated findings (\textcolor{red}{red}).}
\label{fig:qual_success}
\end{figure*}

\subsection{Additional Representative Cases}

Figure~\ref{fig:qual_success} illustrates comparative evaluations between \method and two competitive RL-CoT baselines across diverse modalities. While Med-R1 and MMedExpert-R1 generate extensive reasoning chains, they frequently yield erroneous diagnoses as a result of hallucinatory observations that propagate and amplify throughout the inference process. In the chest X-ray case, Med-R1 fabricates bilateral interstitial opacities and claims sharp costophrenic angles, missing the obvious pleural effusion; MMedExpert-R1 hallucinates a convex border with cavitation and misdiagnoses a lung abscess. In the pathology case, Med-R1 incorrectly describes preserved polarity and intact basement membranes to conclude fibroadenoma, while MMedExpert-R1 fabricates lymphovascular invasion and comedonecrosis to misclassify as invasive lobular carcinoma. In the head CT case, Med-R1 denies the presence of a hyperdense lesion and diagnoses ischemic infarct, whereas MMedExpert-R1 hallucinates ring enhancement with central necrosis and concludes cerebral abscess. In contrast, \method directly identifies the correct findings in 38\textendash43 tokens without explicit CoT, demonstrating that \textit{latent diagnostic memory provides sufficient guidance while avoiding hallucination cascades}.

\subsection{Failure Case Analysis}
\begin{wrapfigure}{r}{0.6\linewidth}
    \centering
    \vspace{-20pt}
    \includegraphics[width=\linewidth]{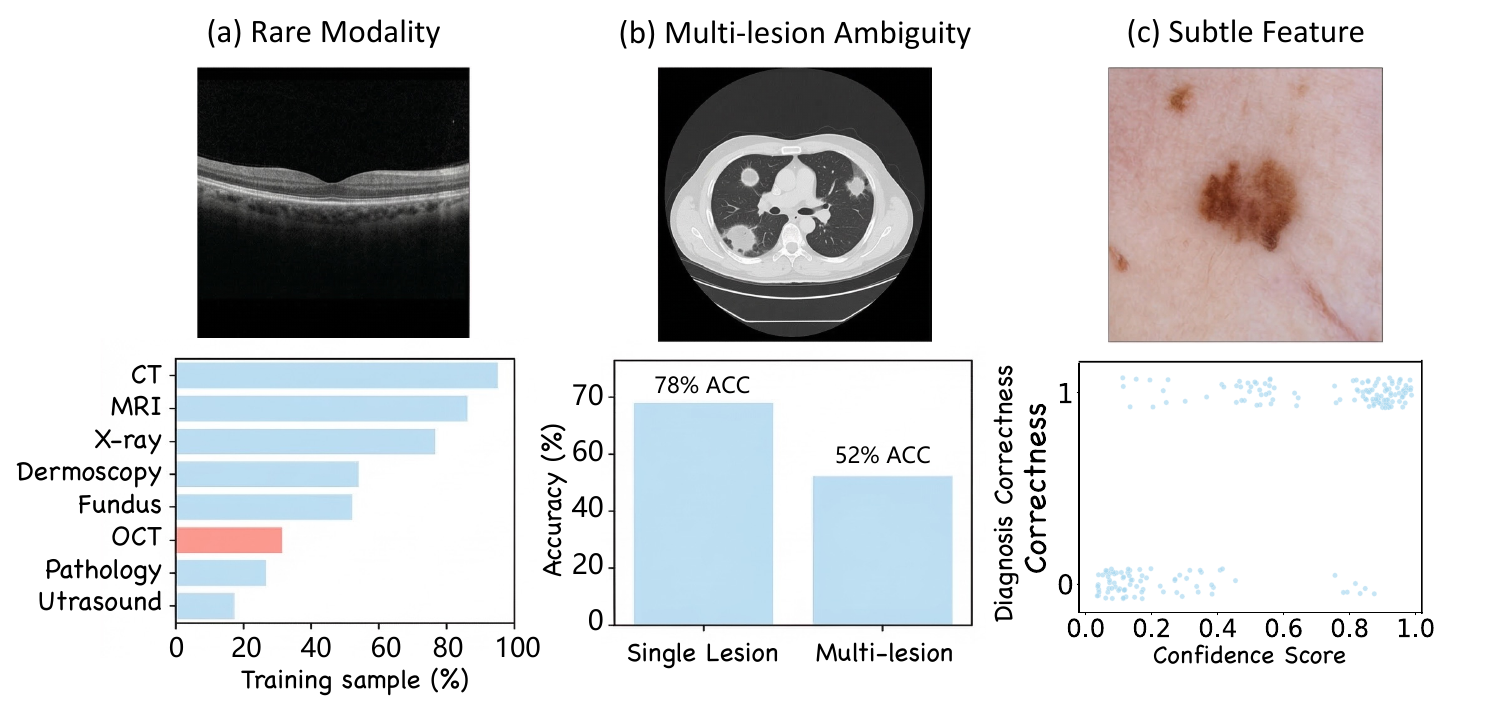}
    \caption{Three representative challenging modes.}
    \label{fig:failure_cases}
    \vspace{-10pt}
\end{wrapfigure}
Figure~\ref{fig:failure_cases} illustrates three primary failure modes.
\textbf{(a)~Rare modality} under-representation: OCT, constituting the smallest training proportion ($\sim$25\%), exhibits the lowest per-modality accuracy, indicating that memory quality degrades when prior exposure is insufficient.
\textbf{(b)~Multi-lesion ambiguity:} accuracy drops from 78\% on single-lesion images to 52\% on multi-lesion cases, as the fixed $N{=}16$ memory pool becomes saturated when multiple co-occurring pathologies compete for representational capacity.
\textbf{(c)~Subtle feature discrimination:} each scatter point represents one evaluation sample from the dermoscopy subset, where the $x$-axis is the model's confidence defined as the mean token-level generation probability $\text{conf}(\mathbf{o}) = \frac{1}{|\mathbf{o}|}\sum_{t} \pi_\theta(\mathbf{o}_t \mid X, q, \mathcal{M}, \mathbf{o}_{<t})$, and the $y$-axis is binary diagnostic correctness (1=correct, 0=incorrect; vertical jitter applied for visibility). While high-confidence predictions are predominantly correct, a notable cluster at $\text{conf} < 0.3$ with correctness$=0$ reveals that borderline cases (\eg, benign vs.\ dysplastic nevi) fall below the memory's discriminative granularity.
These modes point to future directions including balanced modality sampling, adaptive memory pool sizing, and calibrated uncertainty estimation.

% ====================================================================
%  §B  Benchmark Dataset Statistics
% ====================================================================
\section{Benchmark Dataset Statistics}
\label{sec:dataset_details}

We evaluate our model across below medical benchmarks, using official test splits where available. Our evaluation suite covers both closed-ended (CE) and multi-choice (MC) formats: (1) CE tasks include VQA-RAD~\cite{lau2018dataset} (451 radiology questions), SLAKE~\cite{liu2021slake} (1,061 mixed-modality samples), and PathVQA~\cite{he2020pathvqa} (6,719 pathology samples); (2) MC tasks comprise PMC-VQA~\cite{zhang2023pmc} (10,000 samples), the Health \& Medicine track of MMMU~\cite{yue2024mmmu} (150 samples), and the expert-level MedXpertQA-MM~\cite{zuo2025medxpertqa} (960 samples). Additionally, we evaluate on GMAI-MMBench~\cite{ye2024gmai}, a multi-granularity benchmark spanning 38 distinct modalities with 2,847 questions. Accuracy is the primary metric across all benchmarks, except for MedXpertQA-MM which uses a Total Score.

The training pipeline follows three progressive phases to enhance the medical grounding of the model. Stage I involves large scale pre training with 50K image text pairs from PubMedVision~\cite{chen2024towards}. These samples were rigorously curated by medical experts to ensure accurate alignment across radiology modalities like CT, MRI, and X ray along with pathology. Stage II constructs a specialized mixed modality reinforcement learning set of 4K samples. These instances were also selected by clinical professionals to prioritize high diagnostic value, including 3K closed ended VQA samples from expert annotated OmniMedVQA~\cite{hu2024omnimedvqa} and 1K open ended samples from SLAKE and PathVQA. Stage III refines the model by reusing the Stage II data with identical preprocessing to ensure consistent optimization. Importantly, rigorous filtering was applied across all dataset splits to ensure that no evaluation test samples overlap with any training data, maintaining the absolute integrity of our zero shot assessment.

% ====================================================================
%  §D  Additional Analysis
% ====================================================================
\section{Additional Analysis}
\label{sec:additional_analysis}

\subsection{Per-Modality Breakdown on OmniMedVQA}

Table~\ref{tab:per_modality} summarizes performance across the eight imaging modalities in OmniMedVQA.
\method achieves consistent performance gains across all eight OmniMedVQA modalities, with the largest gains on radiology-centric modalities (CT: $+14.4$, MRI: $+14.9$, X-ray: $+13.1$) where structured anatomical priors are most informative.
Notably, although MMedExpert-R1 improves over the Qwen3-VL-8B zero-shot baseline across all modalities through guideline-based RL rewards, the margin remains modest on challenging modalities such as OCT ($+5.6$) and Fundus ($+5.6$), where explicit CoT reasoning struggles to capture subtle spatial patterns.
In contrast, \method's latent memory mechanism yields substantially larger gains on these same modalities ($+11.9$ and $+11.8$), confirming that continuous diagnostic memory encodes fine-grained anatomical features more effectively than discrete token reasoning.

\begin{table*}[t]
\centering
\caption{Per-modality accuracy (\%) on OmniMedVQA. $\Delta$: improvement over Qwen3-VL-8B zero-shot baseline.}
\label{tab:per_modality}
\resizebox{0.95\textwidth}{!}{%
\begin{tabular}{l|>{\centering\arraybackslash}p{1.2cm}>{\centering\arraybackslash}p{1.2cm}>{\centering\arraybackslash}p{1.2cm}>{\centering\arraybackslash}p{1.2cm}>{\centering\arraybackslash}p{1.2cm}>{\centering\arraybackslash}p{1.2cm}>{\centering\arraybackslash}p{1.2cm}>{\centering\arraybackslash}p{1.2cm}|>{\centering\arraybackslash}p{1.2cm}}
\toprule[1.2pt]
\textbf{Method} & \textbf{CT} & \textbf{MRI} & \textbf{X-ray} & \textbf{Derm.} & \textbf{Fundus} & \textbf{OCT} & \textbf{Patho.} & \textbf{US} & \textbf{Avg.} \\
\midrule
Qwen3-VL-8B (zero-shot) & 52.4 & 48.6 & 55.1 & 61.3 & 46.8 & 44.2 & 50.7 & 47.5 & 50.8 \\
MedExpert-R1~\cite{ding2026mmedexpert} & 58.6 & 55.8 & 60.2 & 66.5 & 52.4 & 49.8 & 57.2 & 54.1 & 56.8 \\
\method (IMT) & 66.8 & 63.5 & 68.2 & 72.4 & 58.6 & 56.1 & 62.9 & 60.7 & 63.7 \\
\rowcolor{myyellow}
$\Delta$ & +14.4 & +14.9 & +13.1 & +11.1 & +11.8 & +11.9 & +12.2 & +13.2 & +12.9 \\
\bottomrule[1.2pt]
\end{tabular}%
}
\end{table*}

\subsection{Visualization of Causal Counterfactual Intervention}
\label{sec:ccr_visualization}

\begin{figure}[t]
\centering
\includegraphics[width=1.0\textwidth]{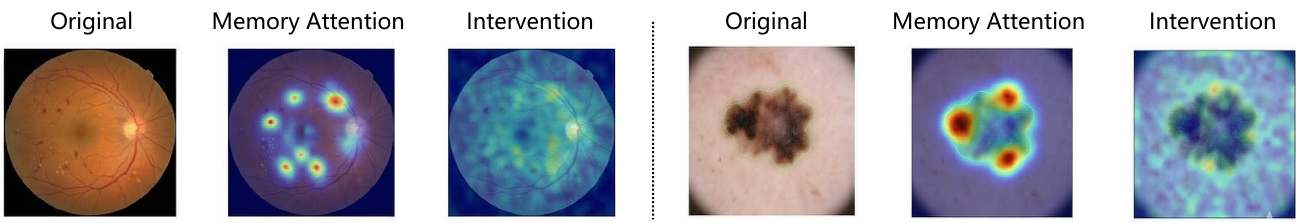}
\caption{\textbf{Causal intervention visualization} on fundus (left group) and dermoscopy (right group).
Each group: original image, MedSAM3 region mask $\mathbf{B}$, and post-CCR memory attention map.
After refinement, memory attention concentrates on diagnostically critical structures while suppressing background.}
\label{fig:ccr_vis}
\end{figure}

%\subsection{Causal Signal Enhancement}
%\vspace{0.5mm}\noindent\textbf{Gradient handling.}
% 这里画一个注意力图
%Importantly, $r_{\textit{causal}}$ is computed using the \emph{old} policy $\pi_{\theta_{\textit{old}}}$ (i.e., the snapshot taken before each GRPO update step) and is treated as a \textbf{scalar constant} with respect to the current policy $\theta$: all log-probabilities in Eq.~5 are evaluated under $\theta_{\textit{old}}$ and \textbf{detached from the computational graph} (stop-gradient). Consequently, $r_{\textit{causal}}$ contributes to the advantage estimate $\hat{A}_i$ without introducing non-stationary reward gradients, maintaining the same bias/variance profile as standard GRPO with a fixed scalar reward. This design ensures that the causal signal modulates \emph{which} trajectories are reinforced, rather than how the policy is differentiated through the reward itself.

Figure~\ref{fig:ccr_vis} illustrates how CCR reshapes the spatial distribution of memory attention through counterfactual intervention.
In the fundus case, MedSAM3 identifies retinal lesions including microaneurysms and hard exudates; after CCR, the memory attention map aligns tightly with these foci while the optic disc and healthy vasculature receive minimal activation.
In the dermoscopy case, post-CCR attention concentrates at the lesion periphery where asymmetry, border irregularity, and color heterogeneity are most pronounced, consistent with the ABCD criteria used in clinical dermoscopic assessment.
The corresponding intervention map again demonstrates substantial attention redistribution upon masking, with activation scattering to non-diagnostic background regions.
These visualizations provide direct evidence that $r_{\textit{causal}}$ successfully enforces a causal dependency between diagnostic memory and pathologically relevant image regions, complementing the quantitative mask robustness analysis in \S\ref{sec:mask_validation}.

\subsection{Inference Latency and Parameter Count}
\label{sec:efficiency}
 
\begin{wraptable}{r}{0.50\linewidth}
    \centering
    \vspace{-32pt}
    \caption{Inference latency (single A100, batch\,=\,1) and parameter count.
    \textit{Prefill}: time to first token.
    \textit{End-to-end}: full autoregressive decoding (max 128 tokens, greedy).}
    \label{tab:inference_param}
    \renewcommand{\arraystretch}{1.05}
    \resizebox{\linewidth}{!}{%
    \begin{tabular}{l|c|c}
    \toprule[1.3pt]
    \textbf{Component} & \textbf{\method (IMT)} & \textbf{Qwen3-VL-8B} \\
    \midrule[1.0pt]
    Visual Encoding & 38\,ms & 38\,ms \\
    Memory Generation ($\mathcal{A}_\psi$) & 4\,ms & -- \\
    First Decoding Step & 60\,ms & 64\,ms \\
    \midrule
    \rowcolor{myyellow}
    \textbf{Prefill (to first token)} & \textbf{102\,ms} & \textbf{102\,ms} \\
    \rowcolor{myyellow}
    \textbf{End-to-end / sample} & \textbf{$\sim$2.6\,s} & \textbf{$\sim$2.8\,s} \\
    \bottomrule[1.2pt]
    \end{tabular}%
    }
    \par\vspace{4pt}
    \resizebox{\linewidth}{!}{%
    \begin{tabular}{l|c|c}
    \toprule[1.2pt]
    \textbf{Module} & \textbf{Parameters} & \textbf{Trainable Stage} \\
    \midrule[1.0pt]
    Qwen3-VL-8B backbone & 8.29B & Frozen \\
    LoRA adapters & 83.9M & Stage II \\
    $\mathcal{P}_\phi$ (Memory Sampler) & 12.6M & Stage I \\
    $\mathcal{A}_\psi$ (Autonomous Module) & 33.6M & Stage III \\
    MedSAM3 encoder & 93.7M & Frozen (removed) \\
    \midrule
    \rowcolor{myyellow}
    \textbf{Inference total} & \textbf{8.41B} & -- \\
    \bottomrule[1.3pt]
    \end{tabular}%
    }
    \vspace{-15pt}
\end{wraptable}
 
The prefill latency of \method (IMT) is 102\,ms, identical to the vanilla baseline; $\mathcal{A}_\psi$ contributes only 4\,ms.
Although prefill overhead is negligible, the 16 memory vectors injected during this stage play a pivotal role in overall efficiency: they become part of the KV cache built at prefill time, so every subsequent decoding step can attend to condensed diagnostic priors \emph{at no extra cost}.
This latent conditioning steers the model toward shorter, more decisive outputs (${\sim}$34--44 answer tokens vs.\ ${\sim}$50--80 for the zero-shot baseline), reducing end-to-end sample latency from ${\sim}$2.8\,s to ${\sim}$2.6\,s.
CoT baselines require 5.2--5.8\,s due to 300--400 autoregressive reasoning tokens.
The inference footprint is 8.41B parameters (1.4\% over the bare backbone), as the $\mathcal{E}_{\textit{ana}}$ is entirely removed.
 
\vspace{1mm}
\noindent\textbf{Note on the ``ms/token'' column in Table~2 of the main text.}
The ms/token values in the main paper's ablation table measure \emph{average per-token decoding latency} (total decode wall time\,/\,number of generated tokens).
The zero-shot Qwen3-VL-8B baseline shows a higher value (126\,ms/token) than \method (${\sim}$102\,ms/token) because, without compact memory conditioning in the KV cache, the model's attention must scatter across the full visual token sequence at each decode step, yielding broader attention patterns and slower per-step computation.
These per-token values should not be confused with the \emph{prefill latency} (102\,ms, Table~\ref{tab:inference_param}) or the \emph{end-to-end sample latency} (${\sim}$2.6\,s, Fig.~5), which cover the complete inference pipeline.

\subsection{Memory Synthesis Design Choices}
\label{sec:injection_ablation}
 
We investigate two design dimensions of the memory synthesis module: (i)~the aggregation strategy for converting anatomical encoder features into compact memory, and (ii)~whether question tokens should condition the autonomous module $\mathcal{A}_\psi$.
 
\begin{table}[t]
\centering
\caption{\textbf{Memory synthesis design ablations.}
\textit{Left}: meta-query sampling vs.\ simpler injection (full MQPM$\to$CCR$\to$IMT pipeline, encoder-free inference).
\textit{Right}: effect of including question tokens $\mathbf{q}$ in $\mathcal{A}_\psi$ input.
Default configurations are highlighted.}
\label{tab:synthesis_ablation}
\renewcommand{\arraystretch}{1.1}
\begin{minipage}[t]{0.62\linewidth}
    \centering
    \resizebox{\linewidth}{!}{%
    \begin{tabular}{l|C{1.4cm}C{1.2cm}C{1.4cm}C{1.4cm}C{1.3cm}|C{1.0cm}}
    \toprule[1.1pt]
    \textbf{Aggregation Strategy} & \textbf{VQA-RAD} & \textbf{SLAKE} & \textbf{PathVQA} & \textbf{PMC-VQA} & \textbf{MMMU*} & \textbf{Avg.} \\
    \midrule[0.9pt]
    No injection (zero-shot) & 58.6 & 66.2 & 55.4 & 42.5 & 48.3 & 54.2 \\
    Avg-pool concat & 63.4 & 70.8 & 58.2 & 48.1 & 51.6 & 58.4 \\
    Linear projector & 65.1 & 72.3 & 59.8 & 50.4 & 53.2 & 60.2 \\
    \rowcolor{myyellow}
    Meta-query $\mathcal{P}_\phi$ (ours) & \textbf{74.2} & \textbf{79.8} & \textbf{64.8} & \textbf{58.5} & \textbf{61.4} & \textbf{67.7} \\
    \bottomrule[1.1pt]
    \end{tabular}%
    }
\end{minipage}%
\hfill
\begin{minipage}[t]{0.36\linewidth}
    \centering
    \resizebox{\linewidth}{!}{%
    \begin{tabular}{l|C{1.4cm}C{1.2cm}|C{1.0cm}}
    \toprule[1.1pt]
    $\mathcal{A}_\psi$ \textbf{Input} & \textbf{VQA-RAD} & \textbf{SLAKE} & \textbf{Avg.} \\
    \midrule[0.9pt]
    Visual only & 72.8 & 78.1 & 67.0 \\
    \rowcolor{myyellow}
    Visual + $\mathbf{q}$ (default) & \textbf{74.2} & \textbf{79.8} & \textbf{67.7} \\
    \bottomrule[1.1pt]
    \end{tabular}%
    }
\end{minipage}
\end{table}
 
\noindent\textbf{Aggregation strategy} (Table~\ref{tab:synthesis_ablation}, left).
Average-pooling MedSAM features into 16 tokens and concatenating them to the input provides +4.2\,pp over zero-shot, confirming that the anatomical encoder supplies useful priors.
A learnable linear projector (MedSAM$\to d_h$) further improves to 60.2\%.
The meta-query sampler $\mathcal{P}_\phi$ outperforms both by a substantial margin (+7.5\,pp over linear projector, +13.5\,pp over zero-shot), demonstrating that \emph{selective, input-conditioned aggregation} of spatial features via cross-attention is critical: static compression discards fine-grained spatial cues that the learnable probes can selectively retain.
 
\noindent\textbf{Query conditioning in $\mathcal{A}_\psi$} (Table~\ref{tab:synthesis_ablation}, right).
Including question tokens as input to $\mathcal{A}_\psi$ yields a consistent +0.7\,pp improvement, as query context enables $\mathcal{A}_\psi$ to generate task-relevant memory rather than generic anatomical summaries.
The gain is modest because the VLM's self-attention already conditions answer generation on $\mathbf{q}$; the additional query signal in $\mathcal{A}_\psi$ primarily helps disambiguate cases where multiple diagnostic hypotheses compete for the same visual features.

\begin{table}[t]
\centering
\caption{\textbf{Effect of mask confidence threshold $\tau$.}
Full MQPM$\to$CCR$\to$IMT pipeline; encoder-free inference.
$|\mathbf{B}|/HW$: average masked fraction.}
\label{tab:mask_threshold}
\renewcommand{\arraystretch}{1.05}
\adjustbox{width=\textwidth}{
\begin{tabular}{C{3.8cm}|C{2.0cm}|C{1.8cm}C{1.6cm}C{1.8cm}C{1.8cm}C{1.6cm}|C{1.5cm}}
\toprule[1.1pt]
\textbf{Threshold $\tau$} & $|\mathbf{B}|/HW$ (\%) & \textbf{VQA-RAD} & \textbf{SLAKE} & \textbf{PathVQA} & \textbf{PMC-VQA} & \textbf{MMMU*} & \textbf{Avg.} \\
\midrule[0.9pt]
0.3 & 42.6 & 71.4 & 77.0 & 62.5 & 55.8 & 58.2 & 65.0 \\
0.5 & 28.3 & 73.1 & 78.6 & 63.8 & 57.4 & 60.1 & 66.6 \\
\rowcolor{myyellow}
0.7 (default) & 15.8 & \textbf{74.2} & \textbf{79.8} & \textbf{64.8} & \textbf{58.5} & \textbf{61.4} & \textbf{67.7} \\
0.8 & 10.1 & 73.8 & 79.2 & 64.2 & 58.0 & 60.8 & 67.2 \\
0.9 & 5.4 & 72.5 & 77.8 & 63.0 & 56.5 & 59.0 & 65.8 \\
\midrule
No mask ($r_{\textit{acc}}$ only) & -- & 70.1 & 75.8 & 61.2 & 54.3 & 56.6 & 63.6 \\
\bottomrule[1.1pt]
\end{tabular}
}
\end{table}

 \begin{table}[t]
\centering
\caption{\textbf{Effect of mask rank selection.}
Rank-1: highest confidence (default); Rank-2: second-highest; Random: uniformly sampled candidate.}
\label{tab:mask_rank}
\renewcommand{\arraystretch}{1.05}
\adjustbox{width=\textwidth}{
\begin{tabular}{l|C{1.8cm}C{1.6cm}C{1.8cm}C{1.8cm}C{1.6cm}|C{1.5cm}|C{2.2cm}}
\toprule[1.1pt]
\textbf{Mask Selection} & \textbf{VQA-RAD} & \textbf{SLAKE} & \textbf{PathVQA} & \textbf{PMC-VQA} & \textbf{MMMU*} & \textbf{Avg.} & $\mathbf{\Delta}$ \\
\midrule[0.9pt]
\rowcolor{myyellow}
Rank-1 (default) & \textbf{74.2} & \textbf{79.8} & \textbf{64.8} & \textbf{58.5} & \textbf{61.4} & \textbf{67.7} & -- \\
Rank-2 & 73.0 & 78.6 & 63.6 & 57.2 & 60.0 & 66.5 & $-1.2$ \\
Random & 71.8 & 77.2 & 62.4 & 56.0 & 58.4 & 65.2 & $-2.5$ \\
\midrule
No mask ($r_{\textit{acc}}$ only) & 70.1 & 75.8 & 61.2 & 54.3 & 56.6 & 63.6 & $-4.1$ \\
\bottomrule[1.1pt]
\end{tabular}
}
\end{table}

\subsection{Robustness of Causal Intervention to Mask Quality}
\label{sec:mask_validation}
 
The causal counterfactual reward $r_{\textit{causal}}$ relies on region masks $\mathbf{B}$ from MedSAM3.
We verify robustness via two ablations: (i)~varying the binarization threshold~$\tau$, and (ii)~replacing the top-1 mask with lower-ranked candidates.
\textbf{\textbf{(i) Confidence threshold $\tau$.}}
Table~\ref{tab:mask_threshold} reports accuracy under five thresholds (default $\tau{=}0.7$).
All thresholds outperform the no-mask baseline (Avg 63.6\%).
Performance is stable across $\tau \in [0.5, 0.8]$ (spread only 1.1\,pp), confirming that $r_{\textit{causal}}$ does not require pixel-perfect boundaries.
Extreme values degrade: $\tau{=}0.3$ masks 42.6\% of the image (intervention too destructive); $\tau{=}0.9$ masks only 5.4\% (intervention too weak).
\textit{\textbf{(ii) Mask rank selection.}}
We replace MedSAM3's top-1 mask with its second-highest confidence candidate (Rank-2) or a random candidate.
Rank-2 retains most of the gain ($-1.2$\,pp vs.\ Rank-1), and even random masks outperform the no-mask baseline by 1.6\,pp.
The monotonic ordering Rank-1 $>$ Rank-2 $>$ Random $>$ None confirms that mask quality helps but is not critical: the causal reward exploits the \emph{relative contrast} between masked and unmasked conditions rather than relying on pixel-precise delineation.

 %\clearpage
%\newpage

% ==========================================
% 方案一代码
% ==========================================

\begin{figure}[t]
\centering
% 使用 tcolorbox 替代 fbox + parbox
\begin{tcolorbox}[
    width=0.92\columnwidth,      % 设置宽度
    boxrule=0.5pt,              % 设置边框粗细（与 fbox 接近）
    colback=gray!10!white,       % 设置非常浅的灰色背景 (10% 灰 + 90% 白)
    colframe=black,             % 设置边框颜色
    arc=0pt,outer arc=0pt,      % 设置直角边框（移除圆角）
    top=3pt,bottom=3pt,left=5pt,right=5pt, % 设置内部边距
    fontupper=\small            % 设置内部字体大小
]
\textbf{System:} You are a helpful medical assistant. Answer the question based on the image.\\[3pt]
\textbf{User:} \texttt{<image>}\\
\texttt{\{question\}}\\
Options:\\
\texttt{(A) \{option\_a\}}\\
\texttt{(B) \{option\_b\}}\\
\texttt{(C) \{option\_c\}}\\
\texttt{(D) \{option\_d\}}\\
Please answer with the option letter only.\\[3pt]
\textbf{Assistant:}
\end{tcolorbox}
\vspace{-2pt}
%{\footnotesize \textit{Note:} For \method (IMT), $\mathcal{A}_\psi$ autonomously generates $\mathcal{M}_{\textit{auto}}$ from VLM visual features and injects them into the hidden stream. No external encoder or special token is involved; the surface-level prompt is identical to that of the base VLM.}
\caption{Prompt template for \textbf{closed-ended} multi-choice VQA (VQA-RAD, SLAKE, PathVQA, PMC-VQA, MMMU*, MedXpertQA-MM, GMAI-MMBench). The number of options varies by dataset (2--5); the template adapts accordingly.}
\label{fig:prompt_closed}
\end{figure}

\begin{figure}[t]
\centering
% 使用 tcolorbox 替代 fbox + parbox
\begin{tcolorbox}[
    width=0.92\columnwidth,      % 设置宽度
    boxrule=0.5pt,              % 设置边框粗细
    colback=gray!10!white,       % 设置非常浅的灰色背景
    colframe=black,             % 设置边框颜色
    arc=0pt,outer arc=0pt,      % 设置直角边框
    top=3pt,bottom=3pt,left=5pt,right=5pt, % 设置内部边距
    fontupper=\small            % 设置内部字体大小
]
\textbf{System:} You are a helpful medical assistant. Provide a concise answer to the question.\\[3pt]
\textbf{User:} \texttt{<image>}\\
\texttt{\{question\}}\\
Answer the question using a single word or phrase.\\[3pt]
\textbf{Assistant:}
\end{tcolorbox}
\vspace{-2pt}
\caption{Prompt template. Notably, $\mathcal{M}_{\textit{auto}}$ is autonomously generated and injected in the hidden stream without altering the text prompt.}
\label{fig:prompt_open}
\end{figure}

\section{Evaluation Prompt Templates}
\label{sec:prompts}

We adopt minimal, zero-shot prompt templates for all evaluations to avoid biasing the model through elaborate instructions and to ensure fair comparison across methods.
Following prior medical VLM evaluation practices~\cite{pan2025medvlm,lai2026med,chen2024towards,mullappilly2026medix}, we use a brief system instruction paired with the clinical query and image, without few-shot exemplars or chain-of-thought elicitation.
This design isolates the effect of each model's intrinsic capabilities (or, in our case, latent diagnostic memory) from prompt engineering.

The qualitative case analyses in the main paper and this supplement uniformly use the open-ended template to reveal each model's complete diagnostic reasoning.
Fig.~\ref{fig:prompt_closed} and~\ref{fig:prompt_open} present the exact prompt templates used for closed-ended and open-ended evaluation, respectively.
For \method, the Autonomous Memory Module $\mathcal{A}_\psi$ generates diagnostic implicit memory $\mathcal{M}_{\textit{auto}} = \{m_1, \ldots, m_{16}\}$ directly from the VLM's own visual encoding features and injects them into the hidden stream between the question encoding and the answer generation position (see \S2.4 in the main text).
The entire process is transparent to the surface-level prompt: \textit{no additional text tokens, special markers, or reasoning elicitation instructions are required}, distinguishing \method from both explicit CoT methods (which append reasoning instructions such as ``\texttt{Let's think step by step}'') and other latent reasoning methods that require special delimiters (\eg, Coconut's \texttt{<bot>}/\texttt{<eot>} markers~\cite{hao2024training} or Heima's \texttt{<CoT>} tokens~\cite{shen2025codi}).

\vspace{1mm}
\noindent\textbf{Answer extraction.}
For closed-ended tasks, we extract the first valid option letter (A/B/C/D/E) from the generated output using regex matching.
For CoT baselines that produce structured tags (\eg, \texttt{<answer>B</answer>}), we parse the content within the answer tags.
If no valid option is detected, the response is marked as incorrect.
For open-ended tasks, we follow prior work~\cite{liu2021slake,he2020pathvqa} and perform exact string matching after lowercasing and stripping punctuation.

\vspace{1mm}
\noindent\textbf{Decoding configuration.} All models are evaluated with greedy decoding (temperature $= 0$, top-$p = 1.0$) to ensure deterministic and reproducible outputs. The maximum generation length is set to 128 tokens for \method and other direct-answer models, and 1024 tokens for CoT baselines to accommodate their verbose reasoning traces. Note that the 16 diagnostic memory vectors ($N{=}16$) are injected into the hidden stream as continuous embeddings and do \emph{not} count toward the generated token budget; the model’s actual text output for closed-ended tasks is typically 1–3 tokens and open-ended for 20-40 tokens.

\begin{table*}[t]
    \centering
    \caption{%
        \textit{Left}: effect of memory injection position on diagnostic accuracy (\%), where $\mathbf{V}$ and $\mathbf{q}$ denote visual and question tokens.
         \textit{Right}: effect of GRPO group size $G$ on accuracy (\%) and training cost.
        All results use the full three-stage pipeline with IMT inference. Default configurations are highlighted.
    }
    \label{tab:ablation_injection_lambda}
    \renewcommand{\arraystretch}{1.1}
    \begin{minipage}[t]{0.50\linewidth}
        \centering
        \resizebox{\linewidth}{!}{%
        \begin{tabular}{l >{\centering\arraybackslash}p{1.6cm}>{\centering\arraybackslash}p{1.4cm}>{\centering\arraybackslash}p{1.5cm}>{\centering\arraybackslash}p{1.6cm}>{\centering\arraybackslash}p{0.9cm}}
        \toprule[1.3pt]
        \textbf{Injection Position} & \textbf{VQA-RAD} & \textbf{SLAKE} & \textbf{PathVQA} & \textbf{PMC-VQA} & \textbf{Avg.} \\
        \midrule[1.0pt]
        Before $\mathbf{V}$ & 70.2 & 75.6 & 61.3 & 54.8 & 65.5 \\
        $\mathbf{V} \to \mathcal{M} \to \mathbf{q}$ & 72.5 & 77.8 & 63.1 & 56.9 & 67.6 \\
        \rowcolor{myyellow}
        After $\mathbf{q}$ (default) & \textbf{74.2} & \textbf{79.8} & \textbf{64.8} & \textbf{58.5} & \textbf{69.3} \\
        Interleaved w/ $\mathbf{q}$ & 73.1 & 78.4 & 63.8 & 57.6 & 68.2 \\
        \bottomrule[1.3pt]
        \end{tabular}%
        }
    \end{minipage}%
    \hfill
    \begin{minipage}[t]{0.48\linewidth}
        \centering
        \resizebox{\linewidth}{!}{%
        \begin{tabular}{>{\centering\arraybackslash}p{0.5cm} >{\centering\arraybackslash}p{1.6cm}>{\centering\arraybackslash}p{1.4cm}>{\centering\arraybackslash}p{1.5cm}>{\centering\arraybackslash}p{1.8cm}>{\centering\arraybackslash}p{0.9cm}>{\centering\arraybackslash}p{1.2cm}}
        \toprule[1.3pt]
        $G$ & \textbf{VQA-RAD} & \textbf{SLAKE} & \textbf{PathVQA} & \textbf{PMC-VQA} & \textbf{Avg.} & \textbf{GPU-h} \\
        \midrule[1.0pt]
        2 & 72.4 & 77.6 & 62.9 & 56.2 & 67.3 & 14 \\
        \rowcolor{myyellow}
        4 & \textbf{74.2} & \textbf{79.8} & \textbf{64.8} & \textbf{58.5} & \textbf{69.3} & 28 \\
        6 & 74.3 & 80.0 & 64.9 & 58.6 & 69.5 & 40 \\
        8 & 74.5 & 80.1 & 65.0 & 58.7 & 69.6 & 55 \\
        \bottomrule[1.3pt]
        \end{tabular}%
        }
    \end{minipage}
\end{table*}

\begin{table*}[t]
    \centering
    \caption{%
        \textit{Left}: comparison of divergence measures for IMT distillation ($\beta$ controls JSD interpolation weight).
          \textit{Right}: sensitivity analysis of causal reward weight $\lambda_{\textit{causal}}$.
        Default configurations are highlighted.
    }
    \label{tab:ablation_div_grpo}
    \renewcommand{\arraystretch}{1.1}
    \begin{minipage}[t]{0.50\linewidth}
        \centering
        \resizebox{\linewidth}{!}{%
        \begin{tabular}{l >{\centering\arraybackslash}p{1.4cm}>{\centering\arraybackslash}p{1.2cm}>{\centering\arraybackslash}p{1.4cm}>{\centering\arraybackslash}p{1.4cm}>{\centering\arraybackslash}p{0.9cm}}
        \toprule[1.3pt]
        \textbf{Divergence} & \textbf{VQA-RAD} & \textbf{SLAKE} & \textbf{PathVQA} & \textbf{PMC-VQA} & \textbf{Avg.} \\
        \midrule[1.0pt]
        Forward KL & 72.1 & 77.5 & 62.8 & 56.4 & 67.2 \\
        Reverse KL & 71.8 & 77.1 & 62.3 & 55.9 & 66.8 \\
        JSD ($\beta{=}0.3$) & 73.5 & 79.0 & 64.0 & 57.8 & 68.6 \\
        \rowcolor{myyellow}
        JSD ($\beta{=}0.5$) & \textbf{74.2} & \textbf{79.8} & \textbf{64.8} & \textbf{58.5} & \textbf{69.3} \\
        JSD ($\beta{=}0.7$) & 73.8 & 79.3 & 64.3 & 58.0 & 68.9 \\
        \bottomrule[1.3pt]
        \end{tabular}%
        }
    \end{minipage}%
    \hfill
    \begin{minipage}[t]{0.46\linewidth}
        \centering
        \resizebox{\linewidth}{!}{%
        \begin{tabular}{>{\centering\arraybackslash}p{0.8cm} >{\centering\arraybackslash}p{1.4cm}>{\centering\arraybackslash}p{1.2cm}>{\centering\arraybackslash}p{1.4cm}>{\centering\arraybackslash}p{1.4cm}>{\centering\arraybackslash}p{0.9cm}}
        \toprule[1.3pt]
        $\lambda_{\textit{causal}}$ & \textbf{VQA-RAD} & \textbf{SLAKE} & \textbf{PathVQA} & \textbf{PMC-VQA} & \textbf{Avg.} \\
        \midrule[1.0pt]
        0.0 & 70.1 & 75.8 & 61.2 & 54.3 & 65.4 \\
        0.3 & 73.4 & 79.0 & 64.1 & 57.8 & 68.6 \\
        \rowcolor{myyellow}
        0.5 & \textbf{74.2} & \textbf{79.8} & \textbf{64.8} & \textbf{58.5} & \textbf{69.3} \\
        0.7 & 73.8 & 79.4 & 64.5 & 58.1 & 69.0 \\
        1.0 & 72.6 & 78.1 & 63.2 & 56.8 & 67.7 \\
        \bottomrule[1.3pt]
        \end{tabular}%
        }
    \end{minipage}
\end{table*}

% ====================================================================
%  §C  Extended Ablation Studies
% ====================================================================
\section{Extended Ablation Studies}
\label{sec:additional_ablation}

Table~\ref{tab:ablation_injection_lambda} and Table~\ref{tab:ablation_div_grpo} present four complementary design analyses that further validate the key design choices of \method. All results use the full three-stage pipeline with IMT inference; default configurations are highlighted.

\textbf{\textit{(i)} Memory injection position.}
Table~\ref{tab:ablation_injection_lambda} (left) examines the effect of injecting diagnostic memory $\mathcal{M}$ at different positions in the input sequence, where $\mathbf{V}$ denotes visual tokens and $\mathbf{q}$ denotes question tokens.
Placing $\mathcal{M}$ after $\mathbf{q}$ and before answer generation (our default) yields the best average of \textbf{69.3\%}, as answer tokens can attend to both visual features and diagnostic memory simultaneously.
Injection before $\mathbf{V}$ degrades performance to 65.5\% because self-attention cannot condition memory on the question context; interleaving with $\mathbf{q}$ (68.2\%) partially recovers but still disrupts the natural query encoding flow.

\textbf{\textit{(ii)} GRPO group size $G$.}
Table~\ref{tab:ablation_injection_lambda} (right) shows that \textit{\textbf{$G{=}4$ achieves the optimal accuracy--cost balance}}: smaller groups ($G{=}2$) yield noisy advantage estimates (67.3\%), while $G{=}6$ and $G{=}8$ provide only marginal gains (+0.1--0.3\,pp) at 1.4--2$\times$ additional GPU hours.
The diminishing returns beyond $G{=}4$ confirm that four trajectories suffice for stable advantage estimation under our composite reward.

\textbf{\textit{(iii)} IMT divergence function.}
Table~\ref{tab:ablation_div_grpo} (left) compares divergence measures for the IMT distillation objective.
Jensen--Shannon divergence with $\beta{=}0.5$ outperforms both forward KL (67.2\%) and reverse KL (66.8\%).
Forward KL causes mode-covering behavior that dilutes diagnostic specificity; reverse KL leads to mode-seeking collapse.
The symmetric JSD provides a balanced learning signal, and performance remains stable across $\beta \in [0.3, 0.7]$.

\textbf{\textit{(iv)} Causal reward weight $\lambda_{\textit{causal}}$.}
Table~\ref{tab:ablation_div_grpo} (right) reveals that performance is robust within $\lambda_{\textit{causal}} \in [0.3, 0.7]$, peaking at \textbf{0.5}.
Setting $\lambda_{\textit{causal}}{=}0$ causes the model to bypass memory via direct shortcuts (65.4\%); excessively high values ($\geq 1.0$) over-penalize trajectories and destabilize training (67.7\%).

\section{Related Works}
\label{sec:related}

\vspace{2mm}
\noindent
\textbf{Latent Computation and Memory Augmented Reasoning.}
Our method is related to latent computation, which leverages continuous latent states to reshape generation in large language models~\cite{geiping2025scaling,deng2024explicit,yu2025vismem,zhang2025memgen}. Existing works include native latent reasoning~\cite{hao2024training,shen2025codi,zhang2025reinforced,tan2025think} and latent regulated generation~\cite{xu2024lars,li2025seek,xu2025softcot,xu2025softcot++,liu2024deliberation}. Memory evolution mechanisms have been explored through progressive experiential compression~\cite{tian2025rgmem}, self evolving visual skill memory~\cite{wang2026atlasva}, compositional concept graph memory~\cite{zhou2026comem}, prompt based and reversible continual learning~\cite{tu2025multiple,10650773,zhu2025ett,xiao2026reversible}, spectral coverage analysis~\cite{wang2026}, rendering aware visual generation~\cite{liang2026render,liang2026vanim}, trajectory anchored memory~\cite{shi2026androtmem}, driver centric latent world models~\cite{chi2026driverwmdrivercentrictrafficconditionedlatent}, hierarchical abductive reasoning~\cite{qiu2026anchorabductivenetworkconstruction}, adaptive state fusion in vision state space models~\cite{ke2026deformbavisionstatespace,ke2025mambev}, temporal point process modeling~\cite{jiang2026danmakutppbench}, knowledge graph reasoning and alignment~\cite{zhang2024question,Li_2025,li2025layerlogitslogicempowering}, psychology based curriculum learning~\cite{meng2025psychology}, and text visual interleaved chain of thought~\cite{hu2026tvi}. Counterfactual explanation for RL agents~\cite{chen2022explain} and causal dynamics of modality arbitration~\cite{zhang2026instruction} motivate our interventional reasoning in latent spaces. Neural signal decoding and reconstruction from EEG~\cite{huang2025ccsumsp,huang2025dual,huang2025mindev,huang2026need} demonstrate how latent spaces bridge perception and cognition. Efficient context utilization via adaptive token pruning~\cite{li2026catp}, input token reduction~\cite{hu2026illava}, correlated prompting for missing modalities~\cite{hu2024deep}, and personalized prompt tuning~\cite{li2026personalize} further inform our design. Our diagnostic implicit memory condenses domain priors into continuous vectors that undergo progressive refinement from external dependency to intrinsic capability.

\vspace{2mm}
\noindent
\textbf{Reinforcement Learning for Vision Language Models.}
RL has become a powerful paradigm for aligning VLMs beyond supervised fine tuning~\cite{schulman2017proximal,rafailov2023direct,ouyang2022training}, with medical applications~\cite{lai2026med,pan2025medvlm,zhu2024mmedpo} and broader tasks including spatial reasoning~\cite{shen2025fine,wang2025monosropenvocabularyspatialreasoning}, cognitive supersensing~\cite{li2026toward}, geometry and expression reasoning~\cite{lin2026synthesizing,lin2026tag}, visual generation~\cite{liang2025multi}, optimization modeling~\cite{liu2026automated}, chart reasoning~\cite{liu2026chartverse}, satirical image comprehension~\cite{jiang2025satiredecoder}, autonomous driving~\cite{Zhang_2026_CVPR,deng2025gaussiandwm3dgaussiandriving,li2026spacedrive}, and GUI reasoning~\cite{li2026whatsmissingscreentoactionuiintheloop}. Post training pipelines combining preference optimization~\cite{zhao2025redone,zhang2026temporal,huang2025adaptive,Huang2026RealTimeAR}, reasoning termination control~\cite{huang2026does}, variation aware entropy scheduling~\cite{wang2026trackingdriftvariationawareentropy}, multi agent collaboration and co evolution~\cite{zhang2026heterogeneous,hu2026seal}, agent benchmarks~\cite{yu2026cirrusbench,guo2025pet,kong2026aiautoresearch}, financial RL and portfolio optimization~\cite{liu2026improving,Cheng2026,song2023deterministic,wang2025gemsllm}, speech LLM training~\cite{yang2025training}, and codebook rebalancing~\cite{fan2026crab} advance alignment. Synthetic data and parameter efficient methods~\cite{liu2024synthvlm,lin2026mmfinereason,liu2025fusion,tao2025autopcr,chen2024can,wei2025identifying} demonstrate data centric approaches. Hallucination mitigation~\cite{jiang2026mm,chen2024detecting,zhang2026mitigating,zhang2025evaluating}, RAG reliability and agent safety~\cite{chen2026doesragknowretrieval,qian2026relevantwarrantedevidenceforcecalibration,wang2026safeskillscollidemeasuring,zhang2026smalllanguagemodelagents,jia-etal-2026-scout}, adversarial robustness and defense~\cite{xu2025one,xu2025clip,xu2026internal,zhang2025dualtapdualtaskadversarialprotector,jiang2026agentic,zhu2026ants,zhu2025knowledge,lin2026reflectguardenhancingllmsafeguards} ensure deployment reliability. For offline RL, federated methods~\cite{qiao2025fova,qiao2026forler}, conservative estimation~\cite{qiao2026less}, collapse suppressed optimization~\cite{qiao2026adamo}, and parameter efficient merging~\cite{du2025graftllm,du2024pcb,du2026dynamic} address training stability. Our CCR builds upon GRPO~\cite{shao2024deepseekmath} with a causal counterfactual reward distinguishing memory utilization from shortcuts.

\vspace{2mm}
\noindent
\textbf{Medical Image Understanding and Efficient Deployment.}
The anatomical encoder in \method derives spatial priors from large scale segmentation pretraining. Annotation efficient and federated medical segmentation~\cite{zhao2024ultrasound,zhao2024sam,zhou2024reducing,zhou2025isosnet,zhao2026divide}, hypergraph based pathological detection~\cite{li2025high}, prototype guided interactive segmentation~\cite{ge2025progis}, curated pathology datasets~\cite{wang2025fully}, similarity aware event prediction~\cite{li2025similarity}, pathology aware multicenter diagnosis~\cite{zhu2025pathology}, progressive diagnosis~\cite{zhu2026medeyes}, adaptive causal reasoning for medical VLMs~\cite{lin2026medcausalx}, histopathological clustering and contrastive learning~\cite{li2022darc,li20252,li2025mico,li2026universal,li2026domain}, topological contrastive learning~\cite{meng2026topocl}, 3D medical segmentation reasoning~\cite{zhang2026seer}, LLM guided few shot segmentation~\cite{xiao6281833focus}, weather aware reasoning segmentation~\cite{du2026weatherreasonseg}, and electron micrograph understanding~\cite{xia2025uniem} advance medical image analysis. Medical reasoning and chain of thought methods~\cite{wang2025v2t,jiang2025comt,jiang2026multi,jiang2025hulu,wu2026better,yang2025med,yang2025qm}, clinical cognition alignment~\cite{zheng2026clinical}, interpretable dermoscopy diagnosis~\cite{jia2026geodesic,jia2026looks,jia2026unsupervised}, knowledge enhanced reasoning~\cite{hu2026bridging,han-etal-2025-black}, domain adaptive healthcare~\cite{hu2025udoncare,hu2026exploring}, temporal mixture of experts~\cite{zhang2026adaptive}, heterogeneous graph learning~\cite{zhang2026heteromile,he2022webmile}, bias mitigation~\cite{salarian2025medequalizer}, auditory attention decoding~\cite{huang2025ssaad}, traffic forecasting and optimal transport~\cite{meng2022early,meng2025efficient,chambers2025stable,meng2026ms}, and clinical workflow analysis~\cite{xia2025association,lew2025association,lou2024secure} collectively advance health informatics. On the deployment side, speculative decoding~\cite{wei2025specasr}, native parallel reading~\cite{wang2026fbsmodelingnativeparallel}, diffusion acceleration~\cite{wei2025orchestrating,wei2026team,zhao2026resilphaseplugandplayphasemapping}, ultra low bit quantization~\cite{zhao2026specquant,zhao2026bwla}, neural parameter search~\cite{du2025nps}, VLM NPU co design~\cite{chen2025autoneuralcodesigningvisionlanguagemodels}, and quantized split learning~\cite{guo2025quantized} validate that not all computation demands equal investment, a principle our 16 memory vectors embody by replacing hundreds of reasoning tokens.

\bibliographystyle{splncs04}
\bibliography{main_525}

@article{lau2018dataset,
  title={A dataset of clinically generated visual questions and answers about radiology images},
  author={Lau, Jason J and Gayen, Soumya and Ben Abacha, Asma and Demner-Fushman, Dina},
  journal={Scientific data},
  volume={5},
  number={1},
  pages={180251},
  year={2018},
  publisher={Nature Publishing Group}
}

@inproceedings{liu2021slake,
  title={Slake: A semantically-labeled knowledge-enhanced dataset for medical visual question answering},
  author={Liu, Bo and Zhan, Li-Ming and Xu, Li and Ma, Lin and Yang, Yan and Wu, Xiao-Ming},
  booktitle={2021 IEEE 18th international symposium on biomedical imaging (ISBI)},
  pages={1650--1654},
  year={2021},
  organization={IEEE}
}

@article{he2020pathvqa,
  title={Pathvqa: 30000+ questions for medical visual question answering},
  author={He, Xuehai and Zhang, Yichen and Mou, Luntian and Xing, Eric and Xie, Pengtao},
  journal={arXiv preprint arXiv:2003.10286},
  year={2020}
}

@article{zhang2023pmc,
  title={Pmc-vqa: Visual instruction tuning for medical visual question answering},
  author={Zhang, Xiaoman and Wu, Chaoyi and Zhao, Ziheng and Lin, Weixiong and Zhang, Ya and Wang, Yanfeng and Xie, Weidi},
  journal={arXiv preprint arXiv:2305.10415},
  year={2023}
}

@inproceedings{yue2024mmmu,
  title={Mmmu: A massive multi-discipline multimodal understanding and reasoning benchmark for expert agi},
  author={Yue, Xiang and Ni, Yuansheng and Zhang, Kai and Zheng, Tianyu and Liu, Ruoqi and Zhang, Ge and Stevens, Samuel and Jiang, Dongfu and Ren, Weiming and Sun, Yuxuan and others},
  booktitle={Proceedings of the IEEE/CVF conference on computer vision and pattern recognition},
  pages={9556--9567},
  year={2024}
}

@article{zuo2025medxpertqa,
  title={Medxpertqa: Benchmarking expert-level medical reasoning and understanding},
  author={Zuo, Yuxin and Qu, Shang and Li, Yifei and Chen, Zhangren and Zhu, Xuekai and Hua, Ermo and Zhang, Kaiyan and Ding, Ning and Zhou, Bowen},
  journal={arXiv preprint arXiv:2501.18362},
  year={2025}
}

@article{wu2025towards,
  title={Towards generalist foundation model for radiology by leveraging web-scale 2d\&3d medical data},
  author={Wu, Chaoyi and Zhang, Xiaoman and Zhang, Ya and Hui, Hui and Wang, Yanfeng and Xie, Weidi},
  journal={Nature Communications},
  volume={16},
  number={1},
  pages={7866},
  year={2025},
  publisher={Nature Publishing Group UK London}
}

@inproceedings{hu2024omnimedvqa,
  title={Omnimedvqa: A new large-scale comprehensive evaluation benchmark for medical lvlm},
  author={Hu, Yutao and Li, Tianbin and Lu, Quanfeng and Shao, Wenqi and He, Junjun and Qiao, Yu and Luo, Ping},
  booktitle={Proceedings of the IEEE/CVF Conference on Computer Vision and Pattern Recognition},
  pages={22170--22183},
  year={2024}
}

@article{ding2026mmedexpert,
  title={MMedExpert-R1: Strengthening Multimodal Medical Reasoning via Domain-Specific Adaptation and Clinical Guideline Reinforcement},
  author={Ding, Meidan and Zhang, Jipeng and Wang, Wenxuan and Zhong, Haiqin and Luo, Xiaoling and Chen, Wenting and Shen, Linlin},
  journal={arXiv preprint arXiv:2601.10949},
  year={2026}
}

@inproceedings{chen2024towards,
  title={Towards injecting medical visual knowledge into multimodal llms at scale},
  author={Chen, Junying and Gui, Chi and Ouyang, Ruyi and Gao, Anningzhe and Chen, Shunian and Chen, Guiming Hardy and Wang, Xidong and Cai, Zhenyang and Ji, Ke and Wan, Xiang and others},
  booktitle={Proceedings of the 2024 conference on empirical methods in natural language processing},
  pages={7346--7370},
  year={2024}
}

@inproceedings{pan2025medvlm,
  title={Medvlm-r1: Incentivizing medical reasoning capability of vision-language models (vlms) via reinforcement learning},
  author={Pan, Jiazhen and Liu, Che and Wu, Junde and Liu, Fenglin and Zhu, Jiayuan and Li, Hongwei Bran and Chen, Chen and Ouyang, Cheng and Rueckert, Daniel},
  booktitle={International Conference on Medical Image Computing and Computer-Assisted Intervention},
  pages={337--347},
  year={2025},
  organization={Springer}
}

@article{lai2026med,
  title={Med-r1: Reinforcement learning for generalizable medical reasoning in vision-language models},
  author={Lai, Yuxiang and Zhong, Jike and Li, Ming and Zhao, Shitian and Li, Yuheng and Psounis, Konstantinos and Yang, Xiaofeng},
  journal={IEEE transactions on medical imaging},
  year={2026},
  publisher={IEEE}
}

@article{mullappilly2026medix,
  title={Medix-r1: Open ended medical reinforcement learning},
  author={Mullappilly, Sahal Shaji and Kurpath, Mohammed Irfan and Mohamed, Omair and Zidan, Mohamed and Khan, Fahad and Khan, Salman and Anwer, Rao and Cholakkal, Hisham},
  journal={arXiv preprint arXiv:2602.23363},
  year={2026}
}

@article{hao2024training,
  title={Training large language models to reason in a continuous latent space},
  author={Hao, Shibo and Sukhbaatar, Sainbayar and Su, DiJia and Li, Xian and Hu, Zhiting and Weston, Jason and Tian, Yuandong},
  journal={arXiv preprint arXiv:2412.06769},
  year={2024}
}

@inproceedings{shen2025codi,
  title={Codi: Compressing chain-of-thought into continuous space via self-distillation},
  author={Shen, Zhenyi and Yan, Hanqi and Zhang, Linhai and Hu, Zhanghao and Du, Yali and He, Yulan},
  booktitle={Proceedings of the 2025 Conference on Empirical Methods in Natural Language Processing},
  pages={677--693},
  year={2025}
}

@article{zhang2025reinforced,
  title={Reinforced latent reasoning for llm-based recommendation},
  author={Zhang, Yang and Xu, Wenxin and Zhao, Xiaoyan and Wang, Wenjie and Feng, Fuli and He, Xiangnan and Chua, Tat-Seng},
  journal={arXiv preprint arXiv:2505.19092},
  year={2025}
}

@article{li2025seek,
  title={Seek in the dark: Reasoning via test-time instance-level policy gradient in latent space},
  author={Li, Hengli and Li, Chenxi and Wu, Tong and Zhu, Xuekai and Wang, Yuxuan and Yu, Zhaoxin and Jiang, Eric Hanchen and Zhu, Song-Chun and Jia, Zixia and Wu, Ying Nian and others},
  journal={arXiv preprint arXiv:2505.13308},
  year={2025}
}

@inproceedings{xu2025softcot,
  title={Softcot: Soft chain-of-thought for efficient reasoning with llms},
  author={Xu, Yige and Guo, Xu and Zeng, Zhiwei and Miao, Chunyan},
  booktitle={Proceedings of the 63rd Annual Meeting of the Association for Computational Linguistics (Volume 1: Long Papers)},
  pages={23336--23351},
  year={2025}
}

@article{xu2025softcot++,
  title={Softcot++: Test-time scaling with soft chain-of-thought reasoning},
  author={Xu, Yige and Guo, Xu and Zeng, Zhiwei and Miao, Chunyan},
  journal={arXiv preprint arXiv:2505.11484},
  year={2025}
}

@article{shao2024deepseekmath,
  title={Deepseekmath: Pushing the limits of mathematical reasoning in open language models},
  author={Shao, Zhihong and Wang, Peiyi and Zhu, Qihao and Xu, Runxin and Song, Junxiao and Bi, Xiao and Zhang, Haowei and Zhang, Mingchuan and Li, YK and Wu, Yang and others},
  journal={arXiv preprint arXiv:2402.03300},
  year={2024}
}

@article{schulman2017proximal,
  title={Proximal policy optimization algorithms},
  author={Schulman, John and Wolski, Filip and Dhariwal, Prafulla and Radford, Alec and Klimov, Oleg},
  journal={arXiv preprint arXiv:1707.06347},
  year={2017}
}

@article{rafailov2023direct,
  title={Direct preference optimization: Your language model is secretly a reward model},
  author={Rafailov, Rafael and Sharma, Archit and Mitchell, Eric and Manning, Christopher D and Ermon, Stefano and Finn, Chelsea},
  journal={Advances in neural information processing systems},
  volume={36},
  pages={53728--53741},
  year={2023}
}

@article{ouyang2022training,
  title={Training language models to follow instructions with human feedback},
  author={Ouyang, Long and Wu, Jeffrey and Jiang, Xu and Almeida, Diogo and Wainwright, Carroll and Mishkin, Pamela and Zhang, Chong and Agarwal, Sandhini and Slama, Katarina and Ray, Alex and others},
  journal={Advances in neural information processing systems},
  volume={35},
  pages={27730--27744},
  year={2022}
}

@article{zhu2024mmedpo,
  title={Mmedpo: Aligning medical vision-language models with clinical-aware multimodal preference optimization},
  author={Zhu, Kangyu and Xia, Peng and Li, Yun and Zhu, Hongtu and Wang, Sheng and Yao, Huaxiu},
  journal={arXiv preprint arXiv:2412.06141},
  year={2024}
}

@inproceedings{wang2025v2t,
  title={V2t-cot: From vision to text chain-of-thought for medical reasoning and diagnosis},
  author={Wang, Yuan and Liu, Jiaxiang and Gao, Shujian and Feng, Bin and Tang, Zhihang and Gai, Xiaotang and Wu, Jian and Liu, Zuozhu},
  booktitle={International Conference on Medical Image Computing and Computer-Assisted Intervention},
  pages={658--668},
  year={2025},
  organization={Springer}
}

@article{deria2026medmo,
  title={MedMO: Grounding and Understanding Multimodal Large Language Model for Medical Images},
  author={Deria, Ankan and Kumar, Komal and Dukre, Adinath Madhavrao and Segal, Eran and Khan, Salman and Razzak, Imran},
  journal={arXiv preprint arXiv:2602.06965},
  year={2026}
}

@article{chen2025reasoning,
  title={Reasoning in the dark: Interleaved vision-text reasoning in latent space},
  author={Chen, Chao and Ma, Zhixin and Li, Yongqi and Hu, Yupeng and Wei, Yinwei and Li, Wenjie and Nie, Liqiang},
  journal={arXiv preprint arXiv:2510.12603},
  year={2025}
}

@article{cheng2023sam,
  title={Sam-med2d},
  author={Cheng, Junlong and Ye, Jin and Deng, Zhongying and Chen, Jianpin and Li, Tianbin and Wang, Haoyu and Su, Yanzhou and Huang, Ziyan and Chen, Jilong and Jiang, Lei and others},
  journal={arXiv preprint arXiv:2308.16184},
  year={2023}
}

@article{liu2024deliberation,
  title={Deliberation in latent space via differentiable cache augmentation},
  author={Liu, Luyang and Pfeiffer, Jonas and Wu, Jiaxing and Xie, Jun and Szlam, Arthur},
  journal={arXiv preprint arXiv:2412.17747},
  year={2024}
}

@article{lin2002divergence,
  title={Divergence measures based on the Shannon entropy},
  author={Lin, Jianhua},
  journal={IEEE Transactions on Information theory},
  volume={37},
  number={1},
  pages={145--151},
  year={2002},
  publisher={IEEE}
}

@article{ye2024gmai,
  title={Gmai-mmbench: A comprehensive multimodal evaluation benchmark towards general medical ai},
  author={Ye, Jin and Wang, Guoan and Li, Yanjun and Deng, Zhongying and Li, Wei and Li, Tianbin and Duan, Haodong and Huang, Ziyan and Su, Yanzhou and Wang, Benyou and others},
  journal={Advances in Neural Information Processing Systems},
  volume={37},
  pages={94327--94427},
  year={2024}
}

@article{brunye2019review,
  title={A review of eye tracking for understanding and improving diagnostic interpretation},
  author={Bruny{\'e}, Tad T and Drew, Trafton and Weaver, Donald L and Elmore, Joann G},
  journal={Cognitive research: principles and implications},
  volume={4},
  number={1},
  pages={7},
  year={2019},
  publisher={Springer}
}

@article{li2023llava,
  title={Llava-med: Training a large language-and-vision assistant for biomedicine in one day},
  author={Li, Chunyuan and Wong, Cliff and Zhang, Sheng and Usuyama, Naoto and Liu, Haotian and Yang, Jianwei and Naumann, Tristan and Poon, Hoifung and Gao, Jianfeng},
  journal={Advances in Neural Information Processing Systems},
  volume={36},
  pages={28541--28564},
  year={2023}
}

@article{le2025s,
  title={S-chain: Structured visual chain-of-thought for medicine},
  author={Le-Duc, Khai and Nguyen, Duy MH and Trinh, Phuong TH and Nguyen, Tien-Phat and Diep, Nghiem T and Ngo, An and Vu, Tung and Vuong, Trinh and Nguyen, Anh-Tien and Nguyen, Mau and others},
  journal={arXiv preprint arXiv:2510.22728},
  year={2025}
}

@article{norman2009dual,
  title={Dual processing and diagnostic errors},
  author={Norman, Geoff},
  journal={Advances in Health Sciences Education},
  volume={14},
  number={Suppl 1},
  pages={37--49},
  year={2009},
  publisher={Springer}
}

@inproceedings{moor2023med,
  title={Med-flamingo: a multimodal medical few-shot learner},
  author={Moor, Michael and Huang, Qian and Wu, Shirley and Yasunaga, Michihiro and Dalmia, Yash and Leskovec, Jure and Zakka, Cyril and Reis, Eduardo Pontes and Rajpurkar, Pranav},
  booktitle={Machine learning for health (ML4H)},
  pages={353--367},
  year={2023},
  organization={PMLR}
}

@article{chen2024detecting,
  title={Detecting and evaluating medical hallucinations in large vision language models},
  author={Chen, Jiawei and Yang, Dingkang and Wu, Tong and Jiang, Yue and Hou, Xiaolu and Li, Mingcheng and Wang, Shunli and Xiao, Dongling and Li, Ke and Zhang, Lihua},
  journal={arXiv preprint arXiv:2406.10185},
  year={2024}
}

@inproceedings{xu2024lars,
  title={LaRS: Latent reasoning skills for chain-of-thought reasoning},
  author={Xu, Zifan and Wang, Haozhu and Bespalov, Dmitriy and Wu, Xian and Stone, Peter and Qi, Yanjun},
  booktitle={Findings of the Association for Computational Linguistics: EMNLP 2024},
  pages={3624--3643},
  year={2024}
}

@article{pham2025multimodal,
  title={Multimodal chain of continuous thought for latent-space reasoning in vision-language models},
  author={Pham, Tan-Hanh and Ngo, Chris},
  journal={arXiv preprint arXiv:2508.12587},
  year={2025}
}

@article{yu2025vismem,
  title={Vismem: Latent vision memory unlocks potential of vision-language models},
  author={Yu, Xinlei and Xu, Chengming and Zhang, Guibin and Chen, Zhangquan and Zhang, Yudong and He, Yongbo and Jiang, Peng-Tao and Zhang, Jiangning and Hu, Xiaobin and Yan, Shuicheng},
  journal={arXiv preprint arXiv:2511.11007},
  year={2025}
}

@article{zhang2025memgen,
  title={Memgen: Weaving generative latent memory for self-evolving agents},
  author={Zhang, Guibin and Fu, Muxin and Yan, Shuicheng},
  journal={arXiv preprint arXiv:2509.24704},
  year={2025}
}

@article{sellergren2025medgemma,
  title={Medgemma technical report},
  author={Sellergren, Andrew and Kazemzadeh, Sahar and Jaroensri, Tiam and Kiraly, Atilla and Traverse, Madeleine and Kohlberger, Timo and Xu, Shawn and Jamil, Fayaz and Hughes, C{\'\i}an and Lau, Charles and others},
  journal={arXiv preprint arXiv:2507.05201},
  year={2025}
}

@article{chen2025think,
  title={Think twice to see more: Iterative visual reasoning in medical vlms},
  author={Chen, Kaitao and Rui, Shaohao and Jiang, Yankai and Wu, Jiamin and Zheng, Qihao and Song, Chunfeng and Wang, Xiaosong and Zhou, Mu and Liu, Mianxin},
  journal={arXiv preprint arXiv:2510.10052},
  year={2025}
}

@inproceedings{gu2024lapa,
  title={Lapa: Latent prompt assist model for medical visual question answering},
  author={Gu, Tiancheng and Yang, Kaicheng and Liu, Dongnan and Cai, Weidong},
  booktitle={Proceedings of the IEEE/CVF Conference on Computer Vision and Pattern Recognition},
  pages={4971--4980},
  year={2024}
}

@article{lewis2020retrieval,
  title={Retrieval-augmented generation for knowledge-intensive nlp tasks},
  author={Lewis, Patrick and Perez, Ethan and Piktus, Aleksandra and Petroni, Fabio and Karpukhin, Vladimir and Goyal, Naman and K{\"u}ttler, Heinrich and Lewis, Mike and Yih, Wen-tau and Rockt{\"a}schel, Tim and others},
  journal={Advances in neural information processing systems},
  volume={33},
  pages={9459--9474},
  year={2020}
}

@inproceedings{van2023open,
  title={Open-ended medical visual question answering through prefix tuning of language models},
  author={Van Sonsbeek, Tom and Derakhshani, Mohammad Mahdi and Najdenkoska, Ivona and Snoek, Cees GM and Worring, Marcel},
  booktitle={International Conference on Medical Image Computing and Computer-Assisted Intervention},
  pages={726--736},
  year={2023},
  organization={Springer}
}

@article{zakka2024almanac,
  title={Almanac—retrieval-augmented language models for clinical medicine},
  author={Zakka, Cyril and Shad, Rohan and Chaurasia, Akash and Dalal, Alex R and Kim, Jennifer L and Moor, Michael and Fong, Robyn and Phillips, Curran and Alexander, Kevin and Ashley, Euan and others},
  journal={Nejm ai},
  volume={1},
  number={2},
  pages={AIoa2300068},
  year={2024},
  publisher={Massachusetts Medical Society}
}

@article{zhang2025patho,
  title={Patho-AgenticRAG: towards multimodal agentic retrieval-augmented generation for pathology VLMs via reinforcement learning},
  author={Zhang, Wenchuan and Guo, Jingru and Zhang, Hengzhe and Zhang, Penghao and Chen, Jie and Zhang, Shuwan and Zhang, Zhang and Yi, Yuhao and Bu, Hong},
  journal={arXiv preprint arXiv:2508.02258},
  year={2025}
}

@article{arasteh2407radiorag,
  title={RadioRAG: factual large language models for enhanced diagnostics in radiology using online retrieval augmented generation 2024},
  author={Arasteh, Soroosh Tayebi and Lotfinia, Mahshad and Bressem, Keno and Siepmann, Robert and Adams, Lisa and Ferber, Dyke and Kuhl, Christiane and Kather, Jakob Nikolas and Nebelung, Sven and Truhn, Daniel},
  journal={arXiv preprint arXiv.2407.15621}
}

@article{wu2024medical,
  title={Medical graph rag: Towards safe medical large language model via graph retrieval-augmented generation},
  author={Wu, Junde and Zhu, Jiayuan and Qi, Yunli and Chen, Jingkun and Xu, Min and Menolascina, Filippo and Grau, Vicente},
  journal={arXiv preprint arXiv:2408.04187},
  year={2024}
}

@inproceedings{zhao2025medrag,
  title={Medrag: Enhancing retrieval-augmented generation with knowledge graph-elicited reasoning for healthcare copilot},
  author={Zhao, Xuejiao and Liu, Siyan and Yang, Su-Yin and Miao, Chunyan},
  booktitle={Proceedings of the ACM on Web Conference 2025},
  pages={4442--4457},
  year={2025}
}

@article{hu2022lora,
  title={Lora: Low-rank adaptation of large language models.},
  author={Hu, Edward J and Shen, Yelong and Wallis, Phillip and Allen-Zhu, Zeyuan and Li, Yuanzhi and Wang, Shean and Wang, Liang and Chen, Weizhu and others},
  journal={Iclr},
  volume={1},
  number={2},
  pages={3},
  year={2022}
}

@article{waite2017interpretive,
  title={Interpretive error in radiology},
  author={Waite, Stephen and Scott, Jinel and Gale, Brian and Fuchs, Travis and Kolla, Srinivas and Reede, Deborah},
  journal={American Journal of Roentgenology},
  volume={208},
  number={4},
  pages={739--749},
  year={2017},
  publisher={American Roentgen Ray Society}
}

@article{yu2025finemedlm,
  title={Finemedlm-o1: Enhancing the medical reasoning ability of llm from supervised fine-tuning to test-time training},
  author={Yu, Hongzhou and Cheng, Tianhao and Cheng, Ying and Feng, Rui},
  journal={arXiv e-prints},
  pages={arXiv--2501},
  year={2025}
}

@article{geiping2025scaling,
  title={Scaling up test-time compute with latent reasoning: A recurrent depth approach},
  author={Geiping, Jonas and McLeish, Sean and Jain, Neel and Kirchenbauer, John and Singh, Siddharth and Bartoldson, Brian R and Kailkhura, Bhavya and Bhatele, Abhinav and Goldstein, Tom},
  journal={arXiv preprint arXiv:2502.05171},
  year={2025}
}

@article{deng2024explicit,
  title={From explicit cot to implicit cot: Learning to internalize cot step by step},
  author={Deng, Yuntian and Choi, Yejin and Shieber, Stuart},
  journal={arXiv preprint arXiv:2405.14838},
  year={2024}
}

@article{su2025gmai,
  title={Gmai-vl-r1: Harnessing reinforcement learning for multimodal medical reasoning},
  author={Su, Yanzhou and Li, Tianbin and Liu, Jiyao and Ma, Chenglong and Ning, Junzhi and Tang, Cheng and Ju, Sibo and Ye, Jin and Chen, Pengcheng and Hu, Ming and others},
  journal={arXiv preprint arXiv:2504.01886},
  year={2025}
}

@article{liu2025medsam3,
  title={MedSAM3: Delving into Segment Anything with Medical Concepts},
  author={Liu, Anglin and Xue, Rundong and Cao, Xu R and Shen, Yifan and Lu, Yi and Li, Xiang and Chen, Qianqian and Chen, Jintai},
  journal={arXiv preprint arXiv:2511.19046},
  year={2025}
}

@article{tan2025think,
  title={Think silently, think fast: Dynamic latent compression of llm reasoning chains},
  author={Tan, Wenhui and Li, Jiaze and Ju, Jianzhong and Luo, Zhenbo and Song, Ruihua and Luan, Jian},
  journal={arXiv preprint arXiv:2505.16552},
  year={2025}
}

@article{wei2022chain,
  title={Chain-of-thought prompting elicits reasoning in large language models},
  author={Wei, Jason and Wang, Xuezhi and Schuurmans, Dale and Bosma, Maarten and Xia, Fei and Chi, Ed and Le, Quoc V and Zhou, Denny and others},
  journal={Advances in neural information processing systems},
  volume={35},
  pages={24824--24837},
  year={2022}
}

@article{gai2024medthink,
  title={Medthink: Explaining medical visual question answering via multimodal decision-making rationale},
  author={Gai, Xiaotang and Zhou, Chenyi and Liu, Jiaxiang and Feng, Yang and Wu, Jian and Liu, Zuozhu},
  journal={arXiv preprint arXiv:2404.12372},
  year={2024}
}

@article{wu2025medreason,
  title={Medreason: Eliciting factual medical reasoning steps in llms via knowledge graphs},
  author={Wu, Juncheng and Deng, Wenlong and Li, Xingxuan and Liu, Sheng and Mi, Taomian and Peng, Yifan and Xu, Ziyang and Liu, Yi and Cho, Hyunjin and Choi, Chang-In and others},
  journal={arXiv preprint arXiv:2504.00993},
  year={2025}
}

@article{sun2025chiron,
  title={Chiron-o1: Igniting multimodal large language models towards generalizable medical reasoning via mentor-intern collaborative search},
  author={Sun, Haoran and Jiang, Yankai and Lou, Wenjie and Zhang, Yujie and Li, Wenjie and Wang, Lilong and Liu, Mianxin and Liu, Lei and Wang, Xiaosong},
  journal={arXiv preprint arXiv:2506.16962},
  year={2025}
}

@article{li2024gmai,
  title={Gmai-vl \& gmai-vl-5.5 m: A large vision-language model and a comprehensive multimodal dataset towards general medical ai},
  author={Li, Tianbin and Su, Yanzhou and Li, Wei and Fu, Bin and Chen, Zhe and Huang, Ziyan and Wang, Guoan and Ma, Chenglong and Chen, Ying and Hu, Ming and others},
  journal={arXiv preprint arXiv:2411.14522},
  year={2024}
}

@article{mullappilly2024bimedix2,
  title={Bimedix2: Bio-medical expert lmm for diverse medical modalities},
  author={Mullappilly, Sahal Shaji and Kurpath, Mohammed Irfan and Pieri, Sara and Alseiari, Saeed Yahya and Cholakkal, Shanavas and Aldahmani, Khaled and Khan, Fahad and Anwer, Rao and Khan, Salman and Baldwin, Timothy and others},
  journal={arXiv preprint arXiv:2412.07769},
  year={2024}
}

@article{bai2025qwen3,
  title={Qwen3-vl technical report},
  author={Bai, Shuai and Cai, Yuxuan and Chen, Ruizhe and Chen, Keqin and Chen, Xionghui and Cheng, Zesen and Deng, Lianghao and Ding, Wei and Gao, Chang and Ge, Chunjiang and others},
  journal={arXiv preprint arXiv:2511.21631},
  year={2025}
}

@article{zhu2025internvl3,
  title={Internvl3: Exploring advanced training and test-time recipes for open-source multimodal models},
  author={Zhu, Jinguo and Wang, Weiyun and Chen, Zhe and Liu, Zhaoyang and Ye, Shenglong and Gu, Lixin and Tian, Hao and Duan, Yuchen and Su, Weijie and Shao, Jie and others},
  journal={arXiv preprint arXiv:2504.10479},
  year={2025}
}

@article{zhao2024ultrasound,
  title={Ultrasound nodule segmentation using asymmetric learning with simple clinical annotation},
  author={Zhao, Xingyue and Li, Zhongyu and Luo, Xiangde and Li, Peiqi and Huang, Peng and Zhu, Jianwei and Liu, Yang and Zhu, Jihua and Yang, Meng and Chang, Shi and others},
  journal={IEEE Transactions on Circuits and Systems for Video Technology},
  volume={34},
  number={10},
  pages={9010--9023},
  year={2024},
  publisher={IEEE}
}

@inproceedings{zhao2024sam,
  title={Sam-Driven Weakly Supervised Nodule Segmentation with Uncertainty-Aware Cross Teaching},
  author={Zhao, Xingyue and Li, Peiqi and Luo, Xiangde and Yang, Meng and Chang, Shi and Li, Zhongyu},
  booktitle={2024 IEEE International Symposium on Biomedical Imaging (ISBI)},
  pages={1--5},
  year={2024},
  organization={IEEE}
}

@inproceedings{zhao2026divide,
  title={Divide, Conquer and Unite: Hierarchical Style-Recalibrated Prototype Alignment for Federated Medical Segmentation},
  author={Zhao, Xingyue and Huang, Wenke and Wang, Xingguang and Zhao, Haoyu and Zhuang, Linghao and Jiang, Anwen and Wan, Guancheng and Ye, Mang},
  booktitle={Proceedings of the AAAI Conference on Artificial Intelligence},
  volume={40},
  number={34},
  pages={28760--28768},
  year={2026}
}

@article{shen2025fine,
  title={Fine-grained preference optimization improves spatial reasoning in vlms},
  author={Shen, Yifan and Liu, Yuanzhe and Zhu, Jingyuan and Cao, Xu and Zhang, Xiaofeng and He, Yixiao and Ye, Wenming and Rehg, James Matthew and Lourentzou, Ismini},
  journal={arXiv preprint arXiv:2506.21656},
  year={2025}
}

@article{li2026toward,
  title={Toward Cognitive Supersensing in Multimodal Large Language Model},
  author={Li, Boyi and Shen, Yifan and Liu, Yuanzhe and Xu, Yifan and Liu, Jiateng and Li, Xinzhuo and Li, Zhengyuan and Zhu, Jingyuan and Zhong, Yunhan and Lan, Fangzhou and others},
  journal={arXiv preprint arXiv:2602.01541},
  year={2026}
}

@article{qiao2025fova,
  author={Qiao, Nan and Yue, Sheng and Ren, Ju and Zhang, Yaoxue},
  journal={IEEE Transactions on Networking}, 
  title={FOVA: Offline Federated Reinforcement Learning With Mixed-Quality Data}, 
  year={2026},
  volume={34},
  pages={2031-2046},
  doi={10.1109/TON.2025.3637043}
}

@inproceedings{qiao2026less,
  title={Less Is More: Clustered Cross-Covariance Control for Offline {RL}},
  author={Nan Qiao and Sheng Yue and Shuning Wang and Yongheng Deng and Ju Ren},
  booktitle={The Fourteenth International Conference on Learning Representations},
  year={2026},
  url={https://openreview.net/forum?id=drOy5wi6Qq}
}

@misc{qiao2026forler,
  title={FORLER: Federated Offline Reinforcement Learning with Q-Ensemble and Actor Rectification}, 
  author={Nan Qiao and Sheng Yue},
  year={2026},
  eprint={2602.02055},
  archivePrefix={arXiv},
  primaryClass={cs.LG}
}

@misc{qiao2026adamo,
  title={AdamO: A Collapse-Suppressed Optimizer for Offline RL}, 
  author={Nan Qiao and Sheng Yue and Shuning Wang and Ju Ren},
  year={2026},
  eprint={2605.01968},
  archivePrefix={arXiv},
  primaryClass={cs.LG}
}

@article{tian2025rgmem,
  title={Rgmem: Renormalization group-based memory evolution for language agent user profile},
  author={Tian, Ao and Lu, Yunfeng and Fan, Xinxin and Wang, Changhao and Zhou, Lanzhi and Zhang, Yeyao and Liu, Yanfang},
  journal={arXiv preprint arXiv:2510.16392},
  year={2025}
}

@inproceedings{wei2025specasr,
  title={SpecASR: Accelerating LLM-based Automatic Speech Recognition via Speculative Decoding},
  author={Wei, Linye and Zhong, Shuzhang and Xu, Songqiang and Wang, Runsheng and Huang, Ru and Li, Meng},
  booktitle={2025 62nd ACM/IEEE Design Automation Conference (DAC)},
  pages={1--7},
  year={2025},
  organization={IEEE}
}

@article{wei2025orchestrating,
  title={Orchestrating Dual-Boundaries: An Arithmetic Intensity Inspired Acceleration Framework for Diffusion Language Models},
  author={Wei, Linye and Chen, Wenjue and Tang, Pingzhi and Guo, Xiaotian and Ye, Le and Wang, Runsheng and Li, Meng},
  journal={arXiv preprint arXiv:2511.21759},
  year={2025}
}

@article{wei2026team,
  title={TEAM: Temporal-Spatial Consistency Guided Expert Activation for MoE Diffusion Language Model Acceleration},
  author={Wei, Linye and Luo, Zixiang and Tang, Pingzhi and Li, Meng},
  journal={arXiv preprint arXiv:2602.08404},
  year={2026}
}

@inproceedings{10650773,
  author={Yi, Huiyu},
  booktitle={2024 International Joint Conference on Neural Networks (IJCNN)}, 
  title={Few-Shot Class-Incremental Learning with Class Centers and Contrastive Learning for Incremental Vehicle Recognition}, 
  year={2024},
  pages={1-8}
}

@inproceedings{tu2025multiple,
  title={Multiple Queries with Multiple Keys: A Precise Prompt Matching Paradigm for Prompt-based Continual Learning},
  author={Tu, Dunwei and Yi, Huiyu and Wang, Yuchi and Xu, Baile and Zhao, Jian and Shen, Furao},
  booktitle={Proceedings of the 33rd ACM International Conference on Multimedia},
  pages={372--381},
  year={2025}
}

@article{lin2026synthesizing,
  title={Synthesizing Multimodal Geometry Datasets from Scratch and Enabling Visual Alignment via Plotting Code},
  author={Lin, Haobo and Bai, Tianyi and Chen, Chen and Zhang, Jiajun and Zeng, Bohan and Zhang, Wentao and Yuan, Binhang},
  journal={arXiv preprint arXiv:2602.18745},
  year={2026}
}

@article{lin2026tag,
  title={TAG: Thinking with Action Unit Grounding for Facial Expression Recognition},
  author={Lin, Haobo and Bai, Tianyi and Zhang, Jiajun and Chang, Xuanhao and Lu, Sheng and Gu, Fangming and Hu, Zengjie and Zhang, Wentao},
  journal={arXiv preprint arXiv:2602.18763},
  year={2026}
}

@inproceedings{zhao2026specquant,
  title={Specquant: Spectral decomposition and adaptive truncation for ultra-low-bit llms quantization},
  author={Zhao, Zhixiong and Liu, Fangxin and Wang, Junjie and Guan, Chenyang and Wang, Zongwu and Jiang, Li and Guan, Haibing},
  booktitle={Proceedings of the AAAI Conference on Artificial Intelligence},
  volume={40},
  number={34},
  pages={28786--28794},
  year={2026}
}

@misc{zhao2026bwla,
  title={BWLA: Breaking the Barrier of W1AX Post-Training Quantization for LLMs}, 
  author={Zhixiong Zhao and Zukang Xu and Dawei Yang},
  year={2026},
  eprint={2605.00422},
  archivePrefix={arXiv},
  primaryClass={cs.LG},
  url={https://arxiv.org/abs/2605.00422}
}

@article{liang2026render,
  title={Render-in-the-Loop: Vector Graphics Generation via Visual Self-Feedback},
  author={Liang, Guotao and Wang, Zhangcheng and Hu, Juncheng and Zhou, Haitao and Xue, Ziteng and Zhang, Jing and Xu, Dong and Yu, Qian},
  journal={arXiv preprint arXiv:2604.20730},
  year={2026}
}

@article{liang2026vanim,
  title={VAnim: Rendering-Aware Sparse State Modeling for Structure-Preserving Vector Animation},
  author={Liang, Guotao and Wang, Zhangcheng and Wang, Chuang and Hu, Juncheng and Zhou, Haitao and Liu, Junhua and Zhang, Jing and Xu, Dong and Yu, Qian},
  journal={arXiv preprint arXiv:2605.01517},
  year={2026}
}

@inproceedings{liang2025multi,
  title={Multi-object sketch animation with grouping and motion trajectory priors},
  author={Liang, Guotao and Hu, Juncheng and Xing, Ximing and Zhang, Jing and Yu, Qian},
  booktitle={Proceedings of the 33rd ACM International Conference on Multimedia},
  pages={9237--9246},
  year={2025}
}

@article{liu2026automated,
  title={Automated optimization modeling via a localizable error-driven perspective},
  author={Liu, Weiting and Wu, Han and Kuang, Yufei and Han, Xiongwei and Zhong, Tao and Feng, Jianfeng and Lu, Wenlian},
  journal={arXiv preprint arXiv:2602.11164},
  year={2026}
}

@inproceedings{zhao2025redone,
  title={RedOne: Revealing Domain-specific LLM Post-Training in Social Networking Services},
  author={Zhao, Fei and Lu, Chonggang and Xie, Zheyong and Liu, Ziyan and Qian, Haofu and Huang, Jianzhao and Shi, Fangcheng and Meng, Zijie and Guo, Hongcheng and He, Mingqian and others},
  booktitle={Proceedings of the 2025 Conference on Empirical Methods in Natural Language Processing: Industry Track},
  pages={2648--2674},
  year={2025}
}

@article{shi2026androtmem,
  title={AndroTMem: From Interaction Trajectories to Anchored Memory in Long-Horizon GUI Agents},
  author={Shi, Yibo and Li, Jungang and Zhang, Linghao and Dongfang, Zihao and Wu, Biao and Tao, Sicheng and Yan, Yibo and Qin, Chenxi and Liu, Weiting and Lin, Zhixin and others},
  journal={arXiv preprint arXiv:2603.18429},
  year={2026}
}

@article{zhang2026temporal,
  title={Temporal Gains, Spatial Costs: Revisiting Video Fine-Tuning in Multimodal Large Language Models},
  author={Zhang, Linghao and Li, Jungang and Hei, Yonghua and Tao, Sicheng and Dai, Song and Yan, Yibo and Dongfang, Zihao and Liu, Weiting and Qin, Chenxi and Li, Hanqian and others},
  journal={arXiv preprint arXiv:2603.17541},
  year={2026}
}

@article{yu2026cirrusbench,
  title={CirrusBench: Evaluating LLM-based Agents Beyond Correctness in Real-World Cloud Service Environments},
  author={Yu, Yi and Hu, Guangquan and Shen, Chenghuang and Liu, Xingyan and Gu, Jing and Sun, Hangyi and Ma, Junzhuo and Liu, Weiting and Liu, Jianfeng and Pu, Mingyue and others},
  journal={arXiv preprint arXiv:2603.28569},
  year={2026}
}

@inproceedings{guo2025pet,
  title={Pet-Bench: Benchmarking the Abilities of Large Language Models as E-Pets in Social Network Services},
  author={Guo, Hongcheng and Xie, Zheyong and Cao, Shaosheng and Wang, Boyang and Liu, Weiting and Ye, Zheyu and Li, Zhoujun and Liu, Zuozhu and Lu, Wei},
  booktitle={Proceedings of the 34th ACM International Conference on Information and Knowledge Management},
  pages={6402--6407},
  year={2025}
}

@inproceedings{du2025graftllm,
  title={Knowledge Fusion of Large Language Models Via Modular Skillpacks},
  author={Guodong Du and Zhuo Li and Xuanning Zhou and Junlin Li and Zesheng Shi and Wanyu Lin and Ho-Kin Tang and Xiucheng Li and Fangming Liu and Wenya Wang and Min Zhang and Jing Li},
  booktitle={Proceedings of the International Conference on Learning Representations (ICLR)},
  year={2026}
}

@inproceedings{du2025nps,
  title={Neural Parameter Search for Slimmer Fine-Tuned Models and Better Transfer},
  author={Guodong Du and Zitao Fang and Jing Li and Junlin Li and Runhua Jiang and Shuyang Yu and Yifei Guo and Yangneng Chen and Sim Kuan Goh and Ho-Kin Tang and Daojing He and Honghai Liu and Min Zhang},
  booktitle={Proceedings of the 63rd Annual Meeting of the Association for Computational Linguistics (Volume 1: Long Papers)},
  year={2025},
  url={https://aclanthology.org/2025.acl-long.1570/},
  doi={10.18653/v1/2025.acl-long.1570}
}

@inproceedings{du2024pcb,
  title={Parameter Competition Balancing for Model Merging},
  author={Guodong Du and Junlin Lee and Jing Li and Runhua Jiang and Yifei Guo and Shuyang Yu and Hanting Liu and Sim Kuan Goh and Ho-Kin Tang and Daojing He and Min Zhang},
  booktitle={The Thirty-eighth Annual Conference on Neural Information Processing Systems (NeurIPS)},
  year={2024}
}

@article{huang2025adaptive,
  title={Adaptive Sample Scheduling for Direct Preference Optimization},
  author={Huang, Zixuan and Ban, Yikun and Fu, Lean and Li, Xiaojie and Dai, Zhongxiang and Li, Jianxin and Wang, Deqing},
  journal={arXiv preprint arXiv:2506.17252},
  year={2025}
}

@inproceedings{Huang2026RealTimeAR,
  title={Real-Time Aligned Reward Model beyond Semantics},
  author={Zixuan Huang and Xin Xia and Yuxi Ren and Jianbin Zheng and Xuefeng Xiao and Hongyan Xie and Huaqiu Li and Songshi Liang and Zhongxiang Dai and Fuzhen Zhuang and Jianxin Li and Yikun Ban and Deqing Wang},
  year={2026},
  url={https://api.semanticscholar.org/CorpusID:285240754}
}

@article{huang2026does,
  title={Does Your Reasoning Model Implicitly Know When to Stop Thinking?},
  author={Huang, Zixuan and Xia, Xin and Ren, Yuxi and Zheng, Jianbin and Wang, Xuanda and Zhang, Zhixia and Xie, Hongyan and Liang, Songshi and Chen, Zehao and Xiao, Xuefeng and others},
  journal={arXiv preprint arXiv:2602.08354},
  year={2026}
}

@article{zhang2026heterogeneous,
  title={Heterogeneous Agent Collaborative Reinforcement Learning},
  author={Zhang, Zhixia and Huang, Zixuan and Xia, Xin and Wang, Deqing and Zhuang, Fuzhen and Ma, Shuai and Ding, Ning and Yang, Yaodong and Li, Jianxin and Ban, Yikun},
  journal={arXiv preprint arXiv:2603.02604},
  year={2026}
}

@inproceedings{li2025high,
  title={High-Precision Mixed Feature Fusion Network Using Hypergraph Computation for Cervical Abnormal Cell Detection},
  author={Li, Jincheng and Dong, Danyang and Zheng, Menglin and Zhang, Jingbo and Hang, Yueqin and Zhang, Lichi and Zhao, Lili},
  booktitle={International Conference on Medical Image Computing and Computer-Assisted Intervention},
  pages={250--259},
  year={2025},
  organization={Springer}
}

@inproceedings{zhu2025ett,
  title={ETT-CKGE: Efficient Task-Driven Tokens for Continual Knowledge Graph Embedding},
  author={Zhu, Lijing and Lan, Qizhen and Tian, Qing and Sun, Wenbo and Yang, Li and Xia, Lu and Xie, Yixin and Xiao, Xi and Duan, Tiehang and Tao, Cui and others},
  booktitle={Joint European Conference on Machine Learning and Knowledge Discovery in Databases},
  pages={481--496},
  year={2025},
  organization={Springer}
}

@misc{qiu2026anchorabductivenetworkconstruction,
      title={ANCHOR: Abductive Network Construction with Hierarchical Orchestration for Reliable Probability Inference in Large Language Models}, 
      author={Wentao Qiu and Guanran Luo and Zhongquan Jian and Jingqi Gao and Meihong Wang and Qingqiang Wu},
      year={2026},
      eprint={2605.10328},
      archivePrefix={arXiv},
      primaryClass={cs.CL},
      url={https://arxiv.org/abs/2605.10328}, 
}

@inproceedings{xu2025one,
  title={One surrogate to fool them all: Universal, transferable, and targeted adversarial attacks with clip},
  author={Xu, Binyan and Dai, Xilin and Tang, Di and Zhang, Kehuan},
  booktitle={Proceedings of the 2025 ACM SIGSAC Conference on Computer and Communications Security},
  pages={3087--3101},
  year={2025}
}

@inproceedings{xu2025clip,
  title={CLIP-Guided Backdoor Defense through Entropy-Based Poisoned Dataset Separation},
  author={Xu, Binyan and Yang, Fan and Dai, Xilin and Tang, Di and Zhang, Kehuan},
  booktitle={Proceedings of the 33rd ACM International Conference on Multimedia},
  pages={7415--7423},
  year={2025}
}

@article{xu2026internal,
  title={From Internal Diagnosis to External Auditing: A VLM-Driven Paradigm for Online Test-Time Backdoor Defense},
  author={Xu, Binyan and Yang, Fan and Dai, Xilin and Tang, Di and Zhang, Kehuan},
  journal={arXiv preprint arXiv:2601.19448},
  year={2026}
}

@misc{wang2026fbsmodelingnativeparallel,
  title={FBS: Modeling Native Parallel Reading inside a Transformer}, 
  author={Tongxi Wang},
  year={2026},
  eprint={2601.21708},
  archivePrefix={arXiv},
  primaryClass={cs.AI},
  url={https://arxiv.org/abs/2601.21708}
}

@misc{wang2026trackingdriftvariationawareentropy,
  title={Tracking Drift: Variation-Aware Entropy Scheduling for Non-Stationary Reinforcement Learning}, 
  author={Tongxi Wang and Zhuoyang Xia and Xinran Chen and Shan Liu},
  year={2026},
  eprint={2601.19624},
  archivePrefix={arXiv},
  primaryClass={cs.LG},
  url={https://arxiv.org/abs/2601.19624}
}

@misc{wang2026,
  title={Stability of In-Context Learning: A Spectral Coverage Perspective}, 
  author={Tongxi Wang and Zhuoyang Xia},
  year={2026},
  eprint={2509.20677},
  archivePrefix={arXiv},
  primaryClass={cs.LG},
  url={https://arxiv.org/abs/2509.20677}
}

@inproceedings{zhu2025pathology,
  title={Pathology-Aware Prototype Evolution via LLM-Driven Semantic Disambiguation for Multicenter Diabetic Retinopathy Diagnosis},
  author={Zhu, Chunzheng and Lin, Yangfang and Shao, Jialin and Lin, Jianxin and Wang, Yijun},
  booktitle={Proceedings of the 33rd ACM International Conference on Multimedia},
  pages={9196--9205},
  year={2025}
}

@inproceedings{zhu2026medeyes,
  title={MedEyes: Learning Dynamic Visual Focus for Medical Progressive Diagnosis},
  author={Zhu, Chunzheng and Lin, Yangfang and Chen, Shen and Wang, Yijun and Lin, Jianxin},
  booktitle={Proceedings of the AAAI Conference on Artificial Intelligence},
  volume={40},
  number={16},
  pages={13916--13924},
  year={2026}
}

@article{lin2026medcausalx,
  title={MedCausalX: Adaptive Causal Reasoning with Self-Reflection for Trustworthy Medical Vision-Language Models},
  author={Lin, Jianxin and Zhu, Chunzheng and Kneuertz, Peter J and Bai, Yunfei and Xue, Yuan},
  journal={arXiv preprint arXiv:2603.23085},
  year={2026}
}

@article{ge2025progis,
  title={Progis: Prototype-guided interactive segmentation for pathological images},
  author={Ge, Jiusong and Zhang, Di and Zhan, Yingkang and Liu, Jiashuai and Gong, Tieliang and Wu, Jialun and Crispin, Mireia and Li, Chen and Gao, Zeyu},
  journal={IEEE Transactions on Medical Imaging},
  year={2025},
  publisher={IEEE}
}

@article{wang2025fully,
  title={A fully annotated pathology slide dataset for early gastric cancer and precancerous lesions},
  author={Wang, Chunbao and Ge, Jiusong and Niu, Yi and Ding, Caixia and Fan, Yangyang and Chang, Hongyun and Yang, Zhe and Ran, Caihong and Teng, Xiali and Wang, Xiaolin and others},
  journal={Scientific Data},
  volume={12},
  number={1},
  pages={1326},
  year={2025},
  publisher={Nature Publishing Group UK London}
}

@inproceedings{li2025similarity,
  title={Similarity-Aware Dual-Perspective Learning for Medical Event Prediction},
  author={Li, Yang and Ge, Jiusong and Cao, Shilei and Zhan, Yingkang and Gong, Zhangpeng and Niu, Jiawei and Liu, Jiashuai and Zhang, Di and Li, Chen},
  booktitle={2025 IEEE International Conference on Bioinformatics and Biomedicine (BIBM)},
  pages={6263--6270},
  year={2025},
  organization={IEEE}
}

@article{liu2026improving,
  title={Improving the Completeness and Comparability of Segment Disclosures: A Large Language Model Approach},
  author={Liu, Yue and Cheng, Zhiyuan and Lai, Longying},
  journal={Available at SSRN 6720239},
  year={2026}
}

@misc{chen2025autoneuralcodesigningvisionlanguagemodels,
  title={AutoNeural: Co-Designing Vision-Language Models for NPU Inference}, 
  author={Wei Chen and Liangmin Wu and Yunhai Hu and Zhiyuan Li and Zhiyuan Cheng and Yicheng Qian and Lingyue Zhu and Zhipeng Hu and Luoyi Liang and Qiang Tang and Zhen Liu and Han Yang},
  year={2025},
  eprint={2512.02924},
  archivePrefix={arXiv},
  primaryClass={cs.CL},
  url={https://arxiv.org/abs/2512.02924}
}

@inproceedings{Cheng2026,
  title={Regime-Dependent Volatility Dynamics: Evidence from Time-Series Analysis},
  author={Kai Cheng and Xiaoxi Qi and Zhiyuan Cheng and Longying Lai and Xuan Liu},
  year={2026},
  booktitle={Proceedings of the 2026 3rd International Conference on Applied Economics, Management Science and Social Development (AEMSS 2026)},
  pages={179-189},
  publisher={Atlantis Press}
}

@article{hu2026bridging,
  title={Bridging stepwise lab-informed pretraining and knowledge-guided learning for diagnostic reasoning},
  author={Hu, Pengfei and Lu, Chang and Wang, Fei and Ning, Yue},
  journal={IEEE Journal of Biomedical and Health Informatics},
  year={2026},
  publisher={IEEE}
}

@inproceedings{han-etal-2025-black,
  title={No Black Boxes: Interpretable and Interactable Predictive Healthcare with Knowledge-Enhanced Agentic Causal Discovery},
  author={Han, Xiaoxue and Hu, Pengfei and Lu, Chang and Ding, Jun-En and Liu, Feng and Ning, Yue},
  booktitle={Findings of the Association for Computational Linguistics: EMNLP 2025},
  year={2025},
  pages={23415--23427}
}

@article{hu2025udoncare,
  title={Udoncare: Hierarchy pruning for unseen domain discovery in predictive healthcare},
  author={Hu, Pengfei and Han, Xiaoxue and Wang, Fei and Ning, Yue},
  journal={arXiv preprint arXiv:2506.06977},
  year={2025}
}

@article{hu2026exploring,
  title={Exploring Accurate and Transparent Domain Adaptation in Predictive Healthcare via Concept-Grounded Orthogonal Inference},
  author={Hu, Pengfei and Lu, Chang and Liu, Feifan and Ning, Yue},
  journal={arXiv preprint arXiv:2602.12542},
  year={2026}
}

@article{liu2024synthvlm,
  title={Synthvlm: High-efficiency and high-quality synthetic data for vision language models},
  author={Liu, Zheng and Liang, Hao and Huang, Xijie and Xiong, Wentao and Yu, Qinhan and Sun, Linzhuang and Chen, Chong and He, Conghui and Cui, Bin and Zhang, Wentao},
  journal={arXiv preprint arXiv:2407.20756},
  volume={3},
  year={2024}
}

@article{liu2025fusion,
  title={FUSION: Fully integration of vision-language representations for deep cross-modal understanding},
  author={Liu, Zheng and Liu, Mengjie and Chen, Jingzhou and Xu, Jingwei and Cui, Bin and He, Conghui and Zhang, Wentao},
  journal={arXiv preprint arXiv:2504.09925},
  year={2025}
}

@article{lin2026mmfinereason,
  title={MMFineReason: Closing the Multimodal Reasoning Gap via Open Data-Centric Methods},
  author={Lin, Honglin and Liu, Zheng and Zhu, Yun and Qin, Chonghan and Lin, Juekai and Shang, Xiaoran and He, Conghui and Zhang, Wentao and Wu, Lijun},
  journal={arXiv preprint arXiv:2601.21821},
  year={2026}
}

@article{liu2026chartverse,
  title={ChartVerse: Scaling Chart Reasoning via Reliable Programmatic Synthesis from Scratch},
  author={Liu, Zheng and Lin, Honglin and Qin, Chonghan and Wang, Xiaoyang and Gao, Xin and Li, Yu and Cai, Mengzhang and Zhu, Yun and Zhong, Zhanping and Pei, Qizhi and others},
  journal={arXiv preprint arXiv:2601.13606},
  year={2026}
}

@misc{ke2026deformbavisionstatespace,
  title={Deformba: Vision State Space Model with Adaptive State Fusion}, 
  author={Hongyu Ke and Jack Morris and Yongkang Liu and Satoshi Kitai and Kentaro Oguchi and Yi Ding and Haoxin Wang},
  year={2026},
  eprint={2605.21308},
  archivePrefix={arXiv},
  primaryClass={cs.CV},
  url={https://arxiv.org/abs/2605.21308}
}

@article{jiang2025hulu,
  title={Hulu-med: A transparent generalist model towards holistic medical vision-language understanding},
  author={Jiang, Songtao and Wang, Yuan and Song, Sibo and Hu, Tianxiang and Zhou, Chenyi and Pu, Bin and Zhang, Yan and Yang, Zhibo and Feng, Yang and Zhou, Joey Tianyi and others},
  journal={arXiv preprint arXiv:2510.08668},
  year={2025}
}

@article{zhou2024reducing,
  title={Reducing manual labeling requirements and improved retinal ganglion cell identification in 3D AO-OCT volumes using semi-supervised learning},
  author={Zhou, Mengxi and Zhang, Yue and Karimi Monsefi, Amin and Choi, Stacey S and Doble, Nathan and Parthasarathy, Srinivasan and Ramnath, Rajiv},
  journal={Biomedical Optics Express},
  volume={15},
  number={8},
  pages={4540--4556},
  year={2024},
  publisher={Optica Publishing Group}
}

@article{zhang2026adaptive,
  title={Adaptive Temporal Mixture of Experts for Predicting Stiffness Metrics from the Ocular Response Analyzer and Identifying Keratoconus},
  author={Zhang, Yue and Padhee, Swati and Yuhas, Phillip T and Roberts, Cynthia J and Parthasarathy, Srinivasan},
  journal={American Journal of Ophthalmology},
  year={2026},
  publisher={Elsevier}
}

@article{zhou2025isosnet,
  title={ISOSNet: a unified framework for cone photoreceptor detection and inner segment and outer segment length measurement from AO-OCT B-scans},
  author={Zhou, Mengxi and Zhang, Yue and Kirkendall, Eli and Karimi Monsefi, Amin and Wolfe, Matthew and Chitkara, Kiran A and Choi, Stacey S and Doble, Nathan and Parthasarathy, Srinivasan and Ramnath, Rajiv},
  journal={Biomedical Optics Express},
  volume={16},
  number={8},
  pages={3237--3254},
  year={2025},
  publisher={Optica Publishing Group}
}

@article{salarian2025medequalizer,
  title={MedEqualizer: A Framework Investigating Bias in Synthetic Medical Data and Mitigation via Augmentation},
  author={Salarian, Sama and Zhang, Yue and Padhee, Swati and Parthasarathy, Srinivasan},
  journal={arXiv preprint arXiv:2511.01054},
  year={2025}
}

@inproceedings{zhang2026heteromile,
  title={Heteromile: a multi-level graph representation learning framework for heterogeneous graphs},
  author={Zhang, Yue and He, Yuntian and Gurukar, Saket and Parthasarathy, Srinivasan},
  booktitle={Proceedings of the Nineteenth ACM International Conference on Web Search and Data Mining},
  pages={63--72},
  year={2026}
}

@article{he2022webmile,
  title={WebMILE: democratizing network representation learning at scale},
  author={He, Yuntian and Zhang, Yue and Gurukar, Saket and Parthasarathy, Srinivasan},
  journal={Proceedings of the VLDB Endowment},
  volume={15},
  number={12},
  year={2022}
}

@article{xia2025association,
  title={Association Between Conversational Multitasking and Clinician Work Behaviors at a Large US Health Care System: Cohort Study},
  author={Xia, Linlin and Lew, Daphne and Baratta, Laura and Eiden, Elise and Lou, Sunny and Kannampallil, Thomas},
  journal={Journal of medical Internet research},
  volume={27},
  pages={e72768},
  year={2025},
  publisher={JMIR Publications Toronto, Canada}
}

@article{lew2025association,
  title={Association of EHR-Integrated Secure Messaging Use with Clinician Workload and Attention Switching},
  author={Lew, Daphne and Baratta, Laura R and Xia, Linlin and Eiden, Elise and Sinsky, Christine A and Kannampallil, Thomas and Lou, Sunny S},
  journal={Journal of general internal medicine},
  volume={40},
  number={10},
  pages={2240--2247},
  year={2025},
  publisher={Springer}
}

@article{lou2024secure,
  title={Secure messaging use and wrong-patient ordering errors among inpatient clinicians},
  author={Lou, Sunny S and Lew, Daphne and Xia, Linlin and Baratta, Laura and Eiden, Elise and Kannampallil, Thomas},
  journal={JAMA Network Open},
  volume={7},
  number={12},
  pages={e2447797},
  year={2024}
}

@inproceedings{tao2025autopcr,
  title={AutoPCR: Automated Phenotype Concept Recognition by Prompting},
  author={Tao, Yicheng and Huang, Yuanhao and Wang, Yiqun and Luo, Xin and Liu, Jie},
  booktitle={arXiv preprint arXiv:2507.19315},
  year={2025}
}

@inproceedings{guo2025quantized,
  title={Quantized-TinyLLaVA: A New Multimodal Foundation Model Enables Efficient Split Learning},
  author={Guo, Jiajun and Luo, Xin and Zheng, Jiayin and Wang, Yiqun and Chang, Kai-Wei and Wang, Wei and Liu, Jie},
  booktitle={arXiv preprint arXiv:2511.23402},
  year={2025}
}

@article{chen2022explain,
  title={Explain the explainer: Interpreting model-agnostic counterfactual explanations of a deep reinforcement learning agent},
  author={Chen, Ziheng and Silvestri, Fabrizio and Tolomei, Gabriele and Wang, Jia and Zhu, He and Ahn, Hongshik},
  journal={IEEE Transactions on Artificial Intelligence},
  volume={5},
  number={4},
  pages={1443--1457},
  year={2022},
  publisher={IEEE}
}

@article{fan2026crab,
  title={CRAB: Codebook Rebalancing for Bias Mitigation in Generative Recommendation},
  author={Fan, Zezhong and Chen, Ziheng and Ma, Luyi and Huang, Jin and Morishetti, Lalitesh and Nag, Kaushiki and Kumar, Sushant and Achan, Kannan},
  journal={arXiv preprint arXiv:2604.05113},
  year={2026}
}

@inproceedings{ke2025mambev,
  title={MamBEV: Enabling State Space Models to Learn Birds-Eye-View Representations},
  author={Ke, Hongyu and Morris, Jack and Oguchi, Kentaro and Cao, Xiaofei and Liu, Yongkang and Wang, Haoxin and Ding, Yi},
  booktitle={The Thirteenth International Conference on Learning Representations},
  year={2025}
}

@inproceedings{jiang2026mm,
  title={MM-Snowball: Evaluating and Mitigating Hallucination Snowballing in Multimodal Multi-turn Dialogue},
  author={Jiang, Yue and Jiang, Xue and Zhang, Lihua and Wang, Zhiqiang and Lu, Yuhang and Wang, Peng and Han, Bo and Zheng, Feng and Yang, Dingkang},
  booktitle={ICML},
  year={2026}
}

@article{jiang2026danmakutppbench,
  title={Danmakutppbench: A multi-modal benchmark for temporal point process modeling and understanding},
  author={Jiang, Yue and Li, Jichu and Liu, Yang and Yang, Dingkang and Zhou, Feng and Kong, Quyu},
  journal={Advances in Neural Information Processing Systems},
  volume={38},
  year={2026}
}

@inproceedings{jiang2025satiredecoder,
  title={SatireDecoder: Visual Cascaded Decoupling for Enhancing Satirical Image Comprehension},
  author={Jiang, Yue and Xue, Haiwei and Han, Minghao and Li, Mingcheng and Hou, Xiaolu and Yang, Dingkang and Zhang, Lihua and Zheng, Xu},
  booktitle={AAAI},
  year={2026}
}

@inproceedings{jiang2025comt,
  title={Comt: Chain-of-medical-thought reduces hallucination in medical report generation},
  author={Jiang, Yue and Chen, Jiawei and Yang, Dingkang and Li, Mingcheng and Wang, Shunli and Wu, Tong and Li, Ke and Zhang, Lihua},
  booktitle={ICASSP 2025-2025 IEEE International Conference on Acoustics, Speech and Signal Processing (ICASSP)},
  pages={1--5},
  year={2025},
  organization={IEEE}
}

@inproceedings{jiang2026multi,
  title={Multi-Agent Diagnostic Collaboration and Segmentation-Aware Residual Decoding for Hallucination-Resistant Medical VQA},
  author={Jiang, Yue and Han, Minghao and Li, Mingcheng and Hou, Xiaolu and Zhang, Hao and Zhu, Wenkai and Li, Hao and He, Yangfan and Wu, Guangxin and Yang, Dingkang and others},
  booktitle={ICASSP 2026-2026 IEEE International Conference on Acoustics, Speech and Signal Processing (ICASSP)},
  pages={11122--11126},
  year={2026},
  organization={IEEE}
}

@inproceedings{chen2024can,
  title={Can llms' tuning methods work in medical multimodal domain?},
  author={Chen, Jiawei and Jiang, Yue and Yang, Dingkang and Li, Mingcheng and Wei, Jinjie and Qian, Ziyun and Zhang, Lihua},
  booktitle={International Conference on Medical Image Computing and Computer-Assisted Intervention},
  pages={112--122},
  year={2024},
  organization={Springer}
}

@inproceedings{wei2025identifying,
  title={Identifying cellular niches in spatial transcriptomics: An investigation into the capabilities of large language models},
  author={Wei, Huanhuan and Luo, Xiao and Yu, Hongyi and Liang, Jinping and Yang, Luning and Lin, Lixing and Popa, Alexandra and Yan, Xiting},
  booktitle={Proceedings of the 63rd Annual Meeting of the Association for Computational Linguistics (Volume 1: Long Papers)},
  pages={9275--9289},
  year={2025}
}

@misc{lin2026reflectguardenhancingllmsafeguards,
  title={Reflect-Guard: Enhancing LLM Safeguards against Adversarial Prompts via Logical Self-Reflection}, 
  author={Lixing Lin and Juli You and Yue Li and Luyun Lin and Yiqing Wang and Zhen Zhang and Moxuan Zheng},
  year={2026},
  eprint={2605.24834},
  archivePrefix={arXiv},
  primaryClass={cs.CR},
  url={https://arxiv.org/abs/2605.24834}
}

@article{zhang2026mitigating,
  title={Mitigating Multimodal Hallucination via Phase-wise Self-reward},
  author={Zhang, Yu and Sun, Chuyang and Chen, Kehai and Bai, Xuefeng and Xiang, Yang and Zhang, Min},
  journal={arXiv preprint arXiv:2604.17982},
  year={2026}
}

@article{zhang2026instruction,
  title={Instruction Anchors: Dissecting the Causal Dynamics of Modality Arbitration},
  author={Zhang, Yu and Xu, Mufan and Bai, Xuefeng and Zhang, Pengfei and Xiang, Yang and Zhang, Min and others},
  journal={arXiv preprint arXiv:2602.03677},
  year={2026}
}

@article{zhang2025evaluating,
  title={Evaluating and steering modality preferences in multimodal large language model},
  author={Zhang, Yu and Ma, Jinlong and Hou, Yongshuai and Bai, Xuefeng and Chen, Kehai and Xiang, Yang and Yu, Jun and Zhang, Min},
  journal={arXiv preprint arXiv:2505.20977},
  year={2025}
}

@inproceedings{zhang2024question,
  title={Question-guided knowledge graph re-scoring and injection for knowledge graph question answering},
  author={Zhang, Yu and Chen, Kehai and Bai, Xuefeng and Kang, Zhao and Guo, Quanjiang and Zhang, Min},
  booktitle={Findings of the Association for Computational Linguistics: EMNLP 2024},
  pages={8972--8985},
  year={2024}
}

@article{du2026dynamic,
  title={Dynamic Model Merging Made Slim},
  author={Du, Guodong and Lin, Wanyu},
  journal={arXiv preprint arXiv:2605.18904},
  year={2026}
}

@misc{li2026whatsmissingscreentoactionuiintheloop,
  title={What's Missing in Screen-to-Action? Towards a UI-in-the-Loop Paradigm for Multimodal GUI Reasoning}, 
  author={Songze Li and Xiaoke Guo and Tianqi Liu and Biao Yi and Zhaoyan Gong and Zhiqiang Liu and Huajun Chen and Wen Zhang},
  year={2026},
  eprint={2604.06995},
  archivePrefix={arXiv},
  primaryClass={cs.AI},
  url={https://arxiv.org/abs/2604.06995}
}

@inproceedings{Li_2025,
  title={Enrich-on-Graph: Query-Graph Alignment for Complex Reasoning with LLM Enriching},
  url={http://dx.doi.org/10.18653/v1/2025.emnlp-main.390},
  DOI={10.18653/v1/2025.emnlp-main.390},
  booktitle={Proceedings of the 2025 Conference on Empirical Methods in Natural Language Processing},
  publisher={Association for Computational Linguistics},
  author={Li, Songze and Liu, Zhiqiang and Gui, Zhengke and Chen, Huajun and Zhang, Wen},
  year={2025},
  pages={7683--7703}
}

@misc{li2025layerlogitslogicempowering,
  title={Last Layer Logits to Logic: Empowering LLMs with Logic-Consistent Structured Knowledge Reasoning}, 
  author={Songze Li and Zhiqiang Liu and Yun Zhu and Huajun Chen and Wen Zhang},
  year={2025},
  eprint={2511.07910},
  archivePrefix={arXiv},
  primaryClass={cs.CL},
  url={https://arxiv.org/abs/2511.07910}
}

@misc{kong2026aiautoresearch,
  title={AI for Auto-Research: Roadmap \& User Guide},
  author={Lingdong Kong and Xian Sun and Wei Chow and Linfeng Li and Kevin Qinghong Lin and Xuan Billy Zhang and Song Wang and Rong Li and Qing Wu and Wei Gao and Yingshuo Wang and Shaoyuan Xie and Jiachen Liu and Leigang Qu and Shijie Li and Lai Xing Ng and Benoit R. Cottereau and Ziwei Liu and Tat-Seng Chua and Wei Tsang Ooi},
  year={2026},
  eprint={2605.18661},
  archivePrefix={arXiv},
  primaryClass={cs.AI},
  url={https://arxiv.org/abs/2605.18661}
}

@article{huang2025ccsumsp,
  title={CCSUMSP: A cross-subject Chinese speech decoding framework with unified topology and multi-modal semantic pre-training},
  author={Huang, Shuai and Wang, Yongxiong and Luo, Huan},
  journal={Information Fusion},
  volume={119},
  pages={103022},
  year={2025},
  publisher={Elsevier}
}

@inproceedings{huang2025ssaad,
  title={SSAAD: A multi-scale temporal-frequency graph network for binary auditory attention detection with self-supervised learning},
  author={Huang, Shuai and Wang, Yongxiong and Luo, Huan and Jia, Shuwen and Chen, Han and Qin, Chendong and He, Zhongcai and Luo, Rui},
  booktitle={ICASSP 2025-2025 IEEE International Conference on Acoustics, Speech and Signal Processing (ICASSP)},
  pages={1--5},
  year={2025},
  organization={IEEE}
}

@article{huang2025dual,
  title={A dual-branch generative adversarial network with self-supervised enhancement for robust auditory attention decoding},
  author={Huang, Shuai and Wang, Yongxiong and Luo, Huan},
  journal={Engineering Applications of Artificial Intelligence},
  pages={111122},
  year={2025},
  publisher={Elsevier}
}

@inproceedings{huang2025mindev,
  title={Mindev: Multi-modal integrated diffusion framework for video reconstruction from eeg signals},
  author={Huang, Shuai and Wang, Yongxiong and Luo, Huan and Jing, Haodong and Qin, Chendong and Tang, Jingqun},
  booktitle={Proceedings of the 33rd ACM International Conference on Multimedia},
  pages={3350--3359},
  year={2025}
}

@article{huang2026need,
  title={Need: Cross-subject and cross-task generalization for video and image reconstruction from eeg signals},
  author={Huang, Shuai and Luo, Huan and Jing, Haodong and Zhang, Qixian and Chang, Litao and Feng, Yating and Lin, Xiao and Qin, Chendong and Chen, Han and Jia, Shuwen and others},
  journal={Advances in Neural Information Processing Systems},
  volume={38},
  pages={173134--173173},
  year={2026}
}

@article{song2023deterministic,
  title     = {From deterministic to stochastic: an interpretable stochastic model-free reinforcement learning framework for portfolio optimization},
  author    = {Song, Zitao and Wang, Yining and Qian, Pin and Song, Sifan and Coenen, Frans and Jiang, Zhengyong and Su, Jionglong},
  journal   = {Applied Intelligence},
  volume    = {53},
  pages     = {15188--15203},
  year      = {2023},
  doi       = {10.1007/s10489-022-04217-5},
  url       = {https://doi.org/10.1007/s10489-022-04217-5}
}

@inproceedings{wang2025gemsllm,
  title     = {{GEMs-LLM}: Integrating Large Language Models with Goal-Aware Exploration for {RL}-Based Portfolio Optimization},
  author    = {Wang, Yining and Lu, Zhixiang and Qian, Pin and Su, Jionglong and Zhou, Mian and Li, Chong and Jiang, Zhengyong},
  booktitle = {Advanced Intelligent Computing Technology and Applications},
  series    = {Communications in Computer and Information Science},
  volume    = {2566},
  pages     = {516--527},
  year      = {2025},
  publisher = {Springer},
  address   = {Singapore},
  doi       = {10.1007/978-981-96-9949-0_43},
  url       = {https://doi.org/10.1007/978-981-96-9949-0_43}
}

@misc{chen2026doesragknowretrieval,
      title={Does RAG Know When Retrieval Is Wrong? Diagnosing Context Compliance under Knowledge Conflict}, 
      author={Yihang Chen and Pin Qian and Su Wang and Sipeng Zhang and Huan Xu and Shuhuai Lin and Xinpeng Wei},
      year={2026},
      eprint={2605.14473},
      archivePrefix={arXiv},
      primaryClass={cs.CL},
      url={https://arxiv.org/abs/2605.14473}, 
}

@misc{qian2026relevantwarrantedevidenceforcecalibration,
      title={Relevant Is Not Warranted: Evidence-Force Calibration for Cited RAG}, 
      author={Pin Qian and Su Wang and Xiaoyuan Wang and Yihang Chen and Wenxuan Xu and Qiaolin Yu and Shuhuai Lin and Sipeng Zhang and Junxian You and Xinpeng Wei},
      year={2026},
      eprint={2605.28044},
      archivePrefix={arXiv},
      primaryClass={cs.AI},
      url={https://arxiv.org/abs/2605.28044}, 
}

@misc{wang2026safeskillscollidemeasuring,
      title={When Safe Skills Collide: Measuring Compositional Risk in Agent Skill Ecosystems}, 
      author={Su Wang and Pin Qian and Yihang Chen and Junxian You and Xiaoyuan Wang and Xiaochong Jiang and Lifei Liu and Haoran Yu and Jingzhou Xu},
      year={2026},
      eprint={2606.00448},
      archivePrefix={arXiv},
      primaryClass={cs.SE},
      url={https://arxiv.org/abs/2606.00448}, 
}

@misc{zhang2026smalllanguagemodelagents,
      title={Small Language Model Agents Enable Efficient and High-Quality Knowledge Mining}, 
      author={Sipeng Zhang and Shuhuai Lin and Xinpeng Wei and Yihang Chen and Pin Qian and Su Wang and Huan Xu},
      year={2026},
      eprint={2510.01427},
      archivePrefix={arXiv},
      primaryClass={cs.AI},
      url={https://arxiv.org/abs/2510.01427}, 
}

@misc{zhang2025dualtapdualtaskadversarialprotector,
      title={DualTAP: A Dual-Task Adversarial Protector for Mobile MLLM Agents}, 
      author={Fuyao Zhang and Jiaming Zhang and Che Wang and Xiongtao Sun and Yurong Hao and Guowei Guan and Wenjie Li and Longtao Huang and Wei Yang Bryan Lim},
      year={2025},
      eprint={2511.13248},
      archivePrefix={arXiv},
      primaryClass={cs.CR},
      url={https://arxiv.org/abs/2511.13248}, 
}

@InProceedings{Zhang_2026_CVPR,
    author    = {Zhang, Zhenguo and Zheng, Haohan and Wang, Yishen and Xu, Le and Deng, Tianchen and Chen, Xuefeng and Chen, Qu and Zhang, Bo and Huang, Wuxiong},
    title     = {OmniDrive-R1: Reinforcement-driven Interleaved Multi-modal Chain-of-Thought for Trustworthy Vision-Language Autonomous Driving},
    booktitle = {Proceedings of the IEEE/CVF Conference on Computer Vision and Pattern Recognition (CVPR) Findings},
    month     = {June},
    year      = {2026},
    pages     = {1106-1116}
}

@article{deng2025gaussiandwm3dgaussiandriving,
      title={GaussianDWM: 3D Gaussian Driving World Model for Unified Scene Understanding and Multi-Modal Generation}, 
      author={Tianchen Deng and Xuefeng Chen and Yi Chen and Qu Chen and Yuyao Xu and Lijin Yang and Le Xu and Yu Zhang and Bo Zhang and Wuxiong Huang and Hesheng Wang},
      year={2025},
      journal={arXiv preprint arXiv:2512.23180}, 
}

@inproceedings{zhu2026ants,
  title={ANTS: Adaptive Negative Textual Space Shaping for OOD Detection via Test-Time MLLM Understanding and Reasoning},
  author={Zhu, Wenjie and Zhang, Yabin and Jin, Xin and Zeng, Wenjun and Zhang, Lei},
  booktitle={Proceedings of the IEEE/CVF Conference on Computer Vision and Pattern Recognition},
  pages={20--30},
  year={2026}
}

@inproceedings{zhu2025knowledge,
  title={Knowledge Regularized Negative Feature Tuning of Vision-Language Models for Out-of-Distribution Detection},
  author={Zhu, Wenjie and Zhang, Yabin and Jin, Xin and Zeng, Wenjun and Zhang, Lei},
  booktitle={Proceedings of the 33rd ACM International Conference on Multimedia},
  pages={3565--3574},
  year={2025}
}

@article{li2022darc,
  title={DARC: Deep adaptive regularized clustering for histopathological image classification},
  author={Li, Junjian and Liu, Jin and Yue, Hailin and Cheng, Jianhong and Kuang, Hulin and Bai, Harrison and Wang, Yuping and Wang, Jianxin},
  journal={Medical image analysis},
  volume={80},
  pages={102521},
  year={2022},
  publisher={Elsevier}
}

@inproceedings{li2025mico,
  title={MiCo: Multiple Instance Learning with Context-Aware Clustering for Whole Slide Image Analysis},
  author={Li, Junjian and Liu, Jin and Kuang, Hulin and Yue, Hailin and He, Mengshen and Wang, Jianxin},
  booktitle={International Conference on Medical Image Computing and Computer-Assisted Intervention},
  pages={376--385},
  year={2025},
  organization={Springer}
}

@article{li20252,
  title={CA2CL: Cluster-Aware Adversarial Contrastive Learning for Pathological Image Analysis},
  author={Li, Junjian and Kuang, Hulin and Liu, Jin and Yue, Hailin and Wang, Jianxin},
  journal={IEEE Journal of Biomedical and Health Informatics},
  year={2025},
  publisher={IEEE}
}

@inproceedings{li2026universal,
  title={Universal-to-Specific: Dynamic Knowledge-Guided Multiple Instance Learning for Few-Shot Whole Slide Image Classification},
  author={Li, Junjian and Kuang, Hulin and Liu, Jin and Yue, Hailin and He, Mengshen and Wang, Jianxin},
  booktitle={Proceedings of the IEEE/CVF Conference on Computer Vision and Pattern Recognition},
  pages={26614--26623},
  year={2026}
}

@article{li2026domain,
  title={Domain-specific self-supervised contrastive learning with contrast-aware pair refinement for pathological image analysis},
  author={Li, Junjian and Kuang, Hulin and Liu, Jin and Yue, Hailin and Wang, Jianxin},
  journal={Big Data Mining and Analytics},
  volume={9},
  number={3},
  pages={705--718},
  year={2026},
  publisher={TUP}
}

@inproceedings{li2026catp,
  title={Catp: Contextually adaptive token pruning for efficient and enhanced multimodal in-context learning},
  author={Li, Yanshu and Yang, Jianjiang and Shen, Zhennan and Han, Ligong and Xu, Haoyan and Tang, Ruixiang},
  booktitle={Proceedings of the AAAI Conference on Artificial Intelligence},
  volume={40},
  number={8},
  pages={6619--6627},
  year={2026}
}

@article{li2026personalize,
  title={Personalize Your Large Vision-language Models With In-context Prompt Tuning},
  author={Li, Yanshu and Li, Jiaqian and Yu, Kuai and Xiao, Xi and Liu, Dongfang and Wang, Tianyang and Tang, Ruixiang},
  journal={arXiv preprint arXiv:2605.31513},
  year={2026}
}

@inproceedings{zhou2026comem,
  title={Comem: Compositional concept-graph memory for vision--language adaptation},
  author={Zhou, Heng and Tang, Jing and Zhang, Jusheng and Li, Yanshu and Xiao, Canran and Hou, Liwei and Ke, Zong and Yao, Jiawei},
  booktitle={The Fourteenth International Conference on Learning Representations},
  year={2026}
}

@article{xiao6281833focus,
  title={Focus Where It Matters: LLM-Guided Attention Priors for Few-Shot Segmentation},
  author={Xiao, Canran and Ke, Zong and Zhao, Puning},
  journal={Available at SSRN 6281833}
}

@inproceedings{xiao2026reversible,
  title={Reversible primitive--composition alignment for continual vision--language learning},
  author={Xiao, Canran and Xu, Tianxiang and Ma, Siyuan and Jiang, Yiyang and Gao, Haoyu and Wu, Yuhan},
  booktitle={The Fourteenth International Conference on Learning Representations},
  year={2026}
}

@article{wu2026better,
  title={Better Eyes, Better Thoughts: Why Vision Chain-of-Thought Fails in Medicine},
  author={Wu, Yuan and Yang, Zongxian and Qian, Jiayu and Gao, Songpan and Chen, Guanxing and Li, Qiankun and Huang, Yu-An and Huang, Zhi-An},
  journal={arXiv preprint arXiv:2603.06665},
  year={2026}
}

@article{yang2025med,
  title={Med-REFL: Medical Reasoning Enhancement via Self-Corrected Fine-grained Reflection},
  author={Yang, Zongxian and Qian, Jiayu and Peng, Zegao and Zhang, Haoyu and Huang, Yu-An and Tan, KC and Huang, Zhi-An},
  journal={arXiv preprint arXiv:2506.13793},
  year={2025}
}

@article{yang2025qm,
  title={Qm-tot: A medical tree of thoughts reasoning framework for quantized model},
  author={Yang, Zongxian and Qian, Jiayu and Tan, Kay Chen and Wong, Hau-San and Chen, Yulong and Zhang, Haoyu and Huang, Zhi-An},
  journal={arXiv preprint arXiv:2504.12334},
  year={2025}
}

@article{hu2026seal,
  title={SEAL: Synergistic Co-Evolution of Agents and Learning Environments},
  author={Hu, Yihao and Wen, Zhihao and Liu, Xiujin and Wang, Pan and Zhang, Xin and Wu, Wei},
  journal={arXiv preprint arXiv:2605.24426},
  year={2026}
}

@article{wang2026atlasva,
  title={AtlasVA: Self-Evolving Visual Skill Memory for Teacher-Free VLM Agents},
  author={Wang, Pan and Hu, Yihao and Liu, Xiujin and Yang, Jingchu and Wang, Hang and Wen, Zhihao},
  journal={arXiv preprint arXiv:2605.17933},
  year={2026}
}

@InProceedings{zhang2026seer,
      author    = { Zhang, Tongrui and Wang, Chenhui and Li, Yongming and Chen, Zhihao and Zhan, Xufeng and Shan, Hongming},
      title     = { Skill-Evolving Grounded Reasoning for Free-Text Promptable 3D Medical Image Segmentation },
      booktitle = { Medical Image Computingand Computer Assisted Intervention },
      year      = { 2026 }
}

@misc{zhao2026resilphaseplugandplayphasemapping,
      title={ResilPhase: Plug-and-Play Phase Mapping and Noise-Resilient Macro-Trajectory Extrapolation for Diffusion Acceleration}, 
      author={Qicheng Zhao and Yu Li and Qi Sun and Zheyu Yan},
      year={2026},
      eprint={2606.26769},
      archivePrefix={arXiv},
      primaryClass={cs.AI},
      url={https://arxiv.org/abs/2606.26769}, 
}

@inproceedings{meng2022early,
  title={Early forecasting of the impact of traffic accidents using a single shot observation},
  author={Meng, Guangyu and Jiang, Qisheng and Fu, Kaiqun and Lin, Beiyu and Lu, Chang-Tien and Chen, Zhqian},
  booktitle={Proceedings of the 2022 SIAM International Conference on Data Mining (SDM)},
  pages={100--108},
  year={2022},
  organization={SIAM}
}

@article{meng2025efficient,
  title={Efficient Approximation of Earth Mover's Distance Based on Nearest Neighbor Search},
  author={Meng, Guangyu and Zhou, Ruyu and Liu, Liu and Liang, Peixian and Liu, Fang and Chen, Danny Z and Niemier, Michael and Hu, Xiaobo Sharon},
  journal={IEEE Transactions on Multimedia},
  year={2025},
  publisher={IEEE}
}

@article{meng2025psychology,
  title={A Psychology-based Unified Dynamic Framework for Curriculum Learning},
  author={Meng, Guangyu and Zeng, Qinkai and Lalor, John P and Yu, Hong},
  journal={Computational Linguistics},
  pages={1--49},
  year={2025},
  publisher={MIT Press 255 Main Street, 9th Floor, Cambridge, Massachusetts 02142, USA~…}
}

@article{chambers2025stable,
  title={A Stable and Theoretically Grounded Gromov-Wasserstein Distance for Reeb Graph Comparison using Persistence Images},
  author={Chambers, Erin W and Meng, Guangyu},
  journal={arXiv preprint arXiv:2507.01171},
  year={2025}
}

@inproceedings{meng2026topocl,
  title={TopoCL: Topological Contrastive Learning for Medical Imaging},
  author={Meng, Guangyu and Gu, Pengfei and Liang, Peixian and Lalor, John P and Chambers, Erin Wolf and Chen, Danny Z},
  booktitle={Proceedings of the IEEE/CVF Conference on Computer Vision and Pattern Recognition},
  pages={42681--42690},
  year={2026}
}

@article{meng2026ms,
  title={MS-COOT: Comparing Morse-Smale Complexes with Co-Optimal Transport},
  author={Meng, Guangyu and Li, Mingzhe and Chambers, Erin Wolf},
  journal={arXiv preprint arXiv:2606.08258},
  year={2026}
}

@article{xia2025uniem,
  title={UniEM-3M: A Universal Electron Micrograph Dataset for Microstructural Segmentation and Generation},
  author={Xia, Zhiyi and Li, Yiming and Tang, Shi and Fan, Zuxin and Fang, Xi and Tao, Haoyi and Cai, Xiaochen and Ke, Guolin and Zhang, Linfeng and Hong, Yanhui and others},
  journal={arXiv preprint arXiv:2508.16239},
  year={2025}
}

@article{du2026weatherreasonseg,
  title={WeatherReasonSeg: A Benchmark for Weather-Aware Reasoning Segmentation in Visual Language Models},
  author={Du, Wanjun and Yuan, Zifeng and Chen, Tingting and Ke, Fucai and Lin, Beibei and Zhang, Shunli},
  journal={arXiv preprint arXiv:2603.17680},
  year={2026}
}

@misc{wang2025monosropenvocabularyspatialreasoning,
      title={MonoSR: Open-Vocabulary Spatial Reasoning from Monocular Images}, 
      author={Qirui Wang and Jingyi He and Yining Pan and Si Yong Yeo and Xulei Yang and Shijie Li},
      year={2025},
      eprint={2511.19119},
      archivePrefix={arXiv},
      primaryClass={cs.CV},
      url={https://arxiv.org/abs/2511.19119}, 
}

@inproceedings{jia-etal-2026-scout,
    title = "{SCOUT}: Selective Coupling via Optimal Unbalanced Transport for Interpretable Text Classification",
    author = "Jia, Junhao  and
      Zheng, Hanwen  and
      Wu, Yueyi  and
      Chen, Huangwei  and
      Wang, Haishuai  and
      Bu, Jiajun  and
      Wu, Lei",
    booktitle = "Proceedings of the 64th Annual Meeting of the {A}ssociation for {C}omputational {L}inguistics (Volume 1: Long Papers)",
    month = jul,
    year = "2026",
    publisher = "Association for Computational Linguistics",
    url = "https://aclanthology.org/2026.acl-long.290/",
    pages = "6413--6431",
    ISBN = "979-8-89176-390-6",
}

@inproceedings{jia2026geodesic,
  title={Geodesic prototype matching via diffusion maps for interpretable fine-grained recognition},
  author={Jia, Junhao and Liu, Yunyou and Sun, Yifei and Chen, Huangwei and Qin, Feiwei and Wang, Changmiao and Peng, Yong},
  booktitle={ICASSP 2026-2026 IEEE International Conference on Acoustics, Speech and Signal Processing (ICASSP)},
  pages={1786--1790},
  year={2026},
  organization={IEEE}
}

@article{jia2026looks,
  title={This Looks Distinctly Like That: Grounding Interpretable Recognition in Stiefel Geometry against Neural Collapse},
  author={Jia, Junhao and Wang, Jiaqi and Liu, Yunyou and Jing, Haodong and Wu, Yueyi and Wu, Xian and Zheng, Yefeng},
  journal={arXiv preprint arXiv:2603.08374},
  year={2026}
}

@article{jia2026unsupervised,
  title={Unsupervised causal prototypical networks for de-biased interpretable dermoscopy diagnosis},
  author={Jia, Junhao and Wu, Yueyi and Chen, Huangwei and Jing, Haodong and Wang, Haishuai and Bu, Jiajun and Wu, Lei},
  journal={arXiv preprint arXiv:2602.23752},
  year={2026}
}

@misc{chi2026driverwmdrivercentrictrafficconditionedlatent,
      title={Driver-WM: A Driver-Centric Traffic-Conditioned Latent World Model for In-Cabin Dynamics Rollout}, 
      author={Haozhuang Chi and Daosheng Qiu and Hao Su and Haochen Liu and Zirui Li and Haoruo Zhang and Chen Lv},
      year={2026},
      eprint={2605.05092},
      archivePrefix={arXiv},
      primaryClass={cs.RO},
      url={https://arxiv.org/abs/2605.05092}, 
}

@article{zheng2026clinical,
  title={Clinical Cognition Alignment for Gastrointestinal Diagnosis with Multimodal LLMs},
  author={Zheng, Huan and Zhou, Yucheng and Yan, Tianyi and Chen, Dubing and Lu, Hongbo and Liao, Wenlong and He, Tao and Peng, Pai and Shen, Jianbing},
  journal={arXiv preprint arXiv:2603.20698},
  year={2026}
}

@inproceedings{jiang2026agentic,
  title={Agentic ai as a cybersecurity attack surface: Threats, exploits, and defenses in runtime supply chains},
  author={Jiang, Xiaochong and Yang, Shiqi and Yang, Wenting and Liu, Yichen and Ji, Cheng},
  booktitle={2026 IEEE Conference on Artificial Intelligence (CAI)},
  pages={2142--2149},
  year={2026},
  organization={IEEE}
}

@inproceedings{yang2025training,
  title={Training Strategies for Speech Large Language Models: A Comprehensive Survey},
  author={Yang, Shiqi and Huang, Ziyi and Xiao, Wengran},
  booktitle={2025 3rd International Conference on Foundation and Large Language Models (FLLM)},
  pages={161--171},
  year={2025},
  organization={IEEE}
}

@inproceedings{hu2026illava,
  title={illava: An image is worth fewer than 1/3 input tokens in large multimodal models},
  author={Hu, Lianyu and Gao, Liqing and Shang, Fanhua and Wan, Liang and Feng, Wei},
  booktitle={International Conference on Learning Representation},
  year={2026}
}

@inproceedings{hu2026tvi,
  title={TVI-CoT: Text-Visual Interleaved Chain-of-Thought Reasoning for Multimodal Understanding},
  author={Hu, Lianyu and Ma, Xiaoyu and Liao, Zeqin and Liu, Yang},
  booktitle={International Conference on Machine Learning},
  year={2026}
}

@article{hu2024deep,
  title={Deep correlated prompting for visual recognition with missing modalities},
  author={Hu, Lianyu and Shi, Tongkai and Feng, Wei and Shang, Fanhua and Wan, Liang},
  journal={Advances in Neural Information Processing Systems},
  volume={37},
  pages={67446--67466},
  year={2024}
}

@inproceedings{li2026spacedrive,
  title={Spacedrive: Infusing spatial awareness into vlm-based autonomous driving},
  author={Li, Peizheng and Zhang, Zhenghao and Holtz, David and Yu, Hang and Yang, Yutong and Lai, Yuzhi and Song, Rui and Geiger, Andreas and Zell, Andreas},
  booktitle={Proceedings of the IEEE/CVF Conference on Computer Vision and Pattern Recognition},
  pages={40096--40107},
  year={2026}
}

\end{document}